\documentclass{article}
\usepackage[utf8]{inputenc}

\usepackage[T1]{fontenc}
\usepackage[english]{babel}
\usepackage{physics,amsmath,amssymb,amsthm,float,tikz,bm,xcolor,amsfonts,mathtools,csquotes,wasysym,cancel,adjustbox,tikzscale}

\usepackage[backend=biber,style=numeric]{biblatex}
\usepackage{caption}
\usepackage{placeins}
\usepackage{array}
\usepackage{authblk}
\usepackage{algorithm}
\usepackage{algpseudocode,setspace}
\usepackage{url,hyperref,lineno,subcaption} 
\usepackage{mdframed}
\usepackage{upgreek}
\usepackage{acro}
\usepackage{xspace}
\usepackage{acro}

\DeclareAcronym{efe}{
    short = EFE,
    long = expected free energy,
    }

\newcommand{\efe}{\ac{efe}\xspace}

\DeclareAcronym{bfe}{
    short = BFE,
    long = Bethe free energy
    }

\newcommand{\bfe}{\ac{bfe}\xspace}

\DeclareAcronym{cbfe}{
    short = CBFE,
    long = constrained Bethe free energy
    }

\newcommand{\cbfe}{\ac{cbfe}\xspace}

\DeclareAcronym{vfe}{
    short = VFE,
    long = variational free energy
    }

\newcommand{\vfe}{\ac{vfe}\xspace}

\DeclareAcronym{gfe}{
    short = GFE,
    long = generalised free energy
    }

\newcommand{\gfe}{\ac{gfe}\xspace}

\DeclareAcronym{aif}{
    short = {AIF},
    long = {Active Inference}
    }

\newcommand{\aif}{\ac{aif}\xspace}
\newcommand{\Aif}{\Ac{aif}\xspace}

\DeclareAcronym{ml}{
    short = {ML},
    long = {machine learning}
    }

\DeclareAcronym{fep}{
    short = FEP,
    long = Free Energy Principle
    }
\newcommand{\fep}{\ac{fep}\xspace}

\DeclareAcronym{ness}{
    short = NESS,
    long = non-equilibrium steady state
    }

\DeclareAcronym{FFG}{
    short = FFG,
    long = Forney-style factor graph
    }
\newcommand{\ffg}{\ac{FFG}\xspace}

\newcommand{\ffgs}{\acp{FFG}\xspace}
\newcommand{\Ffgs}{\Acp{FFG}\xspace}

\DeclareAcronym{cffg}{
    short = CFFG,
    long = constrained Forney-style factor graph
    }

\newcommand{\cffg}{\ac{cffg}\xspace}

\newcommand{\cffgs}{\acp{cffg}\xspace}
\newcommand{\Cffgs}{\Acp{cffg}\xspace}

\DeclareAcronym{bp}{
    short = BP,
    long = belief propagation
    }

\DeclareAcronym{VMP}{
    short = VMP,
    long = variational message passing
    }
\newcommand{\vmp}{\ac{VMP}\xspace}

\DeclareAcronym{mp}{
    short = MP,
    long = message passing
    }
\newcommand{\mpa}{\ac{mp}\xspace}
\newcommand{\Mpa}{\Ac{mp}\xspace}

\DeclareAcronym{EP}{
    short = EP,
    long = expectation propagation
    }
\newcommand{\ep}{\ac{EP}\xspace}

\DeclareAcronym{SI}{
    short = SI,
    long = sophisticated inference
    }

\DeclareAcronym{laif}{
    short = LAIF,
    long = Lagrangian Active Inference
    }
\newcommand{\laif}{\ac{laif}\xspace}
\newcommand{\Laif}{\Ac{laif}\xspace}

\DeclareAcronym{POMDP}{
    short = POMDP,
    long = partially observed Markov decision process
    }
\newcommand{\pomdp}{\ac{POMDP}\xspace}

\DeclareAcronym{HMM}{
    short = HMM,
    long = hidden Markov model
    }

\DeclareAcronym{GMM}{
    short = GMM,
    long = Gaussian mixture model
    }

\DeclareAcronym{GMMM}{
    short = $\Gamma$MM,
    long = Gamma mixture model
    }

\DeclareAcronym{AR}{
    short = AR,
    long = auto-regressive
    }

\DeclareAcronym{TVAR}{
    short = TVAR,
    long = time-varying auto-regressive
    }

\DeclareAcronym{KL}{
    short = KL,
    long = Kullback-Leibler divergence
    }
\newcommand{\kl}{\ac{KL}\xspace}

\DeclareAcronym{ha}{
    short = HA,
    long = hearing aid
    }

\DeclareAcronym{GP}{
    short = GP,
    long = Gaussian process
    }

\DeclareAcronym{gpc}{
    short = GPC,
    long = Gaussian process classifier
    }

\DeclareAcronym{mi}{
    short = MI,
    long = mutual information
    }
\newcommand{\mi}{\ac{mi}\xspace}

\DeclareAcronym{lgds}{
    short = LGDS,
    long = linear Gaussian dynamical system
    }

\usepackage[disable]{todonotes}

\usetikzlibrary{shapes.geometric} 
\usetikzlibrary{positioning}
\usetikzlibrary{calc}
\usetikzlibrary{intersections}
\usepackage{tkz-euclide}
\tikzset{mainstyle/.style={fill=white, draw=black, shape=rectangle, align=center}}

\tikzset{dstyle/.style={mainstyle, minimum size=5mm, inner sep=0pt, text width=4mm}}

\tikzset{astyle/.style={fill=black, draw=black, shape=diamond, minimum size=3mm, inner sep=0pt, text width=0mm}}

\tikzset{sstyle/.style={mainstyle, minimum size=7mm, inner sep=0pt, text width=5mm}}

\tikzset{ostyle/.style={fill=black, draw=black, shape=rectangle, minimum size=0.2cm, inner sep=0pt, text width=2mm}}

\tikzset{sophstyle/.style={fill=white,draw=black, shape=rectangle, minimum size=0.2cm, inner sep=0pt, text width=2mm}}

\tikzset{actionstyle/.style={fill=black, draw=black, shape=diamond, minimum size=0.3cm, inner sep=0pt, text width=2mm}}

\tikzset{estyle/.style={fill=white, draw=black, shape=rectangle, minimum height=0.7cm, minimum width=2.5cm, inner sep=3pt}}

\tikzset{cstyle/.style={fill=white, draw=black, shape=rectangle, minimum height=0.7cm, minimum width=2cm, inner sep=3pt}}

\tikzset{bstyle/.style={fill=white, draw=black, shape=circle, minimum size=0.7cm, inner sep=0pt, text width=4.5mm}}

\tikzstyle{observation}=[ostyle]
\tikzstyle{deterministic}=[dstyle]
\tikzstyle{stochastic}=[sstyle]
\tikzstyle{action}=[astyle]
\tikzstyle{environment}=[estyle]
\tikzstyle{agent}=[bstyle]
\tikzstyle{control}=[cstyle]
\tikzstyle{action}=[actionstyle]
\tikzstyle{soph}=[sophstyle]

\tikzstyle{filter}=[mainstyle, minimum width=1cm, minimum height=0.5cm]
\tikzstyle{selector}=[fill=white, draw=black, shape=trapezium, rotate=180, minimum width=1cm, minimum height=0.5cm]

\renewcommand{\d}[1]{\operatorname{d}\!{#1}}

\newcommand{\Cat}[1]{\mathcal{C}at\!\left({#1}\right)}

\newcommand{\T}{\operatorname{T}}

\newcommand{\GFE}[1]{\mathcal{G}\!\left[\, {#1} \,\right]}

\renewcommand{\H}[1]{\operatorname{H}\!\left[{#1}\right]}

\renewcommand{\bar}[1]{\overline{#1}}
\newcommand{\diag}[1]{\operatorname{diag}\!\left({#1}\right)}

\newcommand{\I}[1]{\operatorname{I}\!\left[\,{#1}\,\right]} 
\newcommand{\seq}[1]{\bm{#1}} 

\newcommand{\vect}[1]{\bm{\mathrm{#1}}} 
\newcommand{\matr}[1]{\bm{\mathrm{#1}}}

\newmdtheoremenv{boxthm}{Theorem}
\newmdtheoremenv{boxlem}{Lemma}
\newmdtheoremenv{boxcor}{Corollary}
\newmdtheoremenv{boxdef}{Definition}

\usepackage{tikz,colortbl}
\usepackage{xifthen}
\usetikzlibrary{intersections}
\usetikzlibrary{arrows.meta}
\usepackage{zref-savepos}
\usepackage{tkz-euclide}

\pgfdeclarelayer{bg}    
\pgfsetlayers{bg,main}  
\definecolor{beige}{RGB}{245, 245, 220}
\definecolor{darkgrey}{RGB}{65, 65, 65}
\definecolor{grey}{RGB}{240, 240, 240}
\definecolor{lightgrey}{RGB}{250, 250, 250}

\newcommand*\tinydarkcircled[1]{\tikz[baseline=(char.base)]{
            \node[shape=circle,draw,inner sep=0pt, text width=2mm, align=center, minimum width=2mm, fill=darkgrey, text=white, font=\bfseries] (char) {\tiny{#1}};}}

\usetikzlibrary{calc, arrows, fit, positioning, patterns, decorations.pathreplacing, shapes}
\tikzstyle{dash} = [dashed]
\tikzstyle{line} = [draw]
\tikzstyle{box} = [draw, minimum size=.8cm]
\tikzstyle{bigbox} = [draw, minimum size=1.6cm]
\tikzstyle{detbox} = [draw, minimum size=.5cm] 
\tikzstyle{smallbox} = [draw, minimum size=5mm]
\tikzstyle{roundbox} = [draw, circle, inner sep=0pt, minimum size=.8cm]
\tikzstyle{clamped} = [draw, fill=black, minimum size=0.35cm]
\tikzstyle{roundclamped} = [draw, circle, color=black, inner sep=0pt, fill=black, text=white, minimum size=0.35cm]
\tikzstyle{optim} = [draw, circle, inner sep=0pt, fill=white, minimum size=0.35cm]
\tikzstyle{msgcircle} = [shape=circle, draw, inner sep=0pt, minimum size=5mm, fill=white]
\tikzstyle{darkmsgcircle} = [shape=circle, draw, inner sep=0pt, minimum size=5mm, fill=darkgrey, text=white, font=\bfseries]
\tikzstyle{msgdoublecircle} = [shape=circle, double, double distance=1.5pt, draw, inner sep=0pt, minimum size=5mm, fill=white]
\tikzstyle{darkmsgdoublecircle} = [shape=circle, double, double distance=1.5pt, draw, inner sep=0pt, minimum size=5mm, fill=darkgrey, text=white, font=\bfseries]

\tikzstyle{q} = [draw, shape=circle, minimum size=3mm, fill=white, anchor=center, inner sep=0pt, text=black] 
\tikzstyle{Q} = [draw, shape=circle, minimum size=6mm, fill=white, anchor=center, inner sep=0pt, text=black] 
\tikzstyle{p} = [draw, minimum size=2.8mm, fill=white, anchor=center, inner sep=0pt, text=black] 
\tikzstyle{v} = [circle split, rotate=90, anchor=center, minimum size=0mm, inner sep=0pt] 
\tikzstyle{h} = [circle split, anchor=center, minimum size=0mm, inner sep=0pt] 
\tikzstyle{s} = [draw, shape=circle, minimum size=3mm, fill=white, anchor=center, inner sep=0pt] 
\tikzstyle{f} = [draw, shape=circle, minimum size=3mm, fill=black, anchor=center, inner sep=0pt, text=white, font=\bfseries] 
\tikzstyle{bc} = [draw, shape=circle, minimum size=6mm, fill=black, anchor=center, inner sep=0pt, text=white, font=\bfseries] 
\tikzstyle{wc} = [draw, shape=circle, minimum size=6mm, fill=white, anchor=center, inner sep=0pt, text=black, font=\bfseries] 
\tikzstyle{data} = [draw, shape=circle, minimum size=3mm, fill=black, anchor=center, inner sep=0pt, text=white, font=\bfseries] 
\tikzstyle{delta} = [draw, shape=circle, minimum size=3mm, fill=white, anchor=center, inner sep=0pt, text=black, font=\bfseries] 

\tikzset{bigsstyle/.style={mainstyle, minimum size=9mm, inner sep=0pt, text width=5mm}}
\tikzset{medsstyle/.style={mainstyle, minimum size=8mm, inner sep=0pt, text width=5mm}}

\newcommand{\arcnw}[1]{
    \draw[black] ($({#1}.west) + (0.05,0.0)$) arc[start angle=180, end angle=90, radius=3.0mm];
    \path[line] ({#1}.west) edge[-, draw=black] ($({#1}.west) + (0.05,0.0)$);
    \path[line] ({#1}.north) edge[-, draw=black] ($({#1}.north) - (0.0,0.06)$);
}

\newcommand{\arcne}[1]{
    \draw[black] ($({#1}.north) - (0.0,0.05)$) arc[start angle=90, end angle=0, radius=3.1mm];
    \path[line] ({#1}.north) edge[-, draw=black] ($({#1}.north) - (0.0,0.05)$);
    \path[line] ({#1}.east) edge[-, draw=black] ($({#1}.east) - (0.05,0.0)$);
}

\newcommand{\arcs}[1]{
    \draw[black] ($({#1}.west) + (0.05,0.0)$) arc[start angle=-180, end angle=0, radius=3.0mm];
    \path[line] ({#1}.west) edge[-, draw=black] ($({#1}.west) + (0.05,0.0)$);
    \path[line] ({#1}.east) edge[-, draw=black] ($({#1}.east) + (-0.06,0.0)$);
}

\newcommand{\bigarcnw}[1]{
    \draw[black] ($({#1}.west) + (0.3,0.0)$) arc[start angle=180, end angle=90, radius=5.1mm];
    \path[line] ({#1}.west) edge[-, draw=black] ($({#1}.west) + (0.3,0.0)$);
    \path[line] ({#1}.north) edge[-, draw=black] ($({#1}.north) - (0.0,0.3)$);
}

\newcommand{\bigarcne}[1]{
    \draw[black] ($({#1}.north) + (0.0,-0.3)$) arc[start angle=90, end angle=0, radius=5.1mm];
    \path[line] ({#1}.east) edge[-, draw=black] ($({#1}.east) + (-0.3,0.0)$);
    \path[line] ({#1}.north) edge[-, draw=black] ($({#1}.north) - (0.0,0.3)$);
}

\newcommand{\bigarcsw}[1]{
    \draw[black] ($({#1}.south) + (0.0,0.3)$) arc[start angle=-90, end angle=-180, radius=5.1mm];
    \path[line] ({#1}.west) edge[-, draw=black] ($({#1}.west) + (0.3,0.0)$);
    \path[line] ({#1}.south) edge[-, draw=black] ($({#1}.south) - (0.0,-0.3)$);
}

\newcommand{\bigarcse}[1]{
    \draw[black] ($({#1}.east) + (-0.3,0.0)$) arc[start angle=0, end angle=-90, radius=5.1mm];
    \path[line] ({#1}.east) edge[-, draw=black] ($({#1}.east) + (-0.3,0.0)$);
    \path[line] ({#1}.south) edge[-, draw=black] ($({#1}.south) - (0.0,-0.3)$);
}

\newcommand{\bigarcs}[1]{
    \draw[black] ($({#1}.west) + (0.3,0.0)$) arc[start angle=-180, end angle=0, radius=5.1mm];
    \path[line] ({#1}.west) edge[-, draw=black] ($({#1}.west) + (0.3,0.0)$);
    \path[line] ({#1}.east) edge[-, draw=black] ($({#1}.east) + (-0.3,0.0)$);
}

\newcommand{\bigstocarcnw}[1]{
    \draw[black] ($({#1}.west) + (0.05,0.0)$) arc[start angle=180, end angle=90, radius=3.5mm];
    \path[line] ({#1}.west) edge[-, draw=black] ($({#1}.west) + (0.05,0.0)$);
    \path[line] ({#1}.north) edge[-, draw=black] ($({#1}.north) + (0.00,-0.06)$);
}

\newcommand{\bigstocarcne}[1]{
    \draw[black] ($({#1}.north) + (0.0,-0.05)$) arc[start angle=90, end angle=0, radius=3.5mm];
    \path[line] ({#1}.north) edge[-, draw=black] ($({#1}.north) + (0.0,-0.05)$);
    \path[line] ({#1}.east) edge[-, draw=black] ($({#1}.east) + (-0.06,0.0)$);
}

\newcommand{\msg}[6]{
      \ifthenelse{\isin{#1}{left} \AND \isin{#2}{down}}{
            \coordinate (anchor) at ($({#3})!{#5}!({#4})$);
            \node[msgcircle, xshift=-5.5mm] at (anchor) {#6};
            \node[xshift=-1.5mm] at (anchor) {$\downarrow$};
      }{}
      \ifthenelse{\isin{#1}{right} \AND \isin{#2}{down}}{
            \coordinate (anchor) at ($({#3})!{#5}!({#4})$);
            \node[msgcircle, xshift=5.5mm] at (anchor) {#6};
            \node[xshift=1.5mm] at (anchor) {$\downarrow$};
      }{}

      \ifthenelse{\isin{#1}{down} \AND \isin{#2}{right}}{
            \coordinate (anchor) at ($({#3})!{#5}!({#4})$);
            \node[msgcircle, yshift=-6.0mm] at (anchor) {#6};
            \node[yshift=-2.0mm] at (anchor) {$\rightarrow$};
      }{}
      \ifthenelse{\isin{#1}{up} \AND \isin{#2}{right}}{
            \coordinate (anchor) at ($({#3})!{#5}!({#4})$);
            \node[msgcircle, yshift=6.0mm] at (anchor) {#6};
            \node[yshift=2.0mm] at (anchor) {$\rightarrow$};
      }{}

      \ifthenelse{\isin{#1}{down} \AND \isin{#2}{left}}{
            \coordinate (anchor) at ($({#3})!{#5}!({#4})$);
            \node[msgcircle, yshift=-6.0mm] at (anchor) {#6};
            \node[yshift=-2.0mm] at (anchor) {$\leftarrow$};
      }{}
      \ifthenelse{\isin{#1}{up} \AND \isin{#2}{left}}{
            \coordinate (anchor) at ($({#3})!{#5}!({#4})$);
            \node[msgcircle, yshift=6.0mm] at (anchor) {#6};
            \node[yshift=2.0mm] at (anchor) {$\leftarrow$};
      }{}

      \ifthenelse{\isin{#1}{left} \AND \isin{#2}{up}}{
            \coordinate (anchor) at ($({#3})!{#5}!({#4})$);
            \node[msgcircle, xshift=-5.5mm] at (anchor) {#6};
            \node[xshift=-1.5mm] at (anchor) {$\uparrow$};
      }{}
      \ifthenelse{\isin{#1}{right} \AND \isin{#2}{up}}{
            \coordinate (anchor) at ($({#3})!{#5}!({#4})$);
            \node[msgcircle, xshift=5.5mm] at (anchor) {#6};
            \node[xshift=1.5mm] at (anchor) {$\uparrow$};
      }{}
}

\newcommand{\darkmsg}[6]{
      \ifthenelse{\isin{#1}{left} \AND \isin{#2}{down}}{
            \coordinate (anchor) at ($({#3})!{#5}!({#4})$);
            \node[darkmsgcircle, xshift=-5.5mm] at (anchor) {#6};
            \node[xshift=-1.5mm] at (anchor) {$\downarrow$};
      }{}
      \ifthenelse{\isin{#1}{right} \AND \isin{#2}{down}}{
            \coordinate (anchor) at ($({#3})!{#5}!({#4})$);
            \node[darkmsgcircle, xshift=5.5mm] at (anchor) {#6};
            \node[xshift=1.5mm] at (anchor) {$\downarrow$};
      }{}

      \ifthenelse{\isin{#1}{down} \AND \isin{#2}{right}}{
            \coordinate (anchor) at ($({#3})!{#5}!({#4})$);
            \node[darkmsgcircle, yshift=-6.0mm] at (anchor) {#6};
            \node[yshift=-2.0mm] at (anchor) {$\rightarrow$};
      }{}
      \ifthenelse{\isin{#1}{up} \AND \isin{#2}{right}}{
            \coordinate (anchor) at ($({#3})!{#5}!({#4})$);
            \node[darkmsgcircle, yshift=6.0mm] at (anchor) {#6};
            \node[yshift=2.0mm] at (anchor) {$\rightarrow$};
      }{}

      \ifthenelse{\isin{#1}{down} \AND \isin{#2}{left}}{
            \coordinate (anchor) at ($({#3})!{#5}!({#4})$);
            \node[darkmsgcircle, yshift=-6.0mm] at (anchor) {#6};
            \node[yshift=-2.0mm] at (anchor) {$\leftarrow$};
      }{}
      \ifthenelse{\isin{#1}{up} \AND \isin{#2}{left}}{
            \coordinate (anchor) at ($({#3})!{#5}!({#4})$);
            \node[darkmsgcircle, yshift=6.0mm] at (anchor) {#6};
            \node[yshift=2.0mm] at (anchor) {$\leftarrow$};
      }{}

      \ifthenelse{\isin{#1}{left} \AND \isin{#2}{up}}{
            \coordinate (anchor) at ($({#3})!{#5}!({#4})$);
            \node[darkmsgcircle, xshift=-5.5mm] at (anchor) {#6};
            \node[xshift=-1.5mm] at (anchor) {$\uparrow$};
      }{}
      \ifthenelse{\isin{#1}{right} \AND \isin{#2}{up}}{
            \coordinate (anchor) at ($({#3})!{#5}!({#4})$);
            \node[darkmsgcircle, xshift=5.5mm] at (anchor) {#6};
            \node[xshift=1.5mm] at (anchor) {$\uparrow$};
      }{}
}

\newcommand{\bwmsg}[6]{
      \ifthenelse{\isin{#1}{left} \AND \isin{#2}{down}}{
            \coordinate (anchor) at ($({#3})!{#5}!({#4})$);
            \node[msgdoublecircle, xshift=-5.5mm] at (anchor) {#6};
            \node[xshift=-1.5mm] at (anchor) {$\downarrow$};
      }{}
      \ifthenelse{\isin{#1}{right} \AND \isin{#2}{down}}{
            \coordinate (anchor) at ($({#3})!{#5}!({#4})$);
            \node[msgdoublecircle, xshift=5.5mm] at (anchor) {#6};
            \node[xshift=1.5mm] at (anchor) {$\downarrow$};
      }{}

      \ifthenelse{\isin{#1}{down} \AND \isin{#2}{right}}{
            \coordinate (anchor) at ($({#3})!{#5}!({#4})$);
            \node[msgdoublecircle, yshift=-6.0mm] at (anchor) {#6};
            \node[yshift=-2.0mm] at (anchor) {$\rightarrow$};
      }{}
      \ifthenelse{\isin{#1}{up} \AND \isin{#2}{right}}{
            \coordinate (anchor) at ($({#3})!{#5}!({#4})$);
            \node[msgdoublecircle, yshift=6.0mm] at (anchor) {#6};
            \node[yshift=2.0mm] at (anchor) {$\rightarrow$};
      }{}

      \ifthenelse{\isin{#1}{down} \AND \isin{#2}{left}}{
            \coordinate (anchor) at ($({#3})!{#5}!({#4})$);
            \node[msgdoublecircle, yshift=-6.0mm] at (anchor) {#6};
            \node[yshift=-2.0mm] at (anchor) {$\leftarrow$};
      }{}
      \ifthenelse{\isin{#1}{up} \AND \isin{#2}{left}}{
            \coordinate (anchor) at ($({#3})!{#5}!({#4})$);
            \node[msgdoublecircle, yshift=6.0mm] at (anchor) {#6};
            \node[yshift=2.0mm] at (anchor) {$\leftarrow$};
      }{}

      \ifthenelse{\isin{#1}{left} \AND \isin{#2}{up}}{
            \coordinate (anchor) at ($({#3})!{#5}!({#4})$);
            \node[msgdoublecircle, xshift=-5.5mm] at (anchor) {#6};
            \node[xshift=-1.5mm] at (anchor) {$\uparrow$};
      }{}
      \ifthenelse{\isin{#1}{right} \AND \isin{#2}{up}}{
            \coordinate (anchor) at ($({#3})!{#5}!({#4})$);
            \node[msgdoublecircle, xshift=5.5mm] at (anchor) {#6};
            \node[xshift=1.5mm] at (anchor) {$\uparrow$};
      }{}
}

\newcommand{\bwdarkmsg}[6]{
      \ifthenelse{\isin{#1}{left} \AND \isin{#2}{down}}{
            \coordinate (anchor) at ($({#3})!{#5}!({#4})$);
            \node[darkmsgdoublecircle, xshift=-5.5mm] at (anchor) {#6};
            \node[xshift=-1.5mm] at (anchor) {$\downarrow$};
      }{}
      \ifthenelse{\isin{#1}{right} \AND \isin{#2}{down}}{
            \coordinate (anchor) at ($({#3})!{#5}!({#4})$);
            \node[darkmsgdoublecircle, xshift=5.5mm] at (anchor) {#6};
            \node[xshift=1.5mm] at (anchor) {$\downarrow$};
      }{}

      \ifthenelse{\isin{#1}{down} \AND \isin{#2}{right}}{
            \coordinate (anchor) at ($({#3})!{#5}!({#4})$);
            \node[darkmsgdoublecircle, yshift=-6.0mm] at (anchor) {#6};
            \node[yshift=-2.0mm] at (anchor) {$\rightarrow$};
      }{}
      \ifthenelse{\isin{#1}{up} \AND \isin{#2}{right}}{
            \coordinate (anchor) at ($({#3})!{#5}!({#4})$);
            \node[darkmsgdoublecircle, yshift=6.0mm] at (anchor) {#6};
            \node[yshift=2.0mm] at (anchor) {$\rightarrow$};
      }{}

      \ifthenelse{\isin{#1}{down} \AND \isin{#2}{left}}{
            \coordinate (anchor) at ($({#3})!{#5}!({#4})$);
            \node[darkmsgdoublecircle, yshift=-6.0mm] at (anchor) {#6};
            \node[yshift=-2.0mm] at (anchor) {$\leftarrow$};
      }{}
      \ifthenelse{\isin{#1}{up} \AND \isin{#2}{left}}{
            \coordinate (anchor) at ($({#3})!{#5}!({#4})$);
            \node[darkmsgdoublecircle, yshift=6.0mm] at (anchor) {#6};
            \node[yshift=2.0mm] at (anchor) {$\leftarrow$};
      }{}

      \ifthenelse{\isin{#1}{left} \AND \isin{#2}{up}}{
            \coordinate (anchor) at ($({#3})!{#5}!({#4})$);
            \node[darkmsgdoublecircle, xshift=-5.5mm] at (anchor) {#6};
            \node[xshift=-1.5mm] at (anchor) {$\uparrow$};
      }{}
      \ifthenelse{\isin{#1}{right} \AND \isin{#2}{up}}{
            \coordinate (anchor) at ($({#3})!{#5}!({#4})$);
            \node[darkmsgdoublecircle, xshift=5.5mm] at (anchor) {#6};
            \node[xshift=1.5mm] at (anchor) {$\uparrow$};
      }{}
}

\tikzset{mainstyle/.style={fill=white, draw=black, shape=rectangle, align=center}}

\tikzset{dstyle/.style={mainstyle, minimum size=5mm, inner sep=0pt, text width=4mm}}

\tikzset{astyle/.style={fill=black, draw=black, shape=diamond, minimum size=3mm, inner sep=0pt, text width=0mm}}

\tikzset{sstyle/.style={mainstyle, minimum size=7mm, inner sep=0pt}}

\tikzset{ostyle/.style={fill=black, draw=black, shape=rectangle, minimum size=0.2cm, inner sep=0pt, text width=2mm}}

\tikzset{sophstyle/.style={fill=white,draw=black, shape=rectangle, minimum size=0.2cm, inner sep=0pt, text width=2mm}}

\tikzset{actionstyle/.style={fill=black, draw=black, shape=diamond, minimum size=0.3cm, inner sep=0pt, text width=2mm}}

\tikzset{estyle/.style={fill=white, draw=black, shape=rectangle, minimum height=0.7cm, minimum width=2.5cm, inner sep=3pt}}

\tikzset{cstyle/.style={fill=white, draw=black, shape=rectangle, minimum height=0.7cm, minimum width=2cm, inner sep=3pt}}

\tikzset{bstyle/.style={fill=white, draw=black, shape=circle, minimum size=0.7cm, inner sep=0pt, text width=4.5mm}}

\tikzset{bigsstyle/.style={mainstyle, minimum size=9mm, inner sep=0pt}}

\tikzstyle{observation}=[ostyle]
\tikzstyle{deterministic}=[dstyle]
\tikzstyle{stochastic}=[sstyle]
\tikzstyle{bigstoc}=[bigsstyle]
\tikzstyle{medstoc}=[medsstyle]
\tikzstyle{action}=[astyle]
\tikzstyle{environment}=[estyle]
\tikzstyle{agent}=[bstyle]
\tikzstyle{control}=[cstyle]
\tikzstyle{action}=[actionstyle]
\tikzstyle{soph}=[sophstyle]

\tikzstyle{filter}=[mainstyle, minimum width=1cm, minimum height=0.5cm]
\tikzstyle{selector}=[fill=white, draw=black, shape=trapezium, rotate=180, minimum width=1cm, minimum height=0.5cm]

\title{Realising Synthetic Active Inference Agents,\\ Part I: Epistemic Objectives and Graphical Specification Language}
\author[1,2]{Magnus Koudahl}
\author[1]{Thijs van de Laar}
\author[1,3]{Bert de Vries}
\date{\today}
\affil[1]{BIASLab, Department of Electrical Engineering, Eindhoven University of Technology, The Netherlands}
\affil[2]{VERSES AI Research Lab, Los Angeles, CA, USA, 90016}
\affil[3]{GN Hearing, JF Kennedylaan 2, 5612 AB Eindhoven, The Netherlands}

\begin{document}

\maketitle


\begin{abstract}
    The \fep is a theoretical framework for describing how (intelligent) systems self-organise into coherent, stable structures by minimising a free energy functional. \Aif is a corollary of the \fep that specifically details how systems that are able to plan for the future (agents) function by minimising particular free energy functionals that incorporate information seeking components.
    This paper is the first in a series of two where we derive a synthetic version of \aif on free form factor graphs. The present paper focuses on deriving a local version of the free energy functionals used for \aif. This enables us to construct a version of \aif which applies to arbitrary graphical models and interfaces with prior work on \mpa algorithms. The resulting messages are derived in our companion paper.
    We also identify a gap in the graphical notation used for factor graphs. While factor graphs are great at expressing a generative model, they have so far been unable to specify the full optimisation problem including constraints. To solve this problem we develop \cffg notation which permits a fully graphical description of variational inference objectives.
    We then proceed to show how \cffgs can be used to reconstruct prior algorithms for \aif as well as derive new ones. The latter is demonstrated by deriving an algorithm that permits direct policy inference for \aif agents, circumventing a long standing scaling issue that has so far hindered the application of \aif in industrial settings. We demonstrate our algorithm on the classic T-maze task and show that it reproduces the information seeking behaviour that is a hallmark feature of \aif.
\end{abstract}

\section{Introduction}
\Acf{aif} is an emerging framework for modelling intelligent agents interacting with an environment. Originating in the field of computational neuroscience, it has since been spread to numerous other fields such as modern machine learning. At its core, \aif relies on variational inference techniques to minimise a free energy functional. A key differentiator of \aif compared to other approaches is the use of custom free energy functionals such as Expected and Generalised free energies (\efe and \gfe, respectively). These functionals are specifically constructed so as to elicit epistemic, information-seeking behaviour when used to infer actions.

Optimisation of these functionals have so far relied on custom algorithms that require evaluating very large search trees which has rendered upscaling of \aif difficult. Many recent works, such as Branching Time \aif \cite{champion_2021_btai} and \acl{SI} \cite{friston_sophisticated_2021} have investigated algorithmic ways to prune the search tree in order to solve this problem.

This paper is the first in a series of two parts where we take a different approach and formulate a variation of the \gfe optimisation problem using a custom Lagrangian derived from \cbfe on factor graphs. Using variational calculus we then derive a custom \mpa algorithm that can directly solve for fixed points of local \gfe terms, removing the need for a search tree. This allows us to construct a purely optimisation-based approach to \aif which we name \laif.

\Laif applies to arbitrary graph topologies and interfaces with generic \mpa algorithms which allow for scaling up of \aif using off-the-shelf tools. We accomplish this by constructing a node-local \gfe on generic factor graphs.

The present paper is structured as follows: In section~\ref{sec:lagrangian}, we review relevant background material concerning \acfp{FFG} and \acf{bfe}. In section~\ref{sec:epistemics} we formalise what we mean by "epistemics" and construct an objective that is local to a single node on an \ffg and possesses an epistemic, information-seeking drive. This objective turns out to be a local version of the \gfe which we review in section \ref{sec:gfe}.

Once we start modifying the free energy functional, we recognise a problem with current \ffg notation. \Ffgs visualise generative models but fail to display a significant part of the optimisation problem, namely, the variational distribution and the functional to be optimised.

To remedy this, we develop the \cffg graphical notation in section~\ref{sec:CFFG} as a method for visualising both the variational distribution and any adaptations to the free energy functional that are needed for \aif. These tools form the basis of the update rules derived in our companion paper \cite{vandelaar2023realising}.

With the \cffg notation in hand, in section~\ref{sec:special_cases} we then proceed to demonstrate how to recover prior algorithms for \aif as \mpa on a \cffg. \Aif is often described as \mpa on a probabilistic graphical model, see \cite{da_costa_active_2020, friston_active_2015, koudahl2023message,smith_step-by-step_2021} for examples. However, this relationship has not been properly formalised before, in part because adequate notation has been lacking. Using \cffgs it is straightforward to accurately write down this relation. Further, due to the modular nature of \cffgs it becomes easy to devise extensions to prior \aif algorithms that can be implemented using off-the-shelf \mpa tools.

Finally, section~\ref{sec:experiments} demonstrates a new algorithm for policy inference using \laif that scales linearly in the planning horizon, providing a solution to a long-standing barrier for scaling \aif to larger models and more complex tasks.

\section{The Lagrangian approach to message passing}\label{sec:lagrangian}
In this section, we review the \acl{bfe} along with \ffgs. These concepts form the foundation from which we will build towards local epistemic, objectives.

As a free energy functional, the \bfe is unique because stationary points of the \bfe correspond to solutions of the \acl{bp} algorithm \cite{pearl_reverend_1982, senoz_variational_2021,zhang2021unifying} which provides exact inference on tree-structured graphs.

Furthermore, by adding constraints to the \bfe using Lagrange multipliers, one can form a custom Lagrangian for an inference problem. Taking the first variation of this Lagrangian, one can then solve for stationary points and obtain \mpa algorithms that solve the desired problem. Prior work \cite{senoz_variational_2021,zhang2021unifying} have shown that adding additional constraints to the \bfe allows for deriving a host of different message passing algorithms including \vmp \cite{winn_variational_2005} and \ep \cite{minka_expectation_2001} among others. We refer interested readers to \cite{senoz_variational_2021} for a comprehensive overview of this technique and how different choices of constraints can lead to different algorithms.

Constraints are specified at the level of nodes and edges, meaning this procedure can produce hybrid \mpa algorithms, foreshadowing the approach we are going to take for deriving a local, epistemic objective for \aif.

\subsection{Bethe free energy and Forney-style factor graphs}\label{sec:bfe_and_ffg}
Throughout the remainder of the paper we will use \ffgs to visualise probabilistic models models. In Section \ref{sec:CFFG} we extend this notation to additionally allow for specifying constraints on the variational optimisation objective.

Following \cite{senoz_variational_2021} we define an \ffg as a graph $\mathcal{G} = (\mathcal{V}, \mathcal{E})$ with nodes $\mathcal{V}$ and edges $\mathcal{E} \subseteq \mathcal{V}\times\mathcal{V}$. For a node $a\in \mathcal{V}$ we denote the connected edges by $\mathcal{E}(a)$. Similarly for an edge $i\in \mathcal{E}$, we denote the connected nodes by $\mathcal{V}(i)$.

An \ffg can be used to represent a factorised function (model) over variables $\seq{s}$, as

\begin{align}
    f(\seq{s}) = \prod_{a\in\mathcal{V}} f_a(\seq{s}_a)\,,\label{eq:f}
\end{align}

where $\seq{s}_a$ collects the argument variables of the factor $f_a$. Throughout this paper we will use $\seq{cursive \, bold}$ font to denote collections of variables. In the corresponding \ffg, the factor $f_a$ is denoted by a square node and connected edges $\mathcal{E}(a)$ represent the argument variables $\seq{s}_a$.

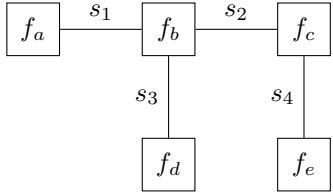
\begin{figure}[H]
    \centering
    \begin{tikzpicture}
[node distance=18mm, auto]

\node[stochastic](fb){$f_b$};
\node[stochastic, right of=fb](fc){$f_c$};

\node[stochastic, left of=fb](fa){$f_a$};

\node[stochastic, below of=fb](fd){$f_d$};

\node[stochastic, below of=fc](fe){$f_e$};

\node[below of=fb](ghost1){};

\draw[](fa)--node[above]{$s_1$}(fb);

\draw[](fb)--node[above]{$s_2$}(fc);

\draw[](fb)--node[left]{$s_3$}(fd);
\draw[](fc)--node[left]{$s_4$}(fe);

\end{tikzpicture}
    \caption{Example of an FFG}
    \label{fig:sample_ffg}
\end{figure}

As an example, we can consider the \ffg shown in Fig.~\ref{fig:sample_ffg} which corresponds to the model

\begin{align}
    f(s_1,s_2,s_3,s_4) &= f_a(s_1)f_b(s_1,s_2,s_3)f_c(s_2,s_4)f_d(s_3)f_e(s_4)
\end{align}

In Fig.~\ref{fig:sample_ffg}, the vertex set is $\mathcal{V} = \{ a, \dots, e\}$ and the edge set is $\mathcal{E} = \{ 1, \dots, 4 \}$. As an example, the neighbouring edges of the node $c$ are given by $\mathcal{E}(c) = \{ 2, 4 \}$. In this way, \ffgs allow for a simple visualisation of the factorisation properties of a high-dimensional function.

The problem we will focus on concerns minimisation of a free energy functional over a generative model. More formally, given a model \eqref{eq:f} and a "variational" distribution $q(\seq{s})$, the \vfe is defined as

\begin{align}
    F[q] \triangleq \int q(\seq{s})\log \frac{q(\seq{s})}{f(\seq{s})} \d{\seq{s}}\,.\label{eq:F}
\end{align}

Variational inference concerns minimising this functional, leading to the solution

\begin{align}
    q^{*} = \arg\min_{q\in\mathcal{Q}} F[q]\,,\label{eq:BFE_problem}
\end{align}

with $\mathcal{Q}$ denoting the admissible family of functions $q$. The optimised \vfe upper-bounds the negative log-evidence (surprisal), as

\begin{align}
    F[q^{*}] = \underbrace{\int q^{*}\!(\seq{s}) \log \frac{q^{*}\!(\seq{s})}{p(\seq{s})} \d{\seq{s}}}_{\text{Posterior divergence}} \underbrace{- \log Z}_{\text{Surprisal}}\,,
\end{align}

with $Z = \int f(\seq{s}) \dd \seq{s}$ is the model evidence and the exact posterior is given by $p(\seq{s}) = f(\seq{s})/Z$.

The \bfe applies the Bethe assumption to the factorisation of $q$ which yields an objective that decomposes into a sum of local free energy terms, each local to a node on the corresponding \ffg.
Each node local free energy will include entropy terms from all connected edges. Since an edge is connected to (at most) two nodes, that means the corresponding entropy term would be counted twice. To prevent overcounting of the edge entropies, the \bfe includes additional terms entropy terms that cancel out overcounted terms.

Under the Bethe approximation $q(\seq{s})$ is given by

\begin{align}\label{eq:bethe}
    q(\seq{s}) = \prod_{a\in\mathcal{V}} q_a(\seq{s}_a) \prod_{i\in\mathcal{E}} q_i(s_i)^{1 - d_i}\,
\end{align}

with $d_i$ the degree of edge $i$. As an example, on the \ffg shown in Fig.~\ref{fig:sample_ffg} this would correspond to a variational distribution of the form

\begin{align}
    q(s_1, \dots, s_4)
    &= \frac{q_a(s_1)q_b(s_1,s_2,s_3)q_c(s_2,s_4)q_d(s_3)q_e(s_4)} {q_1(s_1)q_2(s_2)q_3(s_3)q_4(s_4)}
\end{align}

where we see terms for the edges in the denominator and for the nodes in the numerator. With this definition, the free energy factorises over the \ffg as

\begin{align}
    F[q] = \sum_{a\in\mathcal{V}} \underbrace{\int q_a(\seq{s}_a) \log \frac{q_a(\seq{s}_a)}{f_a(\seq{s}_a)} \d{\seq{s}_a}}_{F[q_a]} + \sum_{i\in\mathcal{E}} (1 - d_i)  \underbrace{\int q_i(s_i) \log \frac{1}{q_i(s_i)} \d{s_i} }_{H[q_i]} \,.\label{eq:bfe}
\end{align}

Eq.~\eqref{eq:bfe} defines the \bfe. Note that $F$ defines a free energy functional which can be either local or global depending on its arguments. More specifically, $F[q]$ defines the free energy for the entire model, while $F[q_a]$ defines a node-local (to the node $a$) \bfe contribution of the same functional form.

Under optimisation of the \bfe - solving Eq.~\eqref{eq:BFE_problem} - the admissible set of functions $\mathcal{Q}$ enforces consistent normalisation and marginalisation of the node and edge local distributions, such that

\begin{subequations} \label{eq:constraints}
\begin{align}
    \int q_i(s_i) \d{s_i} = 1 &\text{ for all } i\in\mathcal{E}\\
    \int q_a(\seq{s}_a) \d{\seq{s}_a} = 1 &\text{ for all } a\in\mathcal{V}\\
    \int q_a(\seq{s}_a) \d{\seq{s}_{a\setminus i}} = q_i(s_i) &\text{ for all } a\in\mathcal{V}, i\in\mathcal{E}(a)\,.
\end{align}
\end{subequations}

Because we always assume the constraints given by Eq.~\eqref{eq:constraints} to be in effect, we will omit the subscript on individual $q$'s moving forwards and instead let the arguments determine which marginal we are referring to, for example writing $q(s_a)$ instead of $q_a(s_a)$.

A core aspect of \ffgs is that they allow for easy visualisation of \mpa algorithms. \Mpa algorithms are a family of distributed inference algorithms with the common trait that they can be viewed as messages flowing on an \ffg.

As a general rule, to infer a marginal for a variable, messages are passed on the graph toward the associated edge for that variable. Multiplication of colliding (forward and backward) messages on an edge yields the desired posterior marginal. We denote a message on an \ffg by arrows pointing in the direction that the message flows.

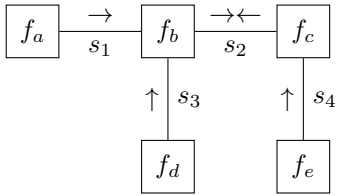
\begin{figure}[H]
    \centering
    \begin{tikzpicture}
[node distance=18mm, auto]

\node[stochastic](fb){$f_b$};
\node[stochastic, right of=fb](fc){$f_c$};

\node[stochastic, left of=fb](fa){$f_a$};

\node[stochastic, below of=fb](fd){$f_d$};

\node[stochastic, below of=fc](fe){$f_e$};

\node[below of=fb](ghost1){};

\draw[](fa)--node[below]{$s_1$}node[above]{$\rightarrow$}(fb);

\draw[](fb)--node[below]{$s_2$}node[above]{$\rightarrow \leftarrow$}(fc);

\draw[](fb)--node[right]{$s_3$}node[left]{$\uparrow$}(fd);
\draw[](fc)--node[right]{$s_4$}node[left]{$\uparrow$}(fe);

\end{tikzpicture}
    \caption{Example of messages flowing on an FFG}
    \label{fig:sample_messages}
\end{figure}

We show an example in Fig.~\ref{fig:sample_messages} of inferring a posterior marginal for the variable $s_2$. In this example, messages are flowing on the \ffg of Fig.~\eqref{fig:sample_ffg} towards the variable $s_2$ where we see two arrows colliding. The exact form of the individual messages depend on the algorithm being used.

\section{Defining epistemic objectives}\label{sec:epistemics}
Now we move on to the central topic of \aif, namely agents that interact with the world they inhabit. Under the heading of \aif, an agent entails a generative model of its environment and is engaged in the process of achieving future goals (specified by a target distribution or goal prior) through actions. This task can be cast as a process of free energy minimisation \cite{friston_free_2019, friston_sophisticated_2021, friston_action_2010}. A natural question to ask is then, what should this free energy functional look like and why? Here we wish to highlight a core feature of \aif that sets it apart from other approaches: Systematic information gathering by targeted exploration of an environment. Whatever the form of our free energy functional, it should lead to agents that possess an exploratory drive consistent with \aif.

While it is tempting to default to a standard \vfe, prior work \cite{schwobel_active_2018} has shown that directly optimising \bfe or \vfe when inferring a sequence of actions (which we will refer to as a \emph{policy}) does not lead to agents that systematically explore their environment. Instead, directly inferring a policy by minimising a \vfe/\bfe leads to \kl-control \cite{kappen_optimal_2012} \footnote{Other works will sometimes use $\pi$ as a symbol to denote a policy}. This means that \vfe/\bfe is not the correct choice when we desire an agent that actively samples its environment with the explicit purpose of gathering information.

Instead a hallmark feature of \aif is the use of alternative functionals in place of either the \vfe or the \bfe, specifically for inferring policies. The goal of these alternative functionals is often specifically to induce an epistemic, explorative term that drives \aif agents to seek out information.

Epistemics, epistemic behaviour or "foraging for information" are commonly used terms in the \aif literature and related fields. While we have used the term colloquially until this point, we now clarify formally how we use the term in the present paper and how it relates to the objective functionals we consider. A core problem is that epistemics is most often defined in terms of the behaviour of agents ("what does my agent \emph{do}?") rather than from a mathematical point of view. Prior work on this point includes \cite{hafner_action_2020,millidge_understanding_2021}.

We take the view that epistemics arise from the optimisation of either an \mi term or a bound thereon. The \mi (between two variables $x$ and $z$) is defined as \cite{mackay_information_2003}

\begin{align}\label{eq:mi}
    \I{x,z} &= \iint p(x,z) \log \frac{p(x,z)}{p(x)p(z)} \dd x \dd z \,.
\end{align}

To gain an intuition for why maximising MI leads to agents that seek out information, we can rewrite MI as

\begin{align}\label{eq:mi_entropy}
\begin{split}
   \I{x,z}
   &= \iint p(x,z) \log \frac{p(z \mid x)}{p(z)} \dd x \dd z \\
   &= \H{z} - \H{z \mid x} = \H{x} - \H{x \mid z}
\end{split}
\end{align}

Since \mi is symmetric in its arguments, Eq.~\eqref{eq:mi_entropy} can equally well be written in terms of $x$ rather than $z$. Eq.~\eqref{eq:mi_entropy} shows that \mi decomposes as the difference between the marginal entropy $z$ and the expected entropy of $z$ conditional on $x$.

If we let $z$ denote an internal state of an agent and $x$ an observation and allow our agent to choose $x$ - for instance through acting on an environment - we can see why maximising Eq.~\eqref{eq:mi} biases the agent towards seeking our observations that reduce entropy in $z$. Maximising Eq.~\eqref{eq:mi} means the agent will prefer observations that provide useful (in the sense of reducing uncertainty) information about its internal states $z$. For this reason \mi is also known as Information Gain.

Actually computing Eq.~\eqref{eq:mi} is often intractable and in practice, a bound is often optimised instead.

\subsection{Constructing a local epistemic objective}
At this point, we have seen that the \bfe is defined over arbitrary \ffgs, yet does not lead to epistemic behaviour. On the other hand, maximising \mi leads to the types of epistemic behaviour we desire, yet is not distributed like the \bfe.
The question becomes whether there is a way to merge the two and obtain a distributed functional - like the \bfe - that includes an epistemic term?

We will now show how to construct such a functional. Our starting point will be the \bfe given by Eq.~\eqref{eq:bfe}. We will focus on a single node $a$ and its associated local energy term and partition the incoming edges into two sets, $\seq{x}$ and $\seq{z}$. This gives the node local free energy

\begin{align}
    F[q_a]
    &= \iint q(\seq{x},\seq{z}) \log \frac{q(\seq{x},\seq{z})}{f(\seq{x},\seq{z})} \dd \seq{x} \dd \seq{z}
\end{align}

Now we need to add on an \mi term to induce epistemics. Since we are minimising our free energy functional and want to maximise \mi, we augment the free energy with a negative \mi term as

\begin{subequations}\label{eq:local_gfe}
\begin{align}
    \GFE{q_a} \label{eq:kl_infogain}
    &= \overbrace{\iint q(\seq{x},\seq{z}) \log \frac{q(\seq{x},\seq{z})}{f(\seq{x},\seq{z})} \dd \seq{x} \dd \seq{z}}^{\text{Variational free energy}}
    + \overbrace{\iint q(\seq{x},\seq{z}) \log \frac{q(\seq{x})q(\seq{z})}{q(\seq{x},\seq{z})} \dd \seq{x} \dd \seq{z}}^{\text{Negative mutual information}} \\
    &= \iint q(\seq{x},\seq{z}) \log \frac{\cancel{q(\seq{x},\seq{z})}}{f(\seq{x},\seq{z})} \dd \seq{x} \dd \seq{z}
    + \iint q(\seq{x},\seq{z}) \log \frac{q(\seq{x})q(\seq{z})}{\cancel{q(\seq{x},\seq{z})}} \dd \seq{x} \dd \seq{z} \\
    &= \iint q(\seq{x},\seq{z}) \log \frac{q(\seq{x})q(\seq{z})}{f(\seq{x},\seq{z})} \dd \seq{x} \dd \seq{z}
\end{align}
\end{subequations}

which we can recognise as a node-local \gfe \cite{parr_generalised_2019}. In section~\ref{sec:gfe} we review the results of \cite{parr_generalised_2019} to expand upon this statement.
Eq.~\eqref{eq:kl_infogain} provides a straightforward explanation of the kinds of behaviour that we can expect out of agents optimising a \gfe. Namely, minimising \bfe with a goal prior corresponds to performing \kl-control \cite{kappen_optimal_2012} while the \mi term adds an epistemic, information-seeking component. Viewed in this way, there is nothing mysterious about the kind of objective optimised by \aif agents: It is simply the sum of two well-known and established objectives that are each widely used within the control and reinforcement learning communities.

\subsection{Generalised free energy}\label{sec:gfe}

The objective derived in Eq.~\eqref{eq:local_gfe} is a node local version of the \gfe originally introduced by \cite{parr_generalised_2019}. In this section, we review the \gfe as constructed by \cite{parr_generalised_2019} in order to relate our construction to prior work on designing \aif functionals. In section~\ref{sec:special_cases}, we show how to reconstruct the exact method of \cite{parr_generalised_2019} using the tools we develop in this paper.
Prior to \cite{parr_generalised_2019}, the functional of choice was the \efe \cite{da_costa_active_2020,friston_active_2015}. \cite{parr_generalised_2019} identified some issues with the \efe and proposed the \gfe as a possible solution.

We show how to reconstruct the original \efe-based algorithm of \cite{friston_active_2015} using our local objective in section~\ref{sec:special_cases}. We also provide a detailed description of the \efe in Appendix.~\ref{sec:efe} since it is still a popular choice for designing \aif agents.

A core issue with the \efe is that it is strictly limited to planning over future timesteps. This means that \aif agents that utilise the \efe functional need to maintain two separate models: One for inferring policies (using \efe) and one for state inference (using \vfe/\bfe) that updates as observations become available. The key advantage of \efe is that it induces the epistemic drive that we desire from \aif agents.

The goal of \cite{parr_generalised_2019} was to extend upon the \efe by introducing a functional that could induce similar epistemic behaviour when used to infer policies while at the same time reducing to a \vfe when dealing with past data points. In this way, an agent would no longer have to maintain two separate models and could instead utilise only one.

The \gfe as introduced by \cite{parr_generalised_2019} is tied to a specific choice of generative model. The model introduced by \cite{parr_generalised_2019} is given by

\begin{align}\label{eq:parr_model}
    p(\vect{x},\vect{z} \mid \vect{\hat{u}}) &\propto
    p(\vect{z}_0) \prod_{k=1}^T p(\vect{x}_k | \vect{z}_k) p(\vect{z}_k | \vect{\hat{u}}_k,\vect{z}_{k-1}) \tilde{p}(\vect{x}_k)
\end{align}

where $\vect{x}$ denotes observations, $\vect{z}$ denotes latent states and $\vect{\hat{u}}$ denotes a fixed policy. Throughout this paper we will use $t$ to refer to the current time step in a given model and denote fixed values by a hat, here exemplified by $\vect{\hat{u}}$. Further, we also use $\vect{z}$ to denote vectors.

Given a current time step $t$, we have that for future time steps $k > t$, $\tilde{p}(\vect{x}_k)$ defines a goal prior over desired future observations. For past time steps $k \leq t$, we instead have $\tilde{p}(\vect{x}_k) = 1$ which makes it uninformative \footnote{\cite{parr_generalised_2019} writes this as the prior being flat to achieve a similar effect}.
Writing the model in this way allows for a single model for both perception (integrating past data points) and action (inferring policies) since both past and future time steps are included. \gfe is defined as \cite{parr_generalised_2019}

\begin{align}\label{eq:gfe}
    \GFE{q ; \hat{\vect{u}}}
    &= \sum_{k=1}^T \iint q(\vect{x}_k | \vect{z}_k) q(\vect{z}_k | \vect{\hat{u}}_k) \log \frac{q(\vect{x}_k | \vect{\hat{u}}_k) q(\vect{z}_k | \vect{\hat{u}}_k)}{\tilde{p}(\vect{x}_k) p(\vect{x}_k,\vect{z}_k| \vect{\hat{u}}_k)} \dd \vect{x}_k \dd \vect{z}_k
\end{align}

where

\begin{align}\label{eq:gfe_assumption}
    q(\vect{x}_k|\vect{z}_k) &=
    \begin{cases}
    \delta(\vect{x}_k - \vect{\hat{x}}_k) & \text{if } k \leq t \\
    p(\vect{x}_k | \vect{z}_k) & \text{if } k > t
    \end{cases} \,.
\end{align}

and $\vect{\hat{x}}_k$ denotes the observed data point at time step $k$. $p(\vect{x}_k,\vect{z}_k \mid \vect{\hat{u}}_k)$ can be found recursively by Bayesian smoothing, see \cite{koudahl_2021_lgds, sarkka_bayesian_2013} for details. To see how the \gfe reduces to a \vfe when data is available, we refer to Appendix~\ref{sec:vfe_and_gfe}.

The \gfe introduced by \cite{parr_generalised_2019} improves upon prior work utilising \efe but still has some issues. The first lies in tying it to the model definition in Eq.~\eqref{eq:parr_model}. Being committed to a model specification apriori severely limits what the \gfe can be applied to since not all problems are going to fit the model specification. Additionally, a more subtle issue lies in the commitment to a hard time horizon. If $t$ denotes the current timestep, at some point time will advance to the point $t > T$. When this happens Eq.~\eqref{eq:parr_model} loses the capacity to plan and becomes static since all observations are clamped by Eq.~\eqref{eq:gfe_assumption}. A further complication arises from $\hat{\vect{u}}$ being a fixed parameter and not a random variable. Being a fixed parameter means it is not possible to perform inference for $\hat{\vect{u}}$ which in turn makes scaling difficult. Moving to a fully local version of the \gfe instead means we can construct a new synthetic approach to \aif that addresses these issues.

\section{LAIF - Lagrangian Active Inference}
Armed with the node local \gfe derived in Eq.~\eqref{eq:local_gfe}, we can construct a Lagrangian for \aif. The goal is to adapt the Lagrangian approach to \mpa sketched in section~\ref{sec:lagrangian} to derive a \mpa algorithm that optimises local \gfe in order to have a distributed inference procedure that incorporates epistemic terms. Naively applying the method of \cite{senoz_variational_2021} to Eq.~\eqref{eq:local_gfe} does not yield useful results because the numerator differs from the term we take the expectation with respect to. To obtain a useful solution we need that

\begin{align}
   q(\seq{x},\seq{z}) &= q(\seq{x} \mid \seq{z}) q(\seq{z})
\end{align}

and add the original assumption in \cite{parr_generalised_2019}, given by Eq.~\eqref{eq:gfe_assumption}

\begin{align}
   q(\seq{x} \mid \seq{z}) &\triangleq p(\seq{x} \mid \seq{z}) \,.
\end{align}

With these additional assumptions, we can obtain meaningful solutions and derive a message passing algorithm that optimises a local \gfe and induces epistemic behaviour which we demonstrate in section\ref{sec:experiments}. The detailed derivation of these results can be found in our companion paper \cite{vandelaar2023realising}. For practical \aif modelling, we provide the relevant messages in Fig.~\ref{fig:messages}.

At this point, we will instead show a different way to arrive at the local \gfe. This approach is more "mechanical" but has the advantage that it can more easily be written in terms of constraints on the \bfe. Our starting point will once again be a free energy term local to the node $a$

\begin{align}\label{eq:local_energy}
    F[q_a] &= \int q(\seq{s}_a) \log \frac{q(\seq{s}_a)}{f(\seq{s}_a)} \dd \seq{s}_a
\end{align}

To turn Eq.~\eqref{eq:local_energy} into a local \gfe we need to perform two steps. The first is to enforce a mean field factorisation

\begin{align}
    q(\seq{s}_a) = \prod_{i \in \mathcal{E}(a)} q(s_i)
\end{align}

The second is to change the expectation to obtain

\begin{align}\label{eq:p_sub_local_gfe}
    \int q(\seq{s}_a) \log \frac{q(\seq{s}_a)}{f(\seq{s}_a)} \dd \seq{s}_a \implies \int p(s_{i} | \seq{s}_{ a\backslash i}) q(\seq{s}_{a\backslash i}) \log \frac{q(\seq{s}_a)}{f(\seq{s}_a)} \dd \seq{s}_a
\end{align}

where $i \in \mathcal{E}(a)$. This partitions the connected variables into two sets: A set of variables where the expectation is modified and any remainder that is not modified. $p(s_{i} \mid \seq{s}_{a \backslash i})$ denotes a conditional probability distribution. We write $p$ here instead of $f$ to emphasise that the conditional needs to be normalised in order for us to be able to take the expectation.

We refer to this move as a \emph{P-substitution}. Once the mean field factorisation is enforced, we can recognise Eq.~\eqref{eq:p_sub_local_gfe} as a node-local \gfe $G[q_a]$.

As an example of constructing a node-local \gfe with this approach, we can consider a node with two connected variables, $\{x, z\}$.

The local free energy becomes

\begin{align}
   F[q_{a}]
   &= \iint q(x,z) \log \frac{q(x,z)}{p(x,z)} \dd x \dd z
\end{align}

Now we apply a mean field factorisation

\begin{align}
    F[q_{a}]
    &= \iint q(x)q(z) \log \frac{q(x)q(z)}{p(x,z)} \dd x \dd z
\end{align}

And finally, perform P-substitution to obtain a node-local \gfe

\begin{align}
    \iint {\color{red} q(x)} q(z) \log \frac{q(x)q(z)}{p(x,z)} \dd x \dd z &\implies  \iint {\color{red} p(x \mid z) } q(z) \log \frac{q(x)q(z)}{p(x,z)} \dd x \dd z
\end{align}

We denote the set of nodes for which we want to perform P-substitution with $\mathcal{P} \subseteq \mathcal{V}$. When performing P-substitution, we are replacing a local \emph{variational} free energy $F[q_a]$ with a local \emph{generalised} free energy $\GFE{q_a}$. Armed with P-substitution, we can now write the simplest instance of a Lagrangian for Active Inference

\begin{align}\label{eq:aif_Lagrangian}
\begin{split}
    \mathcal{L}[q]
    &= \underbrace{\sum_{a \in \mathcal{P}} \GFE{q_a}}_{ \substack{\text{P-substituted subgraph} \\ \text{w / naive mean field} } }
    + \underbrace{\sum_{b \in \mathcal{V} \backslash \mathcal{P}} F[q_b]}_{\substack{\text{Node local} \\ \text{free energies}}}
    + \sum_{i \in \mathcal{E}} (1- d_i) \underbrace{\H{q(s_i)}}_{\text{Edge entropy}}
    \\
    &\quad \quad + \underbrace{\sum_{a \in \mathcal{V}} \sum_{i \in \mathcal{E}} \int \lambda_{ia}(s_i) \Bigg[ q(s_i) - \int q(\seq{s}_a) \dd s_{a \backslash i} \Bigg] \dd s_i}_{\text{Marginalisation}} \\
    & \quad \quad + \underbrace{\sum_{a \in \mathcal{V}} \lambda_a \Bigg[ \int q(\seq{s}_a) \dd \seq{s}_a - 1 \Bigg]}_{\text{Normalisation of node marginals}}
    + \underbrace{\sum_{i \in \mathcal{E}} \lambda_i \Bigg[ \int q(s_i) \dd s_i - 1 \Bigg] }_{\text{Normalisation of edge marginals}} \,.
\end{split}
\end{align}

With the Active Inference Lagrangian in hand, we can now solve for stationary points using variational calculus and obtain \mpa algorithms for \laif. The key insight is that messages flowing out of $\mathcal{P}$ derived from stationary points of Eq.~\eqref{eq:aif_Lagrangian} will correspond to stationary points of the local \gfe rather than \bfe, meaning the result will include an epistemic component. This paves the way for a localised version of \aif that applies to arbitrary graph structures, does not suffer from scaling issues as the planning horizon increases, and can be solved efficiently and asynchronously using \mpa.

\section{Constrained Forney-style factor graphs}\label{sec:CFFG}
While \ffgs are a useful tool for writing down generative models, we have by now established the importance of knowing the exact functional to be minimised. This requires specifying not just the model $f$ but also the family $\mathcal{Q}$ through constraints and any potential P-substitutions. This is important if we want to be able to succinctly specify not just the model but also the exact inference problem we aim to solve.

We will now develop just such a new notation for writing constraints directly as part of the \ffg. We refer to \ffgs with added constraints specification as \Cffgs.

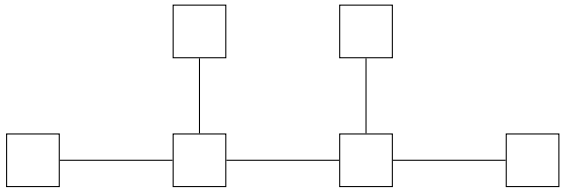
\begin{figure}[H]
    \centering
    \begin{tikzpicture}


\node[stochastic](x){};
\node[stochastic,right= 15mm of x](xyz){};
\node[stochastic,above= 10mm of xyz](y){};
\node[stochastic,right= 15mm of xyz](zvw){};
\node[stochastic,right= 15mm of zvw](w){};
\node[stochastic,above= 10mm of zvw](v){};


\draw (x)--(xyz);
\draw (y)--(xyz);
\draw (xyz)--(zvw);
\draw (v)--(zvw);
\draw (w)--(zvw);

\end{tikzpicture}
    \caption{An example FFG.}
    \label{fig:notation_1}
\end{figure}

Fig~\ref{fig:notation_1} shows a model comprised of five edges and six nodes. \Ffgs traditionally represent the model $f$ using squares connected by lines as shown in Fig.~\ref{fig:notation_1}. The squares represent factors and the connections between them represent variables. Connecting an edge to a square node indicates that the variable on that edge is an argument of the factor it is connected to.

The notation for \cffgs adheres to similar principles when specifying $f$. However, we augment the \ffg with circular beads to indicate the constraints that define our family $\mathcal{Q}$. Each factor of the variational distribution in $q$ will correspond to a bead and the position of a bead indicates to which marginal it refers - a bead on an edge denotes an edge marginal $q(s_i)$ and a bead inside a node denotes a node marginal $q(\seq{s}_a)$. An empty bead will denote the default normalisation constraints while a connection between beads indicates marginalisation constraints following Eq.~\eqref{eq:constraints}. These beads form the basic building blocks of our notation.

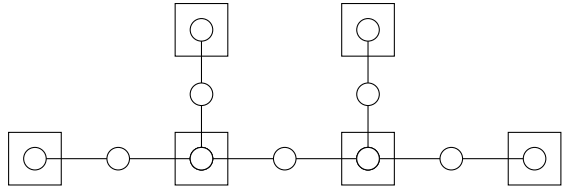
\begin{figure}[H]
    \centering
    \begin{tikzpicture}


\node[stochastic](x){};
\node[stochastic,right= 15mm of x](xyz){};
\node[stochastic,above= 10mm of xyz](y){};
\node[stochastic,right= 15mm of xyz](zvw){};
\node[stochastic,right= 15mm of zvw](w){};
\node[stochastic,above= 10mm of zvw](v){};

\draw (x.center)--node[q,pos=1.0]{}node[q,pos=0.5]{}node[q,pos=0.0]{}(xyz.center);
\draw (y.center)--node[q,pos=1.0]{}node[q,pos=0.5]{}node[q,pos=0.0]{}(xyz.center);
\draw (xyz.center)--node[q,pos=1.0]{}node[q,pos=0.5]{}node[q,pos=0.0]{}(zvw.center);
\draw (v.center)--node[q,pos=1.0]{}node[q,pos=0.5]{}node[q,pos=0.0]{}(zvw.center);
\draw (w.center)--node[q,pos=1.0]{}node[q,pos=0.5]{}node[q,pos=0.0]{}(zvw.center);

\end{tikzpicture}
    \caption{Example \cffg with normalisation and marginalisation constraints.}
    \label{fig:notation_2}
\end{figure}

To write the objective corresponding to the model in Fig.~\ref{fig:notation_1} given the default constraints of Eq.~\eqref{eq:constraints}, we add beads for every term and extend edges through the node boundary to connect variables that are under marginalisation constraints as shown in Fig.~\ref{fig:notation_2}.

\subsection{Factorisation constraints}
We will now extend our notation with the most common types of constraints used for defining $\mathcal{Q}$. A common choice is factorisation of the variational distribution with the most well-known example being the naive mean field approximation. Under a naive mean-field factorisation, all marginals are considered independent. Formally this means we enforce

\begin{align}
    q(\seq{s}_a) &= \prod_{i \in \mathcal{E}(a)} q(s_i) \,.
    \label{eq:mf_constraint}
\end{align}

To write Eq.~\eqref{eq:mf_constraint} on a \cffg we need to replace the joint node marginal with the product of adjacent edge marginals.

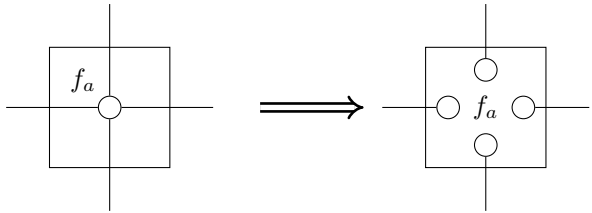
\begin{figure}[H]
    \centering
    \begin{tikzpicture}

\node[bigbox](f){$f_a$};
\node[q,above of=f,node distance=5mm](qnorth){};
\node[q,left of=f, node distance=5mm](qwest){};
\node[q,below of=f,node distance=5mm](qsouth){};
\node[q,right of=f,node distance=5mm](qeast){};

\node[above of=qnorth](north){};
\node[left of=qwest](west){};
\node[right of=qeast](east){};
\node[below of=qsouth](south){};

\draw(qnorth)--(north);
\draw(qsouth)--(south);
\draw(qwest)--(west);
\draw(qeast)--(east);

\node[bigbox, left of=f, node distance=5cm](f2){};
\node[q,left of=f, node distance=5cm](q2){};

\node[above left of=f2, node distance=5mm](fb){$f_a$};
\node[above of=f2,node distance=5mm](qnorth2){};
\node[left of=f2, node distance=5mm](qwest){};
\node[below of=f2,node distance=5mm](qsouth2){};
\node[right of=f2,node distance=5mm](qeast2){};

\node[above of=qnorth2](north2){};
\node[left of=qwest](west2){};
\node[right of=qeast2](east2){};
\node[below of=qsouth2](south2){};

\draw(north2)--(q2);
\draw(south2)--(q2);
\draw(west2)--(q2);
\draw(east2)--(q2);

\draw[-Implies,line width=1,double distance=2pt](east2)++(0.5,0)--(west);

\end{tikzpicture}
    \caption{Changing a local joint factorisation to a naive mean field assumption on a \cffg.}
    \label{fig:mean_field}
\end{figure}

To do this we can replace the bead indicating the joint marginal with a bead for each edge marginal in Eq.~\eqref{eq:mf_constraint} as shown in Fig.~\ref{fig:mean_field}.

The naive mean field is the strongest factorisation possible. It is possible to utilise less aggressive factorisations by appealing to a structured mean field approximation instead. The structured mean field constraint takes the form

\begin{align}
    q(\seq{s}_a) = \prod_{n \in l(a)} q^n(\seq{s}_a^n)
\end{align}

where $l(a)$ denotes a set of one or more edges connected to the node $a$ such that each element in $\mathcal{E}(a)$ can only appear in $l(a)$ once \cite{senoz_variational_2021}. For example if $\mathcal{E}(a) = \{i,j,k\}$ corresponding to variables $\{x, y ,z\}$, we can factorise $q(x,y,z)$ as $q(x)q(y,z)$ or $q(z)q(x,y)$ but not as $q(x,y)q(y,z)$ since $y$ appears twice. The naive mean field is a special case of the structured mean field where every variable appears only by itself.

To write a structured mean field factorisation on a \cffg we can apply a similar logic and replace the single bead denoting the joint with beads that match the structure of $l(a)$. Each set of variables that are factorised together corresponds to a single bead connected to the edges in the set that factor together.

\begin{figure}[H]
    \centering
    \begin{tikzpicture}

\node[bigbox, node distance=5cm](f){};
\node[q](q){};

\node[above left of=f, node distance=5mm](fb){$f_a$};
\node[above of=f,node distance=5mm](qnorth){};
\node[left of=f, node distance=5mm](qwest){};
\node[below of=f,node distance=5mm](qsouth){};
\node[right of=f,node distance=5mm](qeast){};

\node[above of=qnorth](north){};
\node[left of=qwest](west){};
\node[right of=qeast](east){};
\node[below of=qsouth](south){};

\draw(north)--(q);
\draw(south)--(q);
\draw(west)--(q);
\draw(east)--(q);

\node[right of=f, node distance=5cm](ghost1){};

\node[bigbox, above of=ghost1, node distance=2cm](f2){$f_a$};

\bigarcnw{f2};
\node[q,above left of=f2, node distance=5mm](qxy){};
\bigarcse{f2};
\node[q,below right of=f2, node distance=5mm](qzv){};

\node[above of=f2,node distance=5mm](qnorth2){};
\node[left of=f2, node distance=5mm](qwest2){};
\node[below of=f2,node distance=5mm](qsouth2){};
\node[right of=f2,node distance=5mm](qeast2){};
\node[above of=qnorth2](north2){};
\node[left of=qwest2](west2){};
\node[right of=qeast2](east2){};
\node[below of=qsouth2](south2){};
\draw(qnorth2)--(north2);
\draw(qsouth2)--(south2);
\draw(qwest2)--(west2);
\draw(qeast2)--(east2);

\node[bigbox, below of=ghost1, node distance=2cm](f3){$f_a$};

\bigarcsw{f3};
\node[q,below left of=f3, node distance=5mm](qyz){};

\node[q,above of=f3,node distance=5mm](qnorth3){};
\node[left of=f3, node distance=5mm](qwest3){};
\node[below of=f3,node distance=5mm](qsouth3){};
\node[q,right of=f3,node distance=5mm](qeast3){};
\node[above of=qnorth3](north3){};
\node[left of=qwest3](west3){};
\node[right of=qeast3](east3){};
\node[below of=qsouth3](south3){};
\draw(qnorth3)--(north3);
\draw(qsouth3)--(south3);
\draw(qwest3)--(west3);
\draw(qeast3)--(east3);

\draw[-Implies,line width=1,double distance=2pt](f.east)++(0.4,0.7)--++(2,1);
\draw[-Implies,line width=1,double distance=2pt](f.east)++(0.4,-0.7)--++(2,-1);

\end{tikzpicture}
    \caption{Changing a local joint factorisation to structured mean field on a \cffg.
    }
    \label{fig:structured_mean_field}
\end{figure}
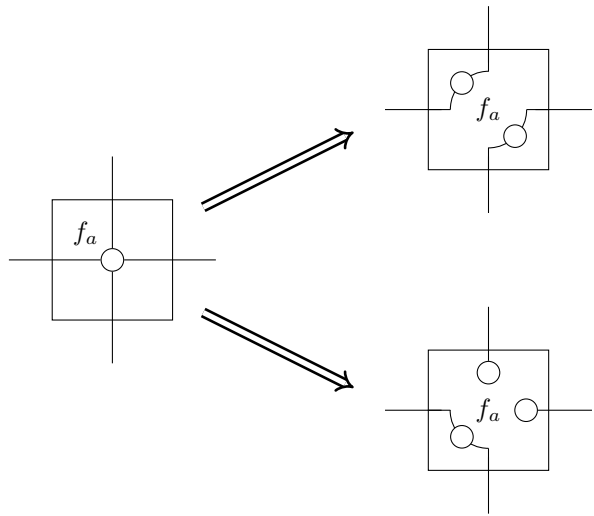

Fig.~\ref{fig:structured_mean_field} shows two example factorisations. The first option factorises the four incoming edges into two sets of two while the second partitions the incoming edges into two sets of one and a single set of two. Using these principles it is possible to specify complex factorisation constraints as part of the \cffg by augmenting each node on the original \ffg.

\begin{figure}[H]
    \centering
    \begin{tikzpicture}

\node[bigbox](f){$f_a$};
\node[q,left of=f, node distance=4mm, yshift=5mm](qup){};
\node[q,left of=f, node distance=4mm, yshift=-5mm](qdown){};

\node[left of=qup](up){};
\node[left of=qdown](down){};

\draw(qdown)--node[pos=0.5](lanchor1){}(down);
\draw(qup)--node[pos=0.5](lanchor2){}(up);

\node[bigbox, right of=f, node distance=45mm](f2){$f_a$};
\node[left of=f2, node distance=8mm, yshift=5mm](qup2){};
\node[left of=f2, node distance=8mm, yshift=-5mm](qdown2){};

\node[left of=qup2](up2){};
\node[left of=qdown2](down2){};

\draw[add=-2.1mm and 1.1mm](up2) to node[pos=0.5](anchor1){} (qup2);
\draw[add=-2.1mm and 1.1mm](down2) to node[pos=0.5](anchor2){} (qdown2);
\draw(qup2) to[out=0, in=0] node[q,pos=0.5](){} (qdown2);

\draw[dotted, thick](anchor1)--(anchor2);
\draw[dotted, thick](lanchor1)--(lanchor2);

\end{tikzpicture}
    \caption{Mean field (left) and joint (right) factorisation constraints for variable numbers of edges on a \cffg.}
    \label{fig:variable_edges}
\end{figure}
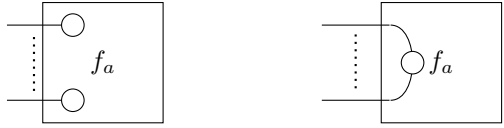

The final situation we need to consider is the case when a single node has a variable or very high number of incoming edges. An example could be a Gaussian Mixture Model with a variable number of mixture components. On a \cffg we indicate variable or large numbers of identical edges by drawing two of the relevant edges and separating them with dots ($\cdots$) as shown in Fig.~\ref{fig:variable_edges}. To indicate factorisation constraints we can write either a joint or a naive mean field factorisation between the two edges, letting the dots denote that a similar factorisation applies to the remaining edges.

\subsection{Form constraints}

We will now extend \cffg notation with constraints on the functional form of nodes and edges. Form constraints are used to enforce a particular form for a local marginal on either an edge or a node. For an edge $s_i$ they enforce

\begin{align}\label{eq:form_constraint}
   \int q(\seq{s}_a) \dd s_{a \backslash i} &= q(s_i) = g(s_i)
\end{align}

where $g(s_i)$ denotes the functional form we are constraining the edge marginal $s_i$ to take. Form constraints on node marginals take the form

\begin{align}
    q(\seq{s}_a) &= g(\seq{s}_a)
\end{align}

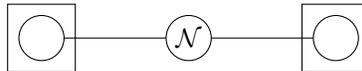
\begin{figure}[H]
    \centering
    \begin{tikzpicture}

\node[bigstoc](left){};
\node[bigstoc,right=30mm of left](right){};

\path[line] (left.center) edge[-] node[Q,pos=0.0](){} node[Q, pos=0.5]{$\mathcal{N}$} node[Q,pos=1.0](){} (right.center);

\end{tikzpicture}
    \caption{Notation for enforcing a Gaussian form constraint on an edge.}
    \label{fig:edge_1}
\end{figure}

Conventionally \ffgs denote the form of a factor by a symbol inside the node. We adopt a similar convention to denote form constraints on $q$ by adding symbols within the corresponding beads. For instance, we can indicate a Gaussian form constraint on an edge as shown in Fig~\ref{fig:edge_1}

Note that this is not dependent on the form of the neighbouring factors. This is a subtle point as it allows us to write approximations into the specification of $\mathcal{Q}$. As an example, the unconstrained marginal in Fig~\ref{fig:edge_1} might be bimodal or highly skewed but by adding a form constraint, we are enforcing a Gaussian approximation. Outside of a few special cases, enforcing form constraints on edges is rarely done in practice since the functional form of $q$ most often follows from optimisation \cite{senoz_variational_2021}.

A special case of form constraints is the case of dangling edges (edges that are not terminated by a factor node). Technically these would not warrant a bead since they would not appear explicitly in the \bfe due to having degree $1$. Intuitively this means that the edge marginal is only counted once and we therefore do not need to correct for overcounting. However, without a bead, there is nowhere to annotate a form constraint which is problematic.

The solution for \cffg notation is to simply draw the bead anyway, in case a form constraint is needed. This is formally equivalent to terminating the dangling edge by a factor node with the node function $f_a(\seq{s}_a) = 1$. Terminating the edge in this way means the edge in question now has degree $2$ and therefore warrants a bead. This is always a valid move since multiplication by $1$ does not change the underlying function \cite{senoz_variational_2021}.

\begin{figure}[H]
    \centering
    \begin{tikzpicture}

\node[bigstoc](left){};
\node[bigstoc,right=30mm of left](right){};

\path[line] (left.center) edge[-] node[Q,pos=0.0](){$\mathcal{N}$} node[Q, pos=0.5]{} node[Q,pos=1.0](){$\mathcal{W}$} (right.center);

\end{tikzpicture}
    \caption{Notation for enforcing form constraints on nodes.}
    \label{fig:edge_2}
\end{figure}
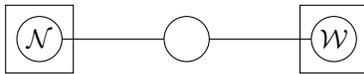

We can denote form constraints on node marginals in the same manner as edge marginals. We show an example in Fig~\ref{fig:edge_2} where we enforce a Gaussian form constraint on one node marginal and a Wishart on the other. Again it is important to note that these are constraints on $q$ and not part of the underlying model specification $f$.

Two kinds of form constraints warrant extra attention: $\delta$-constraints and moment matching. We will now deal with these in turn.

\subsection{$\delta$-constraints and data points}

$\delta$-constraints are the most commonly used form constraints because they allow us to incorporate data points into a model. A $\delta$-constraint on an edge defines the function $g(s_i)$ in Eq.~\eqref{eq:form_constraint} to be

\begin{align}
    g(s_i) &= \delta(s_i - \hat{s}_i)
\end{align}

\begin{figure}[H]
    \centering
    \begin{tikzpicture}

\node[bigstoc](left){};
\node[bc,right=15mm of left](right){$\delta$};

\path[line] (left.center) node[Q, pos=0.0]{} edge[-]  (right);

\node[bigstoc, right=15mm of right](left2){};
\node[bigstoc,right=30mm of left2](right2){};

\path[line] (left.center) node[Q, pos=0.0]{} edge[-]  (right);

\path[line] (left2.center) edge[-] node[Q, pos=0.0]{} node[bc, pos=0.5]{$\delta$} node[Q, pos=1.0]{} (right2.center);

\end{tikzpicture}
    \caption{Terminating and non-terminating notation for data constraints.}
    \label{fig:edge_3}
\end{figure}
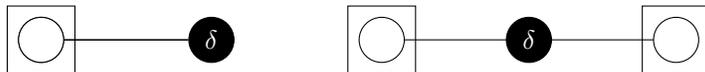

What makes the $\delta$-constraint special is that $\hat{s}_i$ can either be a known value or a parameter to optimise \cite{senoz_variational_2021}. In the case where $\hat{s}_i$ is known, it commonly corresponds to a data point. We will refer to this case as a data constraint and denote it with a filled circle as shown in Fig~\ref{fig:edge_3}

Data constraints are special because they denote observations. They also block any information flow across the edge in question \cite{senoz_variational_2021}. Because they block information flow, \cffg notation optionally allows data constraints to terminate edges.

Here we wish to raise a subtle point about prior \ffg notation. Previous work has used small black \emph{squares} to denote data constraints following \cite{reller2013state}. In keeping with our convention, a small black square on a \cffg denotes a $\delta$-distributed variable in the model $f$ rather than the variational distribution $q$. Being able to differentiate data constrained variables in $q$ and apriori fixed parameter of the model $f$ allows us to be explicit about what actually constitutes a data point for the inference problem at hand \cite{caticha_entropic_2012}.

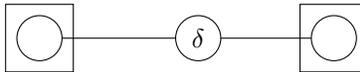
\begin{figure}[H]
    \centering
    \begin{tikzpicture}

\node[bigstoc](left){};
\node[bigstoc,right=30mm of left](right){};

\path[line] (left.center)  node[Q, pos=0.0]{} edge[-] node[wc, pos=0.5](){$\delta$} node[Q, pos=1.0]{}(right.center);

\end{tikzpicture}
    \caption{Notation for $\delta$-constraints.}
    \label{fig:edge_4}
\end{figure}

In the case where $\hat{s}_j$ is not known, it can be treated as a parameter to be optimised. We refer to this case as a $\delta$-constraint or a pointmass constraint and notate it with an unfilled circle as shown in Fig~\ref{fig:edge_4}

Unlike data constraints, the $\delta$-constraint allows messages to pass and is therefore not allowed to terminate an edge. Optimising the value of $\hat{s}_i$ under a $\delta$-constraint leads to EM as message passing \cite{senoz_variational_2021}.

\subsection{Moment matching constraints}
Moment matching constraints are special in that they replace the hard marginalisation constraints of Eq.~\eqref{eq:bfe} with constraints of the form

\begin{align}
    q(s_i) = \int q(\seq{s}_a) \dd s_{a \backslash i} & \implies  \int q(\seq{s}_a) \vect{T}_i(s_i) \dd \seq{s}_a = \int q(s_i) \vect{T}_i(s_i)\dd s_i
\end{align}

where $\vect{T}_i(s_i)$  are the sufficient statistics of an exponential family distribution. This move loosens the marginalisation constraint by instead only requiring that the moments in question align. When taking the first variation and solving, one obtains the \acf{EP} algorithm \cite{minka_expectation_2001}.

For notational purposes, moment matching constraints are unique in that they involve both an edge- and a node-marginal. That means the effects are not localised to a single bead. To indicate which beads are involved, we replace the solid lines between them with dashed lines instead.

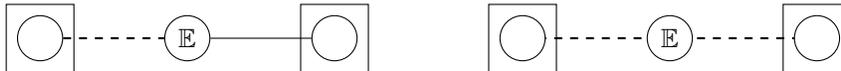
\begin{figure}[H]
    \centering
    \begin{tikzpicture}

\node[bigstoc](left){};
\node[bigstoc, right=30mm of left](right){};
\node[Q] (q1) at ($(left)!0.5!(right)$) {$\mathbb{E}$};
\node[Q] at (left) (qleft1){};
\node[Q] at (right) (qright1){};

\draw[thick, dashed](qleft1)--(q1);
\draw[](q1)--(qright1);


\node[bigstoc, right of=right, node distance=25mm](left2){};
\node[bigstoc, right=30mm of left2](right2){};
\draw[dashed,thick] (left2.center)--node[Q, solid, thin, pos=0.0](){} node[Q, solid, thin, pos=0.5](){$\mathbb{E}$} node[Q, solid, thin, pos=1.0](){} (right2.center);

\end{tikzpicture}
    \caption{Notation for moment matching with a single-sided (left) and double-sided (right) node/edge pairs.}
    \label{fig:ep_edge}
\end{figure}

We denote moment matching constraints by an $\mathbb{E}$ inside the corresponding edge-bead as shown in Fig~\ref{fig:ep_edge}. Choosing the edge-bead over the node-bead is an arbitrary decision made mainly for convenience.

The left side of Fig.~\ref{fig:ep_edge} shows notation for constraining a single node/edge pair by moment matching. If both nodes connected to an edge are under moment matching constraints, the double-sided notation on the right of Fig.~\ref{fig:ep_edge} applies.

Given the modular nature of \cffg notation it is easy to compose different local constraints to accurately specify a Lagrangian and by extension an inference problem. Adding custom marginal constraints to a \cffg is also straightforward as it simply requires defining the meaning of a symbol inside a bead.

\subsection{P-substitution on CFFGs}
The final piece needed to represent the Active Inference Lagrangian on a \cffg is P-substitution. Being able to represent \laif on a \cffg is the reason for constructing the local \gfe using a mean-field factorisation and P-substitution. This construction is much more amenable to the tools we have developed so far as we will now demonstrate.

\begin{figure}[H]
    \centering
    \begin{tikzpicture}
\node[bigbox](f){$f_a$};
\node[q,above of=f,node distance=5mm](qnorth){};
\node[q,left of=f, node distance=5mm](qwest){};
\node[q,below of=f,node distance=5mm](qsouth){};

\node[above of=qnorth](north){};
\node[left of=qwest](west){};
\node[below of=qsouth](south){};

\draw(qnorth)--node[left](){$x$}(north);
\draw(qsouth)--node[left](){$y$}(south);
\draw(qwest)--node[above](){$z$}(west);

\node[bigbox, right of=f, node distance=4.5cm](f2){$f_a$};

\node[q,above of=f2,node distance=5mm](qnorth2){};
\node[q,left of=f2, node distance=5mm](qwest2){};
\node[p,below of=f2,node distance=5mm](qsouth2){};

\node[above of=qnorth2](north2){};
\node[left of=qwest2](west2){};
\node[right of=qeast2](east2){};
\node[below of=qsouth2](south2){};

\draw(qnorth2)--node[left](){$x$}(north2);
\draw(qsouth2)--node[left](){$y$}(south2);
\draw(qwest2)--node[above](){$z$}(west2);

\draw[-Implies,line width=1,double distance=2pt](east)++(-0.5,0)--(west2);

\end{tikzpicture}
    \caption{P-substitution on a \cffg with naive mean field factorisation.}
    \label{fig:simple_epistemic}
\end{figure}
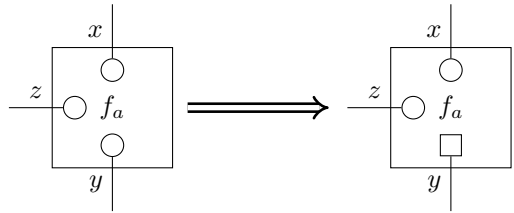

Recall that P-substitution involves substituting part of the model $p$ for $q$ in the expectation only. To write P-substitution on a \cffg, the logical notation is therefore to replace a circle with a square. Fig.~\ref{fig:simple_epistemic} shows an example of adding a P-substitution to a mean-field factorised node marginal

The square notation for P-substitution on \cffgs implies a conditioning of the P-substituted variable on all other connected variables that are not P-substituted. For example, in Fig.~\ref{fig:simple_epistemic} the P-substitution changes local \vfe to a \gfe by

\begin{align}
\begin{split}
    &\iiint {\color{red} q(y)}q(x)q(z) \log \frac{q(y)q(x)q(z)}{p(y,x,z)} \dd y \dd x \dd z \\
    &\quad \quad \implies \iiint {\color{red} p(y \mid x,z)} q(x)q(z) \log \frac{q(y)q(x)q(z)}{p(y,x,z)} \dd y \dd x \dd z
\end{split}
\end{align}

Here, the P-substituted variable is $y$, and the remainder are $\{x,z\}$.

\subsection{CFFG Compression}\label{sec:compressed}

The value of \cffg notation is measured by how much it aids other researchers and practitioners in expressing their ideas accurately and succinctly. We envision two main groups for whom \cffgs might be of particular interest. The first group is comprised of mathematical researchers working on constrained free energy optimisation on \ffgs For this group we expect that the notation developed so far will be both useful and practically applicable since work is often focused on the intricacies of performing local optimisation. Commonly an \ffg in this tradition is small but a very high level of accuracy is desired in order to be mathematically rigorous.

However there is a second group composed of applied researchers for whom the challenge is to accurately specify a larger inference problem and its solution to solve an auxiliary goal - for instance controlling a drone, transmitting a coded message or simulating some phenomenon using AIF. For this group of researchers in particular, \cffg notation as described so far might be too verbose and the overhead of using it may not outweigh the benefits gained. To this end, we now complete \cffg notation by a mandatory compression step.

The compression step is designed to remove redundant information by enforcing an emphasis on \emph{deviations from a default \bfe}. By default, we mean no constraints other than normalisation and marginalisation and with a joint factorisation around every node. Recall that default normalisation constraints are denoted by empty, round beads and marginalisation by connected lines. A joint factorisation means all incoming edges are connected and the node only has a single, internal bead.

To provide a recipe for compressing a \cffg, we will need the concept of a \emph{bead chain}. A bead chain is simply a series of beads connected by edges. In the following recipe, a bead will only be summarised as part of a chain if it contains no additional information, meaning if it is round and empty. To compress a \cffg, we follow a series of four steps:

\begin{enumerate}
    \item Summarise every bead chain by their terminating beads.
    \item For nodes with no factorisation constraints, remove empty, internal beads.
    \item For nodes with no factorisation constraints, remove all internal edges.
    \item For all factor nodes, push remaining internal beads to the border of the corresponding factor node.
\end{enumerate}

\begin{figure}[H]
    \centering
    \begin{tikzpicture}
[node distance=25mm, auto]

\node[bigbox](fa){$f_a$};
\node[bigbox, right of=fa](fb){$f_b$};

\node[bigbox, left of=fa](fc){$f_c$};

\node[bigbox, below of=fc](fd){$f_d$};

\node[observation, below of=fd](obs1){};

\node[bigbox, below of=fb](fe){$f_e$};

\node[above of=fa, node distance=20mm](ghost1){};
\node[data, right of=fb, node distance=15mm](delta1){$\delta$};

\node[bigbox, right of=fd](eq1){$=$};

\node[bigbox, below of=eq1](ff){};
\node[above left of=ff, node distance=5mm](ff_f){$f_f$};

\node[bigbox, below of=fe](fg){};
\node[above left of=fg, node distance=5mm](fg_f){$f_g$};

\bigarcs{fa};
\bigarcnw{fe};
\bigarcne{fd};
\bigarcse{fc};
\bigarcse{eq1};
\bigarcsw{eq1};

\node[q, above right of=fd, node distance=5mm](fd_q){};
\node[q, below right of=fc, node distance=5mm](fc_q){};
\node[q, below of=eq1, node distance=5mm](eq1_q){};

\draw[](ghost1.center)--node[q,pos=1.0](){} ($(fa.north) - (0.0,0.3)$);

\draw[](fa.east)--node[q,pos=0.35](){}node[q,pos=1.0](){}($(fb.center) - (0.5,0,0)$);
\draw[](delta1)--node[q,pos=1.0](){}($(fb.center) + (0.5,0,0)$);

\draw[densely dashed](fe.north)--node[q,  solid, thin, minimum size=4mm, pos=0.35](E1){$\mathbb{E}$} node[q, solid, thin, pos=1.0](){}($(fb.center) - (0.0,0.5)$);
\draw[](fe.north)--(E1); 

\draw[](fg.center)-- node[q, pos=0.0](){}node[q,pos=0.60](){} node[p,pos=1.0](){} ($(fe.south) + (0.0,0.3)$);

\draw[](fd)--(obs1);

\draw[](fa.west)--node[q,pos=0.5](){}(fc);
\draw[](fc)--node[q,pos=0.5](){$\delta$}(fd);
\draw[](fd.east)--node[q,pos=0.5](){}(eq1.west);
\draw[](eq1.east)--node[q,pos=0.5](){}(fe.west);

\draw[](eq1.south)--node[q,pos=0.3](){}node[q,pos=1.0](){}(ff.center);

\node[q, below of=fa, node distance=5mm](){};
\node[q, above left of=fe, node distance=5mm](){};

\end{tikzpicture}
    \caption{Initial CFFG before compression.}
    \label{fig:shorthand_1}
\end{figure}
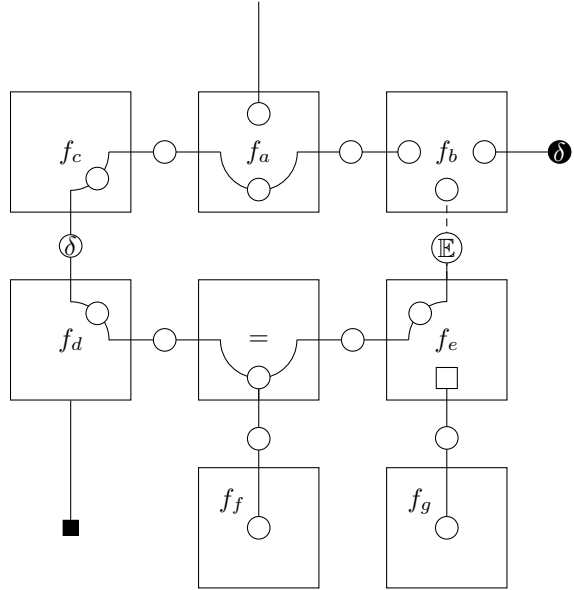

After performing these steps, we are left with a compressed version of the original \cffg where each node can be much smaller and more concise. To exemplify, we apply the recipe to the \cffg in Fig.~\ref{fig:shorthand_1}.

Fig.~\ref{fig:shorthand_1} is a complicated graph with loops, dangling edges, and multiple different factorisations in play. As a result, Fig.~\ref{fig:shorthand_1} covers a lot of special cases that one might encounter on a \cffg. We will now apply the steps in sequence, starting by removing empty beads on chains. This removes most of the beads in the inner loop as shown in Fig.~\ref{fig:shorthand_2}

\begin{figure}[H]
    \centering
    \makebox[0pt][c]{\begin{tikzpicture}
[node distance=25mm, auto]

\node[bigbox](fa){$f_a$};
\node[bigbox, right of=fa](fb){$f_b$};

\node[bigbox, left of=fa](fc){$f_c$};

\node[bigbox, below of=fc](fd){$f_d$};

\node[observation, below of=fd](obs1){};

\node[bigbox, below of=fb](fe){$f_e$};

\node[above of=fa, node distance=20mm](ghost1){};
\node[data, right of=fb, node distance=15mm](delta1){$\delta$};

\node[bigbox, right of=fd](eq1){$=$};

\node[bigbox, below of=eq1](ff){};
\node[above left of=ff, node distance=5mm](ff_f){$f_f$};

\node[bigbox, below of=fe](fg){};
\node[above left of=fg, node distance=5mm](fg_f){$f_g$};

\bigarcs{fa};
\bigarcnw{fe};
\bigarcne{fd};
\bigarcse{fc};
\bigarcse{eq1};
\bigarcsw{eq1};

\draw[](ghost1.center)-- node[q,pos=1.0](){} ($(fa.north) - (0.0,0.3)$);

\draw[](fa.east)--node[q,pos=1.0](){}($(fb.center) - (0.5,0,0)$);
\draw[](delta1)--node[q,pos=1.0](){}($(fb.center) + (0.5,0,0)$);

\draw[densely dashed](fe.north)--node[q,  solid, thin, minimum size=4mm, pos=0.35](E1){$\mathbb{E}$} node[q, solid, thin, pos=1.0](){}($(fb.center) - (0.0,0.5)$);
\draw[](fe.north)--(E1); 

\draw[](fg.center)--node[q,pos=0.0](){}node[p,pos=1.0](){}($(fe.south) + (0.0,0.3)$);

\draw[](fd)--(fd.east);
\draw[](fd)--(obs1);

\draw[](fa.west)--(fc);
\draw[](fc)--node[q,pos=0.5](){$\delta$}(fd);
\draw[](fd.east)--(eq1.west);
\draw[](eq1.east)--(fe.west);

\draw[](eq1.south)--node[q,pos=1.0](){}(ff.center);

\node[bigbox, left of=fa, node distance=90mm](fa2){$f_a$};
\node[bigbox, right of=fa2](fb){$f_b$};

\node[bigbox, left of=fa2](fc){$f_c$};

\node[bigbox, below of=fc](fd){$f_d$};

\node[observation, below of=fd](obs1){};

\node[bigbox, below of=fb](fe){$f_e$};

\node[above of=fa2, node distance=20mm](ghost1){};
\node[data, right of=fb, node distance=15mm](delta1){$\delta$};

\node[bigbox, right of=fd](eq1){$=$};

\node[bigbox, below of=eq1](ff){};
\node[above left of=ff, node distance=5mm](ff_f){$f_f$};

\node[bigbox, below of=fe](fg){};
\node[above left of=fg, node distance=5mm](fg_f){$f_g$};

\bigarcs{fa2};
\bigarcnw{fe};
\bigarcne{fd};
\bigarcse{fc};
\bigarcse{eq1};
\bigarcsw{eq1};
\node[q, above right of=fd, node distance=5mm, color=red](fd_q){};
\node[q, below right of=fc, node distance=5mm, color=red](fc_q){};
\node[q, below of=eq1, node distance=5mm, color=red](eq1_q){};

\draw[](ghost1.center)--node[q,pos=1.0](){} ($(fa2.north) - (0.0,0.3)$);

\draw[](fa2.east)--node[q,pos=0.35, color=red](){}node[q,pos=1.0](){}($(fb.center) - (0.5,0,0)$);
\draw[](delta1)--node[q,pos=1.0](){}($(fb.center) + (0.5,0,0)$);

\draw[densely dashed](fe.north)--node[q,  solid, thin, minimum size=4mm, pos=0.35](E1){$\mathbb{E}$} node[q, solid, thin, pos=1.0](){}($(fb.center) - (0.0,0.5)$);
\draw[](fe.north)--(E1); 

\draw[](fg.center)--node[q,pos=0.0](){}node[q,pos=0.60, color=red](){}node[p,pos=1.0](){}($(fe.south) + (0.0,0.3)$);

\draw[](fd)--(obs1);

\draw[](fa2.west)--node[q,pos=0.5, color=red](){}(fc);
\draw[](fc)--node[q,pos=0.5](){$\delta$}(fd);
\draw[](fd.east)--node[q,pos=0.5, color=red](){}(eq1.west);
\draw[](eq1.east)--node[q,pos=0.5, color=red](){}(fe.west);

\draw[](eq1.south)--node[q,pos=0.3, color=red](){}node[q,pos=1.0](){}(ff.center);

\node[q, below of=fa2, node distance=5mm, color=red](){};
\node[q, above left of=fe, node distance=5mm, color=red](){};

\draw[-Implies,line width=1,double distance=2pt](fe.east)++(0.5,0)--++(1.5,0);

\end{tikzpicture}}
    \caption{Step 1: Removing empty beads from chains. Affected beads are highlighted in {\color{red} red}.
    }
    \label{fig:shorthand_2}
\end{figure}

\begin{figure}[H]
    \centering
    \makebox[0pt][c]{\begin{tikzpicture}
[node distance=25mm, auto]

\node[bigbox](fa){$f_a$};
\node[bigbox, right of=fa](fb){$f_b$};

\node[bigbox, left of=fa](fc){$f_c$};

\node[bigbox, below of=fc](fd){$f_d$};

\node[observation, below of=fd](obs1){};

\node[bigbox, below of=fb](fe){$f_e$};

\node[above of=fa, node distance=20mm](ghost1){};
\node[data, right of=fb, node distance=15mm](delta1){$\delta$};

\node[bigbox, right of=fd](eq1){$=$};

\node[bigbox, below of=eq1](ff){};
\node[above left of=ff, node distance=5mm](ff_f){$f_f$};

\node[bigbox, below of=fe](fg){};
\node[above left of=fg, node distance=5mm](fg_f){$f_g$};

\bigarcs{fa};
\bigarcnw{fe};
\bigarcse{fc};
\bigarcne{fd};
\bigarcse{eq1};
\bigarcsw{eq1};

\draw[](ghost1.center)--node[q,pos=1.0](){}($(fa.north) - (0.0,0.3)$);

\draw[](fa.east)--node[q,pos=1.0](){}($(fb.center) - (0.5,0,0)$);
\draw[](delta1)--node[q,pos=1.0](){}($(fb.center) + (0.5,0,0)$);

\draw[densely dashed](fe.north)--node[q,  solid, thin, minimum size=4mm, pos=0.35](E1){$\mathbb{E}$} node[q, solid, thin, pos=1.0](){}($(fb.center) - (0.0,0.5)$);
\draw[](fe.north)--(E1); 

\draw[](fg.center)--node[p,pos=1.0](){}($(fe.south) + (0.0,0.3)$);

\draw[](fd)--(obs1);

\draw[](fa.west)--(fc);
\draw[](fc)--node[q,pos=0.5](){$\delta$}(fd);
\draw[](fd.east)--(eq1.west);
\draw[](eq1.east)--(fe.west);

\draw[](eq1.south)--(ff.center);

\node[bigbox, left of=fa, node distance=90mm](fa2){$f_a$};
\node[bigbox, right of=fa2](fb){$f_b$};

\node[bigbox, left of=fa2](fc){$f_c$};

\node[bigbox, below of=fc](fd){$f_d$};

\node[observation, below of=fd](obs1){};

\node[bigbox, below of=fb](fe){$f_e$};

\node[above of=fa2, node distance=20mm](ghost1){};
\node[data, right of=fb, node distance=15mm](delta1){$\delta$};

\node[bigbox, right of=fd](eq1){$=$};

\node[bigbox, below of=eq1](ff){};
\node[above left of=ff, node distance=5mm](ff_f){$f_f$};

\node[bigbox, below of=fe](fg){};
\node[above left of=fg, node distance=5mm](fg_f){$f_g$};

\bigarcs{fa2};
\bigarcnw{fe};
\bigarcne{fd};
\bigarcse{fc};
\bigarcse{eq1};
\bigarcsw{eq1};

\draw[](ghost1.center)--node[q,pos=1.0](){}($(fa2.north) - (0.0,0.3)$);

\draw[](fa2.east)--node[q,pos=1.0](){}($(fb.center) - (0.5,0,0)$);
\draw[](delta1)--node[q,pos=1.0](){}($(fb.center) + (0.5,0,0)$);

\draw[densely dashed](fe.north)--node[q,  solid, thin, minimum size=4mm, pos=0.35](E1){$\mathbb{E}$} node[q, solid, thin, pos=1.0](){}($(fb.center) - (0.0,0.5)$);
\draw[](fe.north)--(E1); 

\draw[](fg.center)--node[q,pos=0.0, color=red](){}node[p,pos=1.0](){}($(fe.south) + (0.0,0.3)$);

\draw[](fd)--(fd.east);
\draw[](fd)--(obs1);

\draw[](fa2.west)--(fc);
\draw[](fc)--node[q,pos=0.5](){$\delta$}(fd);
\draw[](fd.east)--(eq1.west);
\draw[](eq1.east)--(fe.west);

\draw[](eq1.south)--node[q,pos=1.0, color=red](){}(ff.center);

\draw[-Implies,line width=1,double distance=2pt](fe.east)++(0.5,0)--++(1.5,0);

\end{tikzpicture}}
    \caption{Step 2. Removing beads on nodes with default factorisation. Affected nodes are highlighted in {\color{red} red}.}
    \label{fig:shorthand_3}
\end{figure}
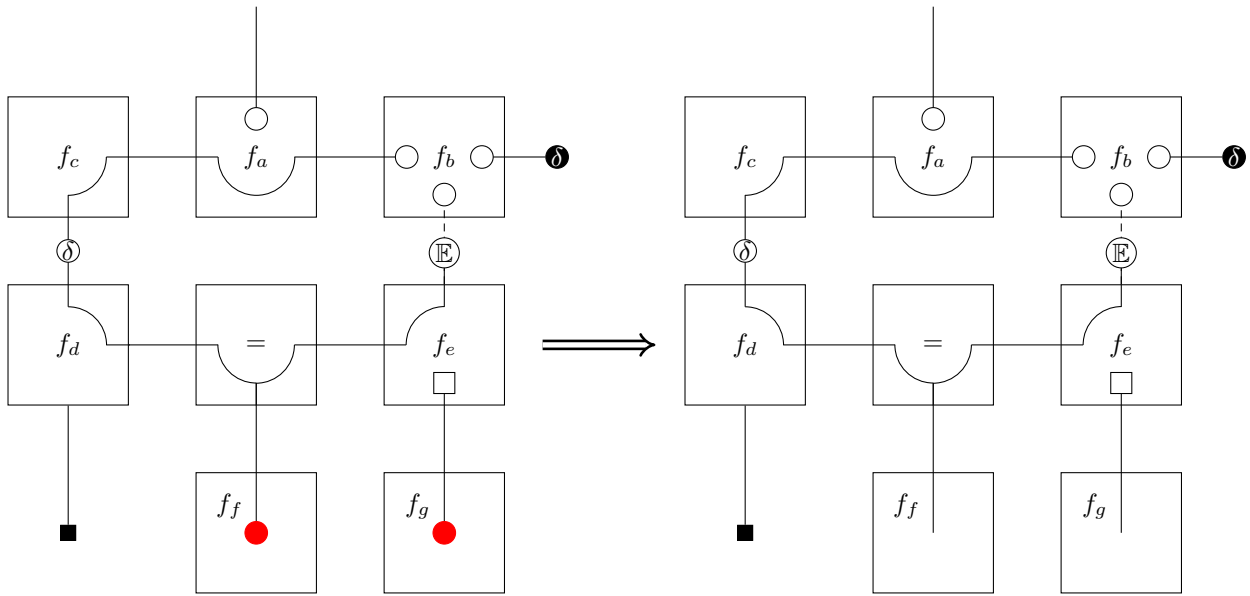

Note that the dangling edge extending from $f_a$ is treated as terminated by a bead and the bead on the edge is removed. Next, we remove any internal beads for nodes with no factorisation constraints. We show this step in Fig.~\ref{fig:shorthand_3}

\begin{figure}[H]
    \centering
    \makebox[0pt][c]{\begin{tikzpicture}
[node distance=25mm, auto]

\node[bigbox](fa){$f_a$};
\node[bigbox, right of=fa](fb){$f_b$};

\node[bigbox, left of=fa](fc){$f_c$};

\node[bigbox, below of=fc](fd){$f_d$};

\node[observation, below of=fd](obs1){};

\node[bigbox, below of=fb](fe){$f_e$};

\node[above of=fa, node distance=20mm](ghost1){};
\node[data, right of=fb, node distance=15mm](delta1){$\delta$};

\node[bigbox, right of=fd](eq1){$=$};

\node[bigbox, below of=eq1](ff){$f_f$};

\node[bigbox, below of=fe](fg){$f_g$};

\bigarcs{fa};
\bigarcne{fd};
\bigarcnw{fe};

\draw[](ghost1.center)--node[q,pos=1.0](){}($(fa.north) - (0.0,0.3)$);


\draw[](fa.east)--node[q,pos=1.0](){}($(fb.center) - (0.5,0.0)$);
\draw[](delta1)--node[q,pos=1.0](){}($(fb.center) + (0.5,0.0)$);

\draw[densely dashed](fe.north)--node[q,  solid, thin, minimum size=4mm, pos=0.35](E1){$\mathbb{E}$} node[q, solid, thin, pos=1.0](){}($(fb.center) - (0.0,0.5)$);
\draw[](fe.north)--(E1); 

\draw[](fg.north)--node[p,pos=1.0](){}($(fe.south) + (0.0,0.3)$);

\draw[](fa.west)--(fc.east);
\draw[](fc.south)--node[q,pos=0.5](){$\delta$}(fd.north);
\draw[](fd.east)--(eq1.west);
\draw[](fd)--(obs1);
\draw[](eq1.east)--(fe.west);

\draw[](eq1.south)--(ff.north);

\node[bigbox, left of=fa, node distance=90mm](fa2){$f_a$};
\node[bigbox, right of=fa2](fb){$f_b$};

\node[bigbox, left of=fa2](fc){$f_c$};

\node[bigbox, below of=fc](fd){$f_d$};

\node[observation, below of=fd](obs1){};

\node[bigbox, below of=fb](fe){$f_e$};

\node[above of=fa2, node distance=20mm](ghost1){};
\node[data, right of=fb, node distance=15mm](delta1){$\delta$};

\node[bigbox, right of=fd](eq1){$=$};

\node[bigbox, below of=eq1](ff){};
\node[above left of=ff, node distance=5mm](ff_f){$f_f$};

\node[bigbox, below of=fe](fg){};
\node[above left of=fg, node distance=5mm](fg_f){$f_g$};

\bigarcs{fa2};
\bigarcnw{fe};
\bigarcne{fd};

\draw[red] ($(fc.east) + (-0.3,0.0)$) arc[start angle=0, end angle=-90, radius=5.1mm];
\path[line] (fc.east) edge[-, draw=red] ($(fc.east) + (-0.3,0.0)$);
\path[line] (fc.south) edge[-, draw=red] ($(fc.south) + (0.0,0.3)$);

\draw[red] ($(eq1.east) + (-0.3,0.0)$) arc[start angle=0, end angle=-180, radius=5.1mm];
\path[line] (eq1.east) edge[-, draw=red] ($(eq1.east) + (-0.3,0.0)$);
\path[line] (eq1.south) edge[-, draw=red] ($(eq1.south) + (0.0,0.3)$);
\path[line] (eq1.west) edge[-, draw=red] ($(eq1.west) + (0.3,0.0)$);

\draw[](ghost1.center)--node[q,pos=1.0](){}($(fa2.north) - (0.0,0.3)$);

\draw[](fa2.east)--node[q,pos=1.0](){}($(fb.center) - (0.5,0,0)$);
\draw[](delta1)--node[q,pos=1.0](){}($(fb.center) + (0.5,0,0)$);

\draw[densely dashed](fe.north)--node[q,  solid, thin, minimum size=4mm, pos=0.35](E1){$\mathbb{E}$} node[q, solid, thin, pos=1.0](){}($(fb.center) - (0.0,0.5)$);
\draw[](fe.north)--(E1); 

\draw[](fg.north)--node[p,pos=1.0](){}($(fe.south) + (0.0,0.3)$);
\draw[red](fg.north)--(fg.center);

\draw[](fd)--(obs1);

\draw[](fa2.west)--(fc);
\draw[](fc)--node[q,pos=0.5](){$\delta$}(fd);
\draw[](fd.east)--(eq1.west);
\draw[](eq1.east)--(fe.west);

\draw[](eq1.south)--(ff.north);
\draw[red](ff.north)--(ff.center);

\draw[-Implies,line width=1,double distance=2pt](fe.east)++(0.5,0)--++(1.5,0);

\end{tikzpicture}}
    \caption{Step 3. Removing internal edges. Affected edges are highlighted in {\color{red} red}.}
    \label{fig:shorthand_4}
\end{figure}
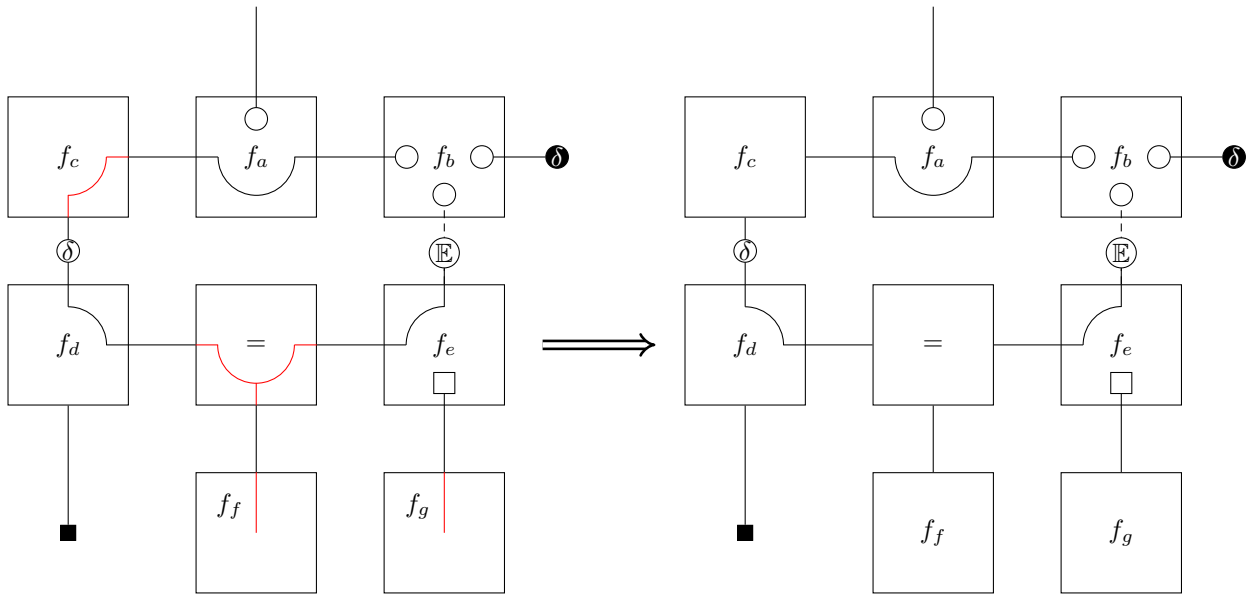

The next step is to remove any internal edge extensions for nodes with no factorisation constraints. We demonstrate this step in Fig.~\ref{fig:shorthand_4}. Note that the internal edge of the node $f_d$ does not get cancelled since one of the connected edges is a pointmass in the model and therefore not present in $q$, meaning $f_d$ does not have a default factorisation.

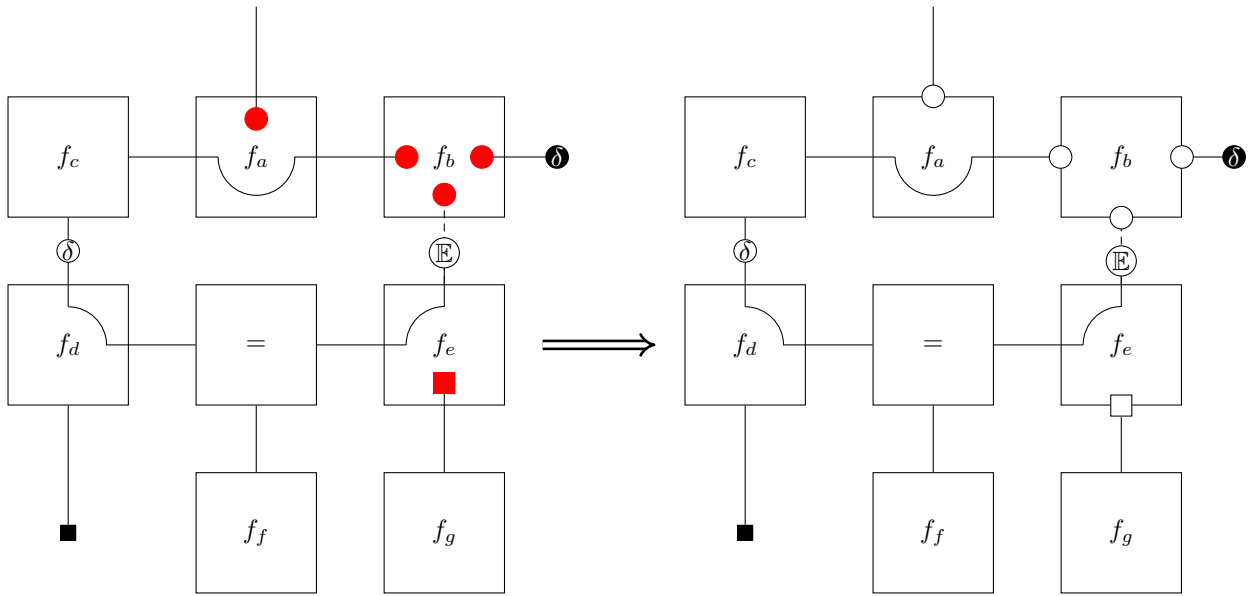
\begin{figure}[H]
    \centering
    \makebox[0pt][c]{\begin{tikzpicture}
[node distance=25mm, auto]

\node[bigbox](fa){$f_a$};
\node[bigbox, right of=fa](fb){$f_b$};

\node[bigbox, left of=fa](fc){$f_c$};

\node[bigbox, below of=fc](fd){$f_d$};

\node[observation, below of=fd](obs1){};

\node[bigbox, below of=fb](fe){$f_e$};

\node[above of=fa, node distance=20mm](ghost1){};
\node[data, right of=fb, node distance=15mm](delta1){$\delta$};

\node[bigbox, right of=fd](eq1){$=$};

\node[bigbox, below of=eq1](ff){$f_f$};

\node[bigbox, below of=fe](fg){$f_g$};

\bigarcs{fa};
\bigarcne{fd};
\bigarcnw{fe};

\draw[](ghost1.center)--node[q,pos=1.0, color=red](){}($(fa.north) - (0.0,0.3)$);


\draw[](fa.east)--node[q,pos=1.0, color=red](){}($(fb.center) - (0.5,0.0)$);
\draw[](delta1)--node[q,pos=1.0, color=red](){}($(fb.center) + (0.5,0.0)$);


\draw[densely dashed](fe.north)--node[q,  solid, thin, minimum size=4mm, pos=0.35](E1){$\mathbb{E}$} node[q, solid, thin, pos=1.0, color=red](){}($(fb.center) - (0.0,0.5)$);
\draw[](fe.north)--(E1); 

\draw[](fg.north)--node[p,pos=1.0, color=red](){}($(fe.south) + (0.0,0.3)$);

\draw[](fa.west)--(fc.east);
\draw[](fc.south)--node[q,pos=0.5](){$\delta$}(fd.north);
\draw[](fd.east)--(eq1.west);
\draw[](fd)--(obs1);
\draw[](eq1.east)--(fe.west);

\draw[](eq1.south)--(ff.north);

\draw[-Implies,line width=1,double distance=2pt](fe.east)++(0.5,0)--++(1.5,0);

\node[bigbox, right of=fa, node distance=90mm](fa2){$f_a$};
\node[bigbox, right of=fa2](fb){$f_b$};

\node[bigbox, left of=fa2](fc){$f_c$};

\node[bigbox, below of=fc](fd){$f_d$};

\node[observation, below of=fd](obs1){};

\node[bigbox, below of=fb](fe){$f_e$};

\node[above of=fa2, node distance=20mm](ghost1){};
\node[data, right of=fb, node distance=15mm](delta1){$\delta$};

\node[bigbox, right of=fd](eq1){$=$};

\node[bigbox, below of=eq1](ff){$f_f$};

\node[bigbox, below of=fe](fg){$f_g$};

\bigarcs{fa2};
\bigarcne{fd};
\bigarcnw{fe};

\draw[](ghost1.center)--node[q,pos=1.0](){}(fa2.north);

\draw[](fa2.east)--node[q,pos=1.0](){}(fb.west);
\draw[](delta1)--node[q,pos=1.0](){}(fb.east);


\draw[densely dashed](fe.north)--node[q,  solid, thin, minimum size=4mm, pos=0.35](E1){$\mathbb{E}$} node[q, solid, thin, pos=1.0](){}(fb.south);
\draw[](fe.north)--(E1); 

\draw[](fg.north)--node[p,pos=1.0](){}(fe.south);

\draw[](fa2.west)--(fc.east);
\draw[](fc.south)--node[q,pos=0.5](){$\delta$}(fd.north);
\draw[](fd.east)--(eq1.west);
\draw[](fd)--(obs1);
\draw[](eq1.east)--(fe.west);

\draw[](eq1.south)--(ff.north);
\end{tikzpicture}}
    \caption{Step 4. Pushing beads to node borders. Affected beads are highlighted in {\color{red} red}.}
    \label{fig:shorthand_5}
\end{figure}

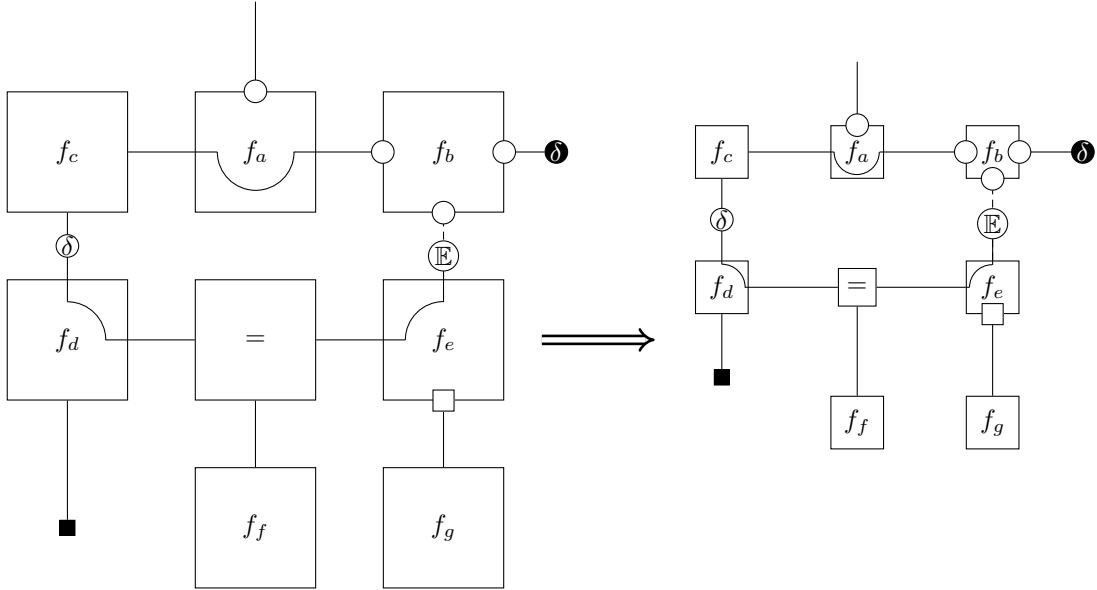
\begin{figure}[H]
    \centering
    \makebox[0pt][c]{\begin{tikzpicture}
[node distance=18mm, auto]

\node[stochastic](fa){$f_a$};
\node[stochastic, right of=fa](fb){$f_b$};

\node[stochastic, left of=fa](fc){$f_c$};

\node[stochastic, below of=fc](fd){$f_d$};

\node[observation, below of=fd, node distance=12mm](obs1){};

\node[stochastic, below of=fb](fe){$f_e$};

\node[above of=fa, node distance=12mm](ghost1){};
\node[data, right of=fb, node distance=12mm](delta1){$\delta$};

\node[deterministic, right of=fd](eq1){$=$};

\node[stochastic, below of=eq1](ff){$f_f$};

\node[stochastic, below of=fe](fg){$f_g$};

\arcs{fa};
\arcne{fd};
\arcnw{fe};

\draw[](ghost1.center)--node[q,pos=1.0](){}(fa.north);

\draw[](fa.east)--node[q,pos=1.0](){}(fb.west);
\draw[](delta1)--node[q,pos=1.0](){}(fb.east);

\draw[densely dashed](fe.north)--node[q,  solid, thin, minimum size=4mm, pos=0.45](E1){$\mathbb{E}$} node[q, solid, thin, pos=1.0](){}(fb.south);
\draw[](fe.north)--(E1); 

\draw[](fg.north)--node[p,pos=1.0](){}(fe.south);

\draw[](fa.west)--(fc.east);
\draw[](fc.south)--node[q,pos=0.5](){$\delta$}(fd.north);
\draw[](fd.east)--(eq1.west);
\draw[](fd)--(obs1);
\draw[](eq1.east)--(fe.west);

\draw[](eq1.south)--(ff.north);

\node[bigbox, left of=fa, node distance=80mm](fa2){$f_a$};
\node[bigbox, right of=fa2, node distance=25mm](fb){$f_b$};

\node[bigbox, left of=fa2, node distance=25mm](fc){$f_c$};

\node[bigbox, below of=fc, node distance=25mm](fd){$f_d$};

\node[observation, below of=fd, node distance=25mm](obs1){};

\node[bigbox, below of=fb, node distance=25mm](fe){$f_e$};

\node[above of=fa2, node distance=20mm](ghost1){};
\node[data, right of=fb, node distance=15mm](delta1){$\delta$};

\node[bigbox, right of=fd, node distance=25mm](eq1){$=$};

\node[bigbox, below of=eq1, node distance=25mm](ff){$f_f$};

\node[bigbox, below of=fe, node distance=25mm](fg){$f_g$};

\bigarcs{fa2};
\bigarcne{fd};
\bigarcnw{fe};

\draw[](ghost1.center)--node[q,pos=1.0](){}(fa2.north);

\draw[](fa2.east)--node[q,pos=1.0](){}(fb.west);
\draw[](delta1)--node[q,pos=1.0](){}(fb.east);

\draw[densely dashed](fe.north)--node[q,  solid, thin, minimum size=4mm, pos=0.35](E1){$\mathbb{E}$} node[q, solid, thin, pos=1.0](){}(fb.south);
\draw[](fe.north)--(E1); 

\draw[](fg.north)--node[p,pos=1.0](){}(fe.south);

\draw[](fa2.west)--(fc.east);
\draw[](fc.south)--node[q,pos=0.5](){$\delta$}(fd.north);
\draw[](fd.east)--(eq1.west);
\draw[](fd)--(obs1);
\draw[](eq1.east)--(fe.west);

\draw[](eq1.south)--(ff.north);

\draw[-Implies,line width=1,double distance=2pt](fe.east)++(0.5,0)--++(1.5,0);

\end{tikzpicture}}
    \caption{Compression allows for much more compact \cffgs which in turn are better suited for larger inference problems.}
    \label{fig:shorthand_6}
\end{figure}

Finally, we can push any remaining internal beads to their node border, illustrated in Fig.~\ref{fig:shorthand_5}. This step allows for writing the \cffg much more compactly, as we can see in Fig.~\ref{fig:shorthand_6}.

At this point, we have removed most of the beads and edges and still retain most of the relevant information around all nodes. What is left is only what deviates from a default \bfe specification. The goal here is, as stated initially, to make it easier to work with \cffg's for larger models which necessitate working with smaller nodes.

From Fig.~\ref{fig:shorthand_6} it is clear that compression allows for much more compact \cffg's which in turn are more amenable to larger \cffg's. This is especially true in the absence of any factorisation or form constraints and P-substitutions in which case the compressed \cffg and underlying \ffg are identical. We see this with nodes $f_c$, $f_f$, $f_g$ and $=$ in Fig.~\ref{fig:shorthand_6}.

\section{Classical AIF and the original GFE algorithm as special cases of LAIF}\label{sec:special_cases}

An integral part of working with \mpa algorithms is the choice of schedule - the order in which messages are passed. Many \mpa algorithms are iterative in nature and can therefore be sensitive to scheduling. \Laif is also an iterative \mpa algorithm and might consequently be sensitive to the choice of schedule.

A particularly interesting observation is that we can recover both the classical \aif planning algorithm of \cite{friston_active_2015} and the scheme proposed by \cite{parr_generalised_2019} as special cases of \laif by carefully choosing the schedule and performing model comparison. This result extends upon prior work by \cite{koudahl2023message} who reinterpreted the classical algorithm through the lens of \mpa on an \ffg.

\subsection{Example: CFFG and message updates for a discrete observation model with goals}

\begin{figure}[H]
    \centering
    \begin{tikzpicture}

\node[bigbox,minimum size=11mm](f){$T$};
\node[bigbox,minimum size=11mm, below of=f, node distance=20mm](g){$\mathcal{C}at$};
\node[above of=f, node distance=18mm](z){};
\node[left of=f, node distance=18mm](A){};

\node[below of=g, node distance=18mm](c){};

\draw[dashed](-0.9,0.9) rectangle (0.9,-2.9);
%
\draw[](f)--node[p,pos=0.0](){} node[left](){$\vect{x}$} node[q,pos=1.0](){}(g);
\draw[](f.north)--node[q, pos=0.0](){} node[left, pos=0.70](){$\vect{z}$} (z);
\draw[](f.west)--node[q, pos=0.0](){} node[above, pos=0.70](beadA){$\matr{A}$} (A);
\draw[](g.south)--node[q, pos=0.0](){} node[left, pos=0.70](){$\vect{c}$} (c);

\darkmsg{right}{down}{g}{c}{0.75}{1};
\darkmsg{right}{up}{f}{z}{0.75}{2};
\darkmsg{down}{left}{f}{A}{0.75}{3};
%

\end{tikzpicture}
%
%
%
%
%
%
%
    \caption{\cffg of composite node for \laif on discrete state spaces, reproduced from \cite{vandelaar2023realising}.
    }
    \label{fig:composite_node}
\end{figure}
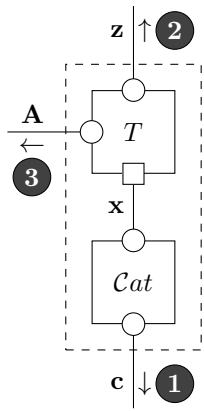

In order to both demonstrate \laif and recover prior work, we need to derive the required messages. We refer to our companion paper \cite{vandelaar2023realising} for the detailed derivations and summarise the results here. For the remaining paragraphs, we will use $\vect{x}$ to denote that $x$ is vector-valued and $\matr{A}$ to denote that $A$ is either a matrix or a tensor. For the discrete case, we work with a composite node corresponding to the factor

\begin{subequations}
\begin{align}\label{eq:composite_node}
   p(\vect{x} \mid \matr{A}, \vect{z}) &= \mathcal{C}at( \vect{x} \mid \matr{A}\vect{z})  \\
   \tilde{p}(\vect{x} \mid \vect{c}) &= \mathcal{C}at(\vect{x} \mid \vect{c}) \,.
\end{align}
\end{subequations}

We show the corresponding \cffg in Fig.~\ref{fig:composite_node} where we indicate the necessary factorisation and P-substitution. In the parlance of \aif Eq.~\ref{eq:composite_node} defines a model composed of a discrete state transition ($T$) with transition matrix $\matr{A}$ and a categorical goal prior ($\mathcal{C}at$) with parameter vector $\vect{c}$. Before stating the messages, we define the vector

\begin{align}
    \vect{h}(\matr{A}) &= -\mathrm{diag}(\matr{A}^T \log \matr{A})
\end{align}

This term is often denoted H in other works \cite{da_costa_active_2020,parr_generalised_2019}. We instead opt for using lowercase since the term is a vector and for making the dependence on $\matr{A}$ explicit.
In the following section, we will sometimes use an $\overline{\mathrm{overbar}}$ to denote expectations, such that $\mathbb{E}_{q(x)}[g(x)] = \overline{g(x)}$. Additionally, we define

\begin{subequations}
\begin{align}
    \vect{\upxi}(\matr{A}) &= \matr{A}^{\T}\!\left(\overline{\log\vect{c}} - \log(\overline{\matr{A}}\overline{\vect{z}})\right) - \vect{h}(\matr{A})\\
    \vect{\uprho} &= \overline{\matr{A}}^{\T}\!\left(\overline{\log\vect{c}} - \log(\overline{\matr{A}}\overline{\vect{z}})\right) - \overline{\vect{h}(\matr{A})}\,.
\end{align}
\end{subequations}

in order to make the expressions more concise. Where expectations cannot be computed in closed form, we instead resort to Monte Carlo estimates. With this notation in place, we can write the required messages as

\begin{figure}[H]
    \centering
    \def\arraystretch{1.5}
\begin{tabular}{r l}
    \hline
    $\mu_{\tinydarkcircled{1}}(\vect{c})$\hspace{-2.5mm} & $\propto \mathcal{D}ir (\vect{c} | \bar{\matr{A}}\bar{\vect{z}} + \vect{1})$ \\
    \hline
    \rowcolor{lightgray}
    $\mu_{\tinydarkcircled{2}}(\vect{z})$\hspace{-2.5mm} & $\propto \mathcal{C}at(\vect{z}|\sigma(\vect{\uprho}))$\\
    \hline
    Solve: $\bar{\vect{z}}$\hspace{-2.5mm} & $\stackrel{!}{=} \sigma\!\left(\vect{\uprho}(\bar{\vect{z}}) + \log\vect{d}\right)$\\
    $\mu_{\tinydarkcircled{2}}(\vect{z})$\hspace{-2.5mm} & $\propto \Cat{\vect{z}|\sigma(\log \bar{\vect{z}}^* - \log\vect{d})}$\\
    \hline
    \rowcolor{lightgray}
    $\log \mu_{\tinydarkcircled{3}}(\matr{A})$\hspace{-2.5mm} & $= \bar{\vect{z}}^{\T}\vect{\upxi}(\matr{A})$\\
    \hline
    $U_{\vect{x}}$\hspace{-2.5mm} & $= -\bar{\vect{z}}^{\T}\vect{\uprho}$\\
    \hline
\end{tabular}
    \caption{Message updates for the discrete composite node, reproduced from \cite{vandelaar2023realising}.}
    \label{fig:messages}
\end{figure}

Fig.~\ref{fig:messages} shows the message updates towards all connected variables as well as the average energy term ($U_x$) for the composite node. $\vect{d}$ denotes the parameters of the incoming message from the rest of the \cffg on the edge $z$. Interestingly, the energy term $U_x$ corresponds exactly to the \efe as used in standard \aif \cite{da_costa_active_2020, friston_active_2015}.

Special attention needs to be given to the messages $\mu_{\tinydarkcircled{2}}(\vect{z})$ and $\mu_{\tinydarkcircled{3}}(\matr{A})$. While $\mu_{\tinydarkcircled{2}}(\vect{z})$ can be solved for in closed form, in practice applying this result directly can lead to unstable solutions that fluctuate between multiple extrema. For this reason, we opt for parameterising the message by $\overline{\vect{z}}$ and solving for parameters of the marginal directly using Newtons method. Having found the optimum $\overline{\vect{z}}^*$ we can then substitute the result into the message expression to obtain a stable solution.

The message $\mu_{\tinydarkcircled{3}}(\matr{A})$ does not follow a nice exponential family distribution. To circumvent this problem, we can pass on the logpdf directly. When we need to compute the marginal $q(\matr{A})$, we can then use sampling procedures to estimate the necessary expectations \cite{akbayrak2021extended}. Further details and full derivations can be found in our companion paper \cite{vandelaar2023realising}.

\subsection{Reconstructing classical AIF}
With our new messages in hand and equipped with \cffg notation, we can now restate prior work unambiguously, starting with the classical algorithm of \cite{friston_active_2015}.
To reconstruct the algorithm of \cite{friston_active_2015}, we start by defining the generative model. The generative model is a discrete \pomdp over future time steps given by

\begin{align}\label{eq:hmm}
    p(\underbrace{\vect{x}_{t+1:T},\vect{z}_{t:T}}_{\text{Future}}& \mid \underbrace{\vect{\hat{u}}_{t+1:T}}_{\text{Policy}}, \underbrace{\vect{x}_{1:t},\vect{u}_{1:t}}_{\text{Past}} ) \nonumber \\
    &\propto \underbrace{p(\vect{z}_t \mid \vect{x}_{1:t},\vect{u}_{1:t})}_{\text{State Prior}}
   \prod_{k=t+1}^T \underbrace{p(\vect{x}_k \mid \vect{z}_k)}_{\text{Likelihood}}
   \underbrace{p(\vect{z}_k \mid \vect{z}_{k-1}, \vect{\hat{u}}_k)}_{\text{State Transition}}
   \underbrace{\tilde{p}(\vect{x}_k)}_{\text{Goal prior}}
\end{align}

where

\begin{align}
    p(\vect{z}_t \mid \vect{x}_{1:t}, \vect{u}_{1:t}) &= \mathcal{C}at( \vect{z}_t \mid \vect{d}) \label{eq:px0} \\
    p(\vect{z}_k \mid \vect{z}_{k-1}, \vect{\hat{u}}_k) &= \mathcal{C}at( \vect{z}_k \mid \matr{B}_{\hat{u}_k} \vect{z}_{k-1}) \\
    p(\vect{x}_k \mid \vect{z}_k) &= \mathcal{C}at( \vect{x}_k \mid \matr{A} \vect{z}_k) \\
    \tilde{p}(\vect{x}_k) &= \mathcal{C}at(\vect{x}_k \mid \vect{c}_k) \,.
\end{align}

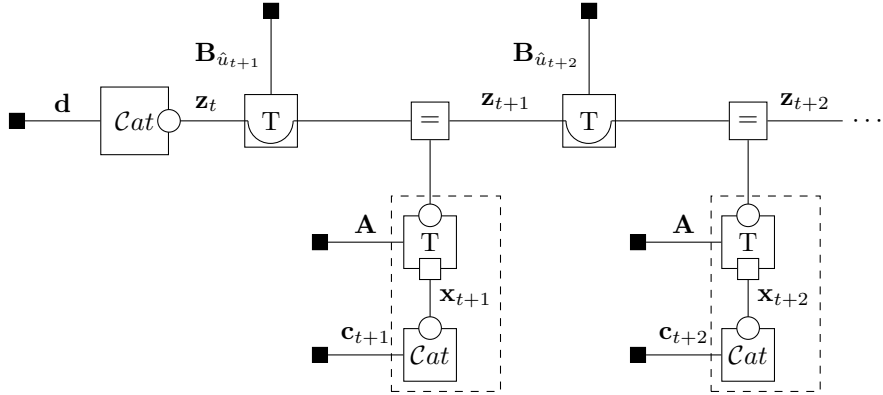
\begin{figure}[H]
    \centering
    \begin{tikzpicture}

\node[stochastic](f){T};
\node[bigstoc, left=10mm of f](cat){$\mathcal{C}at$};
\node[observation, left=10mm of cat](D){};
\node[deterministic,right= 15mm of f](eq1){$=$};
\node[stochastic,below= 10mm of eq1](g1){T};
\node[observation,above= 10mm of f](u1){};
\node[stochastic,right= 15mm of eq1](f2){T};
\node[observation,left= 10mm of g1](A1){};

\node[below= 1mm of f](equation){};
\node[right= 1mm of g1](equation){};

\node[observation,above= 10mm of f2](u2){};
\node[deterministic,right= 15mm of f2](eq2){$=$};
\node[stochastic,below= 10mm of eq2](g2){T};
\node[observation,left= 10mm of g2](A2){};

\node[stochastic, below of=g1, node distance=15mm](goal1){$\mathcal{C}at$};
\node[stochastic, below of=g2, node distance=15mm](goal2){$\mathcal{C}at$};

\node[observation, below of=A1, node distance=15.0mm](c1){};
\node[observation, below of=A2, node distance=15.0mm](c2){};

\node[right=10mm of eq2](ghost2){$\cdots$};

\draw[](cat)--node[above]{$\vect{z}_t$} node[q, pos=0.0](){}(f);
\draw[](D)--node[above]{$\vect{d}$}(cat);
\draw[](u1)--node[left]{$\matr{B}_{\hat{u}_{t+1}}$}(f);
\draw[](f)-- (eq1);
\draw[](eq1)-- node[q,pos=1.0]{} (g1);
\draw[](g1)--node[right]{$\vect{x}_{t+1}$}node[p,pos=0.0]{} node[q,pos=1.0]{} (goal1);
\draw[](A1)--node[above]{$\matr{A}$}(g1);
\draw[](c1)--node[above]{$\vect{c}_{t+1}$}(goal1);

\draw[](eq1)--node[above]{$\vect{z}_{t+1}$}(f2);
\draw[](u2)--node[left]{$\matr{B}_{\hat{u}_{t+2}}$}(f2);
\draw[](f2)--(eq2);
\draw[](eq2)-- node[q,pos=1.0]{} (g2);
\draw[](g2)--node[right]{$\vect{x}_{t+2}$}node[p,pos=0.0]{} node[q,pos=1.0]{} (goal2);
\draw[](A2)--node[above]{$\matr{A}$}(g2);
\draw[](c2)--node[above]{$\vect{c}_{t+2}$}(goal2);

\draw[](eq2)--node[above]{$\vect{z}_{t+2}$}(ghost2);

\arcs{f}
\arcs{f2}

\draw[dashed](1.6,-1.0) rectangle (3.05,-3.6);
\draw[dashed](5.85,-1.0) rectangle (7.3,-3.6);
\end{tikzpicture}
    \caption{\cffg of discrete \pomdp as used for planning in standard \aif models.}
    \label{fig:hmm}
\end{figure}

Here we let $\vect{x}_k$ denote observations, $\vect{z}_k$ latent states, and $\vect{\hat{u}}_k$ a fixed control, with the subscript $k$ indicating the future time step in question. The initial state $\vect{z}_t$ represents the filtering solution given the agents trajectory so far which we summarise in the parameter vector $\vect{d}$. Control signals correspond to particular transition matrices $\matr{B}_{\vect{\hat{u}}_k}$ where we use the subscript to emphasize that each $\matr{B}$ matches a particular control $\vect{\hat{u}}_k$. The observation model is given by the known matrix $\matr{A}$. Note that Eq.~\eqref{eq:hmm} is not normalised as it includes goal priors over $\vect{x}$. The \cffg of \eqref{eq:hmm} is shown in Fig~\ref{fig:hmm}.

where $T$ nodes denote a discrete state transition (multiplication of a categorical variable by a transition matrix). Given a generative model with a fixed set of controls, the next step is to compute the \efe \cite{da_costa_active_2020,friston_active_2015,koudahl2023message,koudahl_2021_lgds}. We provide a brief description here and refer to Appendix \ref{sec:efe} and \cite{da_costa_active_2020, koudahl2023message, koudahl_2021_lgds, sajid_active_2020} for more detailed descriptions. The \efe is given by

\begin{align}
    G(\hat{\vect{u}}_{t+1:T})
    &= \sum_{k=t}^T \iint p(\vect{x}_k \mid \vect{z}_k) q(\vect{z}_k \mid \vect{\hat{u}}_k) \log \frac{q(\vect{z_k} \mid \vect{\hat{u}}_k)}{p(\vect{x}_k, \vect{z}_k \mid \vect{\hat{u}}_k) } \dd \vect{x}_k \dd \vect{z}_k
\end{align}

With a slight abuse of notation, we can compute \efe by first applying transition matrices to the latent state and generating predicted observations as

\begin{align} \label{eq:efe_messages}
\begin{split}
   \vect{z}_k &= \matr{B}_{\hat{u}_k} \vect{z}_{k-1} \\
   \vect{x}_k &= \matr{A} \vect{z}_k \,.
\end{split}
\end{align}

In Eq.~\eqref{eq:efe_messages} we slightly abuse notation by having $\vect{z}_k$ (resp. $\vect{x}_k$)  refer to the prediction after applying the transition matrix $\matr{B}_{\hat{u}_k}$ (resp. $\matr{A}$) instead of the random variable as we have done elsewhere.

We can recognise these operations as performing a forwards \mpa sweep using belief propagation messages. With this choice of generative model, the \efe of a policy $\vect{\hat{u}}_{t+1:T}$ is found by \cite[Eq. D.2-3]{da_costa_active_2020}.

\begin{align}\label{eq:computation}
    G(\hat{\vect{u}}_{t+1:T}) &= \sum_{k=t+1}^{T} - \diag{\matr{A}^{T} \log \matr{A}}^{T} \vect{z}_k + \vect{x}_k^{T} (\log \vect{x}_k - \log \vect{c}_k)
\end{align}

where $\vect{c}_k$ denotes the parameter vector of a goal prior at the $k$'th time step. To select a policy we simply pick the sequence $\vect{\hat{u}}_{t+1:T}$ that results in the lowest numerical value when solving Eq.~\eqref{eq:computation}.

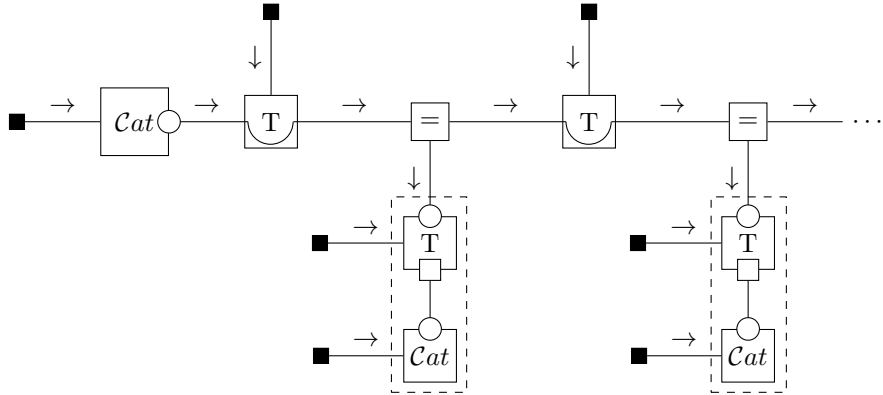
\begin{figure}[H]
    \centering
    \begin{tikzpicture}

\node[stochastic](f){T};
\node[bigstoc, left=10mm of f](cat){$\mathcal{C}at$};
\node[observation, left=10mm of cat](D){};
\node[deterministic,right= 15mm of f](eq1){$=$};
\node[stochastic,below= 10mm of eq1](g1){T};
\node[observation,above= 10mm of f](u1){};
\node[stochastic,right= 15mm of eq1](f2){T};
\node[observation,left= 10mm of g1](A1){};

\node[below= 1mm of f](equation){};
\node[right= 1mm of g1](equation){};

\node[observation,above= 10mm of f2](u2){};
\node[deterministic,right= 15mm of f2](eq2){$=$};
\node[stochastic,below= 10mm of eq2](g2){T};
\node[observation,left= 10mm of g2](A2){};

\node[stochastic, below of=g1, node distance=15mm](goal1){$\mathcal{C}at$};
\node[stochastic, below of=g2, node distance=15mm](goal2){$\mathcal{C}at$};

\node[observation, left of=goal1, node distance=14.5mm](c1){};
\node[observation, left of=goal2, node distance=14.5mm](c2){};

\node[right=10mm of eq2](ghost2){$\cdots$};

\draw[](cat)--node[above]{$\rightarrow$}node[q,pos=0.0]{}(f);
\draw[](D)--node[above]{$\rightarrow$}(cat);
\draw[](u1)--node[left]{$\downarrow$}(f);
\draw[](f)--node[above]{$\rightarrow$}(eq1);
\draw[](eq1)--node[left]{$\downarrow$} node[q,pos=1.0](){} (g1);
\draw[](g1)--node[p,pos=0.0]{}node[q,pos=1.0]{}(goal1);
\draw[](A1)--node[above]{$\rightarrow$}(g1);
\draw[](c1)--node[above]{$\rightarrow$}(goal1);

\draw[](eq1)--node[above]{$\rightarrow$}(f2);
\draw[](u2)--node[left]{$\downarrow$}(f2);
\draw[](f2)--node[above]{$\rightarrow$}(eq2);
\draw[](eq2)--node[left]{$\downarrow$} node[q,pos=1.0](){} (g2);
\draw[](g2)--node[p,pos=0.0]{}node[q,pos=1.0]{}(goal2);
\draw[](A2)--node[above]{$\rightarrow$}(g2);
\draw[](c2)--node[above]{$\rightarrow$}(goal2);

\draw[](eq2)--node[above]{$\rightarrow$}(ghost2);

\draw[dashed](1.6,-1.0) rectangle (2.6,-3.6);
\draw[dashed](5.85,-1.0) rectangle (6.85,-3.6);

\arcs{f}
\arcs{f2}

\end{tikzpicture}
    \caption{Classical \efe computation on the corresponding \cffg.}
    \label{fig:efe_schedule}
\end{figure}

To write this method on the \cffg in Fig.~\ref{fig:hmm}, we can simply add messages to the \cffg and note that the sum of energy terms of the P-substituted composite nodes in Fig.~\ref{fig:messages} matches Eq.~\eqref{eq:computation}. We show this result in Fig.~\ref{fig:efe_schedule}.

Comparing different policies (choices of $\vect{\hat{u}}_{t+1:T}$) and computing the energy terms of the P-substituted composite nodes is then exactly equal to the \efe-computation detailed in \cite{friston_active_2015}.

\subsection{Reconstructing the original GFE method}
We can further exemplify the capabilities of \cffg notation by recapitulating the update rules given in the original \gfe paper \cite{parr_generalised_2019}. To recover the procedure of \cite{parr_generalised_2019}, we need to extend the generative model to encompass past observations as

\begin{align}
\begin{split}
    p(\vect{x}_{1:T} &,\vect{z}_{0:T} \mid \vect{\hat{u}}_{1:T} ) \\
    &\propto p(\vect{z}_0)
    \underbrace{\prod_{l=1}^t p(\vect{x}_l \mid \vect{z}_l) p(\vect{z}_l \mid \vect{z}_{l-1}, \vect{\hat{u}}_l)}_{\text{Past}}
    \underbrace{ \prod_{k=t+1}^T p(\vect{x}_k \mid \vect{z}_k)
   p(\vect{z}_k \mid \vect{z}_{k-1}, \vect{\hat{u}}_k)
   \overbrace{\tilde{p}(\vect{x}_k)}^{\text{Goal prior}}}_{\text{Future}}
\end{split}
\end{align}

where

\begin{align}
\begin{split}
    p(\vect{z}_0) &= \mathcal{C}at( \vect{z}_0 \mid \vect{d}) \\
    p(\vect{z}_t \mid \vect{z}_{t-1},\vect{\hat{u}}_t) &= \mathcal{C}at( \vect{z}_t \mid \matr{B}_{\vect{\hat{u}}_t} \vect{z}_{t-1}) \\
    p(\vect{x}_t \mid \vect{z}_t) &= \mathcal{C}at( \vect{x}_t \mid \matr{A} \vect{z}_t) \\
    \tilde{p}(\vect{x}_k) &= \mathcal{C}at(\vect{x}_k \mid \vect{c}_k) \,.
\end{split}
\end{align}

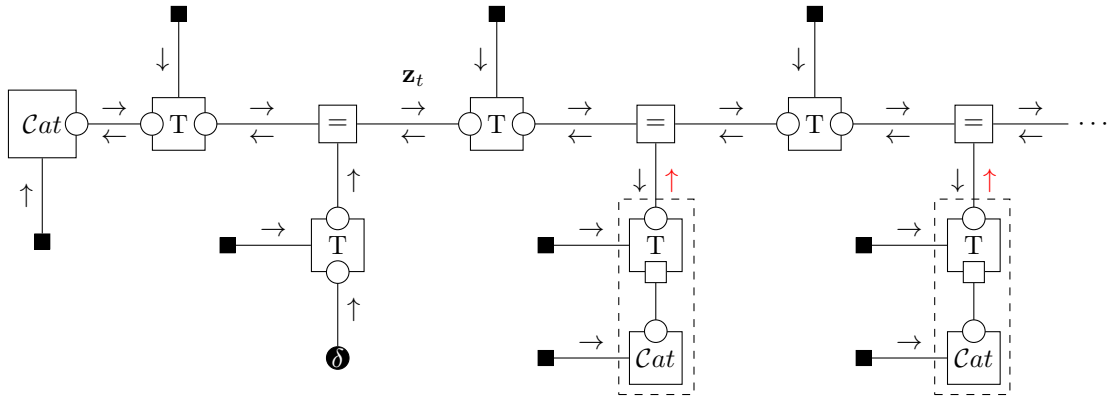
\begin{figure}[H]
    \centering
    \begin{tikzpicture}

\node[stochastic](f){T};
\node[bigstoc, left=10mm of f](cat){$\mathcal{C}at$};
\node[observation, below=10mm of cat](D){};
\node[deterministic,right= 15mm of f](eq1){$=$};
\node[stochastic,below= 10mm of eq1](g1){T};
\node[observation,above= 10mm of f](u1){};
\node[observation,left= 10mm of g1](A1){};
\node[data, below of=g1, node distance=15mm](goal1){$\delta$};

\node[stochastic,right= 15mm of eq1](f2){T};
\node[observation,above= 10mm of f2](u2){};
\node[deterministic,right= 15mm of f2](eq2){$=$};
\node[stochastic,below= 10mm of eq2](g2){T};
\node[observation,left= 10mm of g2](A2){};
\node[stochastic, below of=g2, node distance=15mm](goal2){$\mathcal{C}at$};

\node[stochastic,right= 15mm of eq2](f3){T};
\node[observation,above= 10mm of f3](u3){};
\node[deterministic,right= 15mm of f3](eq3){$=$};
\node[stochastic,below= 10mm of eq3](g3){T};
\node[observation,left= 10mm of g3](A3){};
\node[stochastic, below of=g3, node distance=15mm](goal3){$\mathcal{C}at$};

\node[observation, below of=A2, node distance=15mm](c2){};
\node[observation, below of=A3, node distance=15mm](c3){};

\node[right=10mm of eq3](ghost2){$\cdots$};

\draw[](cat)--node[above]{$\rightarrow$}node[below]{$\leftarrow$}node[q,pos=1.0]{}node[q,pos=0.0]{}(f);
\draw[](D)--node[left]{$\uparrow$}(cat);
\draw[](u1)--node[left]{$\downarrow$}(f);
\draw[](f)--node[q,pos=0.0]{}node[above]{$\rightarrow$}node[below]{$\leftarrow$}(eq1);
\draw[](eq1)--node[right]{$\uparrow$}(g1)node[q,pos=1.0]{};
\draw[](g1)--node[q,pos=0.0]{}node[right]{$\uparrow$}(goal1);
\draw[](A1)--node[above]{$\rightarrow$}(g1);

\draw[](eq1)--node[above](arrow){$\rightarrow$}node[below]{$\leftarrow$}(f2)node[q,pos=1.0]{};
\draw[](u2)--node[left]{$\downarrow$}(f2);
\draw[](f2)--node[q,pos=0.0]{}node[above]{$\rightarrow$}node[below]{$\leftarrow$}(eq2);
\draw[](eq2)--node[left]{$\downarrow$}node[right]{$\textcolor{red}{\uparrow}$}(g2)node[q,pos=1.0]{};
\draw[](g2)--node[p,pos=0.0]{}node[q,pos=1.0]{}(goal2);
\draw[](A2)--node[above]{$\rightarrow$}(g2);
\draw[](c2)--node[above]{$\rightarrow$} (goal2);

\node[above= 0mm of arrow](marker){$\vect{z}_t$};

\draw[](eq2)--node[above]{$\rightarrow$}node[below]{$\leftarrow$}(f3)node[q,pos=1.0]{};
\draw[](u3)--node[left]{$\downarrow$}(f3);
\draw[](f3)--node[q,pos=0.0]{}node[above]{$\rightarrow$}node[below]{$\leftarrow$}(eq3);
\draw[](eq3)--node[left]{$\downarrow$}node[right]{$\textcolor{red}{\uparrow}$}(g3)node[q,pos=1.0]{};
\draw[](g3)--node[p,pos=0.0]{} node[q,pos=1.0]{} (goal3);
\draw[](A3)--node[above]{$\rightarrow$}(g3);
\draw[](c3)--node[above]{$\rightarrow$}(goal3);

\draw[](eq3)--node[above]{$\rightarrow$}node[below]{$\leftarrow$}(ghost2);

\draw[dashed](5.85,-1.0) rectangle (6.85,-3.6);
\draw[dashed](10.05,-1.0) rectangle (11.05,-3.6);

\end{tikzpicture}
    \caption{\gfe computation on the \cffg.}
    \label{fig:gfe_schedule}
\end{figure}

For past time steps, we add data constraints and for future time steps we perform P-substitution. We also apply a naive mean field factorisation for every node. The final ingredient we need is a schedule that includes both forwards and backwards passes as hinted at in \cite{koudahl2023message}. For $t=1$ the corresponding \cffg is shown in Fig.~\ref{fig:gfe_schedule} with messages out of the P-substituted nodes highlighted in \textcolor{red}{red}. These messages are given by $\mu_{\tinydarkcircled{2}}(\vect{z})$ in Fig.~\ref{fig:messages}.

With these choices, the update equations become identical to those of \cite{parr_generalised_2019}. Indeed, careful inspection of the update rules given in \cite{parr_generalised_2019} reveals the component parts of the P-substituted message. However, by using \cffgs and P-substitution, we can cast their results as \mpa on generic graphs which immediately generalises their results to free-form graphical models.

Once inference has converged, we note that \cite{parr_generalised_2019} shows that \gfe evaluates identically to the \efe, meaning we can use the same model comparison procedure to select between policies as we used for reconstructing the classical algorithm. This shows how we can obtain the algorithm of \cite{parr_generalised_2019} as a special case of \laif.

\section{LAIF for policy inference}\label{sec:experiments}

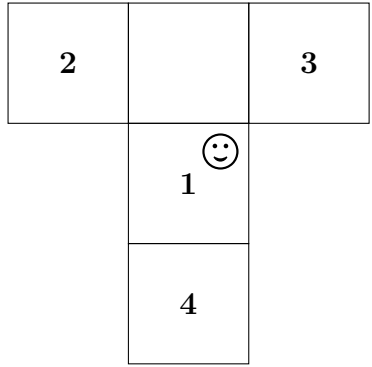
\begin{figure}[H]
    \centering
    \begin{tikzpicture}
[node distance=16mm, auto]
[font=\large\bfseries, auto]

\node[bigbox](1){1};
\node[bigbox, above of=1](5){};
\node[bigbox, right of=5](3){3};
\node[bigbox, left of=5](2){2};
\node[bigbox, below of=1](4){4};

\node[above right of=1, node distance=6mm](){\huge{\textmd{\smiley}}};

\end{tikzpicture}
    \caption{The T-maze environment}
    \label{fig:tmaze}
\end{figure}

The tools presented in this chapter are not limited to restating prior work. Indeed, \laif offers several advantages over prior methods, one of which is the ability to directly infer a policy instead of relying on a post hoc comparison of energy terms. To demonstrate, we solve two variations of the classic T-maze task \cite{friston_active_2015}. This is a well-studied setting within the \aif literature and therefore constitutes a good minimal benchmark. In the T-maze experiment, the agent lives in a maze with four locations as depicted in Fig.~\ref{fig:tmaze}. The agent (\smiley) starts in position 1 and knows that a reward is present at either position 2 or 3, but not which one. At position 4 is a cue that informs the agent which arm contains the reward. The optimal action to take is therefore to first visit 4 and learn which arm contains the reward before going to the rewarded arm. Because this course of action requires delaying the reward, an agent following a greedy policy behaves sub-optimally. The T-maze is therefore considered a reasonable minimal example of the epistemic, information-seeking behaviour that is a hallmark of \aif agents. We implemented our experiments in the reactive \mpa toolbox \texttt{RxInfer} \cite{bagaev_reactive_2022}. The source code for our simulations is available at \url{https://github.com/biaslab/LAIF}.

\subsection{Model specification}
The generative model for the T-maze is an adaptation of the discrete \pomdp used by \cite{friston_active_2015, parr_generalised_2019}. We assume the agent starts at some current timestep $t$ and want to infer a policy up to a known time horizon $T$. The generative model is then

\begin{align}
    p(\seq{x},\seq{z}, \seq{u})
    &\propto \underbrace{p(\vect{z}_t)}_{\substack{\text{Initial}\\ \text{state}}} \prod_{k=t+1}^T
    \underbrace{p(\vect{x}_k \mid \vect{z}_k)}_{\substack{\text{Observation}\\\text{model}}}
    \underbrace{p(\vect{z}_k \mid \vect{z}_{k-1}, \vect{u}_k)}_{\substack{\text{Transition}\\\text{model}}}
    \underbrace{p(\vect{u}_k)}_{\substack{\text{Control}\\ \text{prior}}}
    \underbrace{\Tilde{p}(\vect{x}_k)}_{\substack{\text{Goal}\\ \text{prior}}}
\end{align}

where

\begin{subequations}
\begin{align}
    p(\vect{z}_t) &= \mathcal{C}at(\vect{z}_t | \vect{d}) \\
    p(\vect{x}_k | \vect{z}_k) &= \mathcal{C}at(\vect{x}_k | \matr{A} \vect{z}_k) \\
    p(\vect{z}_k | \vect{z}_{k-1}, \vect{u}_k) &= \prod_n \mathcal{C}at( \vect{z}_k | \matr{B}_n \vect{z}_{k-1})^{\vect{u}_{nk}} \label{eq:transition_mixture}\\
    p(\vect{u}_k) &= \mathcal{C}at(\vect{u}_k | \vect{e}_k) \\
    \tilde{p}(\vect{x}_k) &= \mathcal{C}at(\vect{x}_k | \vect{c}_k) \,.
\end{align}
\end{subequations}

\begin{figure}[H]
    \centering
    \begin{tikzpicture}
    [node distance=20mm,auto]

   \node[observation](d){}; 
   \node[bigstoc, right of=d](z0){$\mathcal{C}at$};
   \node[bigstoc, minimum size=8mm, right of=z0](t1){$\mathcal{T}M$};
   \node[deterministic, right of=t1](eq1){=};
   \node[bigstoc, minimum size=8mm, right of=eq1](t2){$\mathcal{T}M$};
   \node[right of=t2](eq2){};

    \draw(t1.south)++(-2.0mm,0.0mm)--++(0.0,-1.0) node[observation,pos=1.0, xshift=-1.2mm](){}node[left, pos=0.5](){$\matr{B}_1$};
    \draw(t1.south)++(2.0mm,0.0mm)--++(0.0,-1.0) node[observation,pos=1.0, xshift=-1.2mm](){} node[right, pos=0.5](){$\matr{B}_4$}node[pos=0.5](b41){};
    \node[left of=b41, node distance=3mm](){$\cdots$};
    
    \draw(t2.south)++(-2.0mm,0.0mm)--++(0.0,-1.0) node[observation,pos=1.0, xshift=-1.2mm](){}node[left, pos=0.5](){$\matr{B}_1$};
    \draw(t2.south)++(2.0mm,0.0mm)--++(0.0,-1.0) node[observation,pos=1.0, xshift=-1.2mm](){} node[right, pos=0.5](){$\matr{B}_4$}node[pos=0.5](b42){};
    \node[left of=b42, node distance=3mm](){$\cdots$};

    \bigstocarcnw{t1};
    \bigstocarcne{t1};
    \bigstocarcnw{t2};
    \bigstocarcne{t2};

   \node[deterministic, minimum size=8mm, below of=eq1](obs1){$T$};
   \node[deterministic, minimum size=8mm, below of=eq2](obs2){$T$};
   \node[stochastic, below of=obs1](c1){$\mathcal{C}at$};
   \node[stochastic, below of=obs2](c2){$\mathcal{C}at$};
   \node[observation, node distance=18mm, left of= obs1](a1){};
   \node[observation, node distance=18mm, left of= obs2](a2){};
   \node[observation, node distance=18mm, left of= c1](goal1){};
   \node[observation, node distance=18mm, left of= c2](goal2){};

   \node[stochastic,above of=t1](u1){$\mathcal{C}at$};
   \node[stochastic,above of=t2](u2){$\mathcal{C}at$};
   \node[observation, left of=u1](e1){};
   \node[observation, left of=u2](e2){};

   \path[line] (d) edge[-, draw=black] node[above]{$\vect{d}$}(z0);
   \path[line] (z0) edge[-, draw=black] node[above]{$\vect{z}_t$} node[q,pos=0.0]{} (t1);
   \path[line] (t1) edge[-, draw=black] (eq1);
   \path[line] (eq1) edge[-, draw=black] node[above]{$\vect{z}_{t+1}$}(t2);
   \path[line] (t2) edge[-, draw=black] (eq2.center);

   \path[line] (u1) edge[-, draw=black] node[right]{$\vect{u}_t$} node[q,pos=0]{} (t1);
   \path[line] (u2) edge[-, draw=black] node[right]{$\vect{u}_{t+1}$} node[q,pos=0]{} (t2);
   \path[line] (e1) edge[-, draw=black] node[above]{$\vect{e}_t$} (u1);
   \path[line] (e2) edge[-, draw=black] node[above]{$\vect{e}_{t+1}$} (u2);
   
   \path[line] (eq1) edge[-, draw=black] node[q, pos=1]{} (obs1);
   \path[line] (obs1) edge[-, draw=black] node[p, pos=0]{} node[q,pos=1]{} (c1);
   
   \path[line] (eq2.center) edge[-, draw=black] node[q, pos=1]{} (obs2);
   \path[line] (obs2) edge[-, draw=black] node[p,pos=0]{} node[q,pos=1]{} (c2);

   \path[line] (a1) edge[-, draw=black] node[above, pos=0.4]{$\matr{A}_t$} (obs1);
   \path[line] (a2) edge[-, draw=black] node[above, pos=0.4]{$\matr{A}_{t+1}$} (obs2);
   
   \path[line] (goal1) edge[-, draw=black] node[above,pos=0.4]{$\vect{c}_t$} (c1);
   \path[line] (goal2) edge[-, draw=black] node[above,pos=0.4]{$\vect{c}_{t+1}$} (c2);

   \draw[densely dashed] (5.3,-1.3) rectangle ++(1.3, -3.3);
   \draw[densely dashed] (9.3,-1.3) rectangle ++(1.3, -3.3);
\end{tikzpicture}
    \caption{\cffg for the T-maze experiment}
    \label{fig:tmaze_model}
\end{figure}
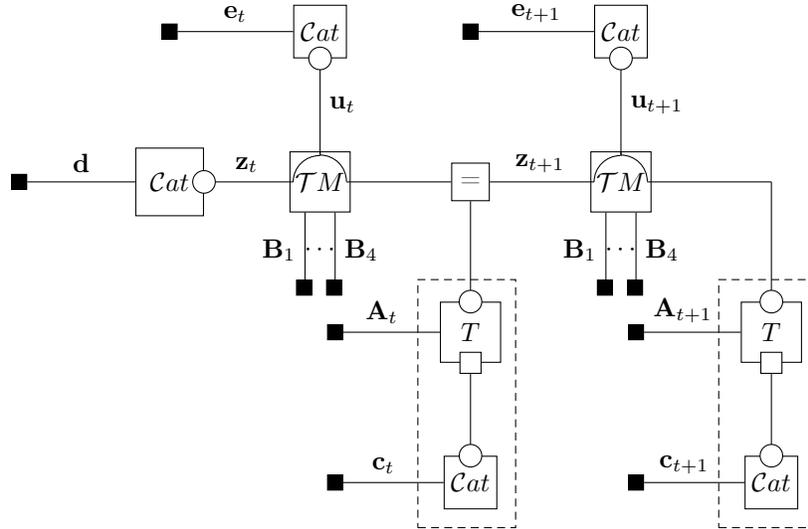

$\vect{u}_k$ is a one-hot encoded vector of length $n$. The notation $\vect{u}_{nk}$ picks the $n$'th entry of $\vect{u}_k$. We show the corresponding \cffg in Fig~\ref{fig:tmaze_model}.

\begin{figure}[H]
    \centering
    \begin{tikzpicture}
    [node distance=20mm,auto]

   \node[observation](d){}; 
   \node[bigstoc, right of=d](z0){$\mathcal{C}at$};
   \node[bigstoc, minimum size=8mm, right of=z0](t1){$\mathcal{T}M$};
   \node[deterministic, right of=t1](eq1){=};
   \node[bigstoc, minimum size=8mm, right of=eq1](t2){$\mathcal{T}M$};
   \node[right of=t2](eq2){};

    \draw(t1.south)++(-2.0mm,0.0mm)--++(0.0,-1.0) node[observation,pos=1.0, xshift=-1.2mm](){}node[left, pos=0.5](){$\uparrow$};
    \draw(t1.south)++(2.0mm,0.0mm)--++(0.0,-1.0) node[observation,pos=1.0, xshift=-1.2mm](){} node[right, pos=0.5](){$\uparrow$}node[pos=0.5](b41){};
    \node[left of=b41, node distance=3mm](){$\cdots$};

    \draw(t2.south)++(-2.0mm,0.0mm)--++(0.0,-1.0) node[observation,pos=1.0, xshift=-1.2mm](){}node[left, pos=0.5](){$\uparrow$};
    \draw(t2.south)++(2.0mm,0.0mm)--++(0.0,-1.0) node[observation,pos=1.0, xshift=-1.2mm](){} node[right, pos=0.5](){$\uparrow$}node[pos=0.5](b42){};
    \node[left of=b42, node distance=3mm](){$\cdots$};
    
    \bigstocarcnw{t1};
    \bigstocarcne{t1};
    \bigstocarcnw{t2};
    \bigstocarcne{t2};

   \node[deterministic, minimum size=8mm, below of=eq1](obs1){$T$};
   \node[deterministic, minimum size=8mm, below of=eq2](obs2){$T$};
   \node[stochastic, below of=obs1](c1){$\mathcal{C}at$};
   \node[stochastic, below of=obs2](c2){$\mathcal{C}at$};
   \node[observation, node distance=18mm, left of= obs1](a1){};
   \node[observation, node distance=18mm, left of= obs2](a2){};
   \node[observation, node distance=18mm, left of= c1](goal1){};
   \node[observation, node distance=18mm, left of= c2](goal2){};

   \node[stochastic,above of=t1](u1){$\mathcal{C}at$};
   \node[stochastic,above of=t2](u2){$\mathcal{C}at$};
   \node[observation, left of=u1](e1){};
   \node[observation, left of=u2](e2){};

   \path[line] (d) edge[-, draw=black] node[above]{$\rightarrow$}(z0);
   \path[line] (z0) edge[-, draw=black] node[above]{$\rightarrow$}node[below]{$\leftarrow$} node[q,pos=0.0]{} (t1);
   \path[line] (t1) edge[-, draw=black] node[above]{$\rightarrow$}node[below]{$\leftarrow$}(eq1);
   \path[line] (eq1) edge[-, draw=black] node[above]{$\rightarrow$}node[below]{$\leftarrow$}(t2);
   \path[line] (t2) edge[-, draw=black]  node[above]{$\rightarrow$}(eq2.center);

   \path[line] (u1) edge[-, draw=black] node[left]{$\uparrow$}node[right]{$\downarrow$} node[q,pos=0.0]{} (t1);
   \path[line] (u2) edge[-, draw=black] node[left]{$\uparrow$}node[right]{$\downarrow$} node[q,pos=0.0]{} (t2);
   \path[line] (e1) edge[-, draw=black] node[above]{$\rightarrow$} (u1);
   \path[line] (e2) edge[-, draw=black] node[above]{$\rightarrow$} (u2);
   
   \path[line] (eq1) edge[-, draw=black] node[left]{$\textcolor{red}{\uparrow}$}node[right]{$\downarrow$}node[q, pos=1]{} (obs1);
   \path[line] (obs1) edge[-, draw=black] node[p, pos=0]{} node[q,pos=1.0]{} (c1);
   
   \path[line] (eq2.center) edge[-, draw=black] node[left]{$\textcolor{red}{\uparrow}$} node[q, pos=1]{} (obs2);
   \path[line] (obs2) edge[-, draw=black] node[p, pos=0]{} node[q,pos=1.0]{} (c2);

   \path[line] (a1) edge[-, draw=black] node[above, pos=0.4]{$\rightarrow$} (obs1);
   \path[line] (a2) edge[-, draw=black] node[above, pos=0.4]{$\rightarrow$} (obs2);
   
   \path[line] (goal1) edge[-, draw=black] node[above, pos=0.4]{$\rightarrow$} (c1);
   \path[line] (goal2) edge[-, draw=black] node[above, pos=0.4]{$\rightarrow$} (c2);

   \draw[densely dashed] (5.3,-1.3) rectangle ++(1.3, -3.3);
   \draw[densely dashed] (9.3,-1.3) rectangle ++(1.3, -3.3);
\end{tikzpicture}
    \caption{Message passing schedule for the T-maze experiment}
    \label{fig:tmaze_schedule}
\end{figure}
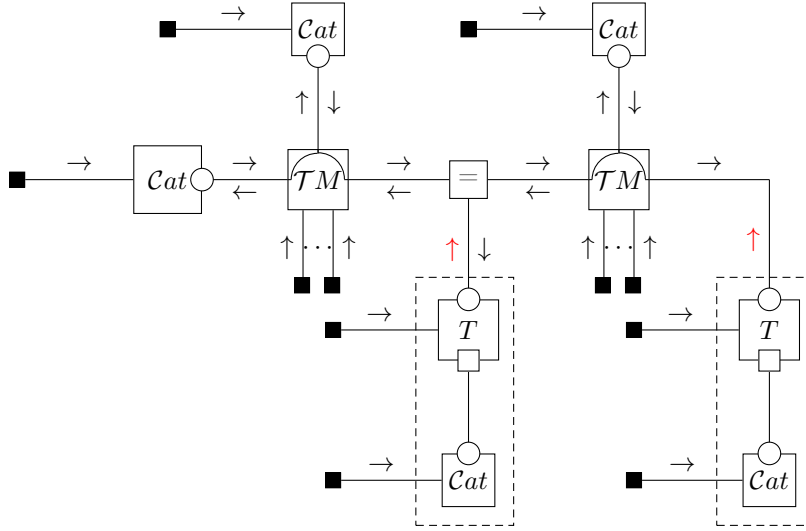

Eq.~\eqref{eq:transition_mixture} defines a mixture model over candidate transition matrices indexed by $\vect{u}_k$. We give the details of this node function and the required messages in Appendix~\ref{sec:transition_mixture}. The \mpa schedule is shown in Fig.~\ref{fig:tmaze_schedule} with \gfe-based messages highlighted in \textcolor{red}{red}

Following \cite{friston_active_2015} we define the initial state and control prior as

\begin{subequations}
\begin{align}
    \vect{d} &= (1,0,0,0)^T \otimes (0.5,0.5)^T\\
    \vect{e}_k &= (0.25,0.25,0.25,0.25) \forall k
\end{align}
\end{subequations}

with $\otimes$ denoting the Kronecker product. The transition mixture node requires a set of candidate transition matrices. The T-maze utilises four possible transitions, given below

\begin{align}
\begin{split}
    \matr{B}_1 = \begin{bmatrix}
    1 & 1 & 1 & 1 \\
    0 & 0 & 0 & 0 \\
    0 & 0 & 0 & 0 \\
    0 & 0 & 0 & 0
    \end{bmatrix} \otimes \matr{I}_2 \,,
    \matr{B}_2 = \begin{bmatrix}
    0 & 1 & 1 & 0 \\
    1 & 0 & 0 & 1 \\
    0 & 0 & 0 & 0 \\
    0 & 0 & 0 & 0
    \end{bmatrix} \otimes \matr{I}_2 \\
    \matr{B}_3 = \begin{bmatrix}
    0 & 1 & 1 & 0 \\
    0 & 0 & 0 & 0 \\
    1 & 0 & 0 & 1 \\
    0 & 0 & 0 & 0
    \end{bmatrix} \otimes \matr{I}_2 \,,
    \matr{B}_4 = \begin{bmatrix}
    0 & 1 & 1 & 0 \\
    0 & 0 & 0 & 0 \\
    0 & 0 & 0 & 0 \\
    1 & 0 & 0 & 1
    \end{bmatrix} \otimes \matr{I}_2
\end{split}
\end{align}

where $\matr{I}_2$ denotes a $2 \times 2$ identity matrix. Note that these differ slightly from the original implementation of \cite{friston_active_2015}. In \cite{friston_active_2015} invalid transitions are represented by an identity mapping where we instead model invalid transitions by sending the agent back to position 1. The likelihood matrix $\matr{A}$ is given by four blocks, corresponding to the observation likelihood in each position.

\begin{align}
    \matr{A} = \begin{bmatrix}
    \matr{A}_1 & & & \\
    & \matr{A}_2 & & \\
    & & \matr{A}_3 & \\
    & & & \matr{A}_4
    \end{bmatrix} \,,
\end{align}

with everything outside the blocks being set to 0. The blocks are

\begin{align}
    \notag &\matr{A}_1 = \begin{bmatrix}
    0.5 & 0.5 \\
    0.5 & 0.5 \\
    0 & 0 \\
    0 & 0
    \end{bmatrix}\,,
    &\matr{A}_2 = \begin{bmatrix}
    0 & 0 \\
    0 & 0  \\
    \alpha & 1 - \alpha \\
    1 - \alpha & \alpha
    \end{bmatrix} \\
   &\matr{A}_3 = \begin{bmatrix}
    0 & 0 \\
    0 & 0  \\
    1 - \alpha & \alpha \\
    \alpha & 1 - \alpha
    \end{bmatrix} \,,
    &\matr{A}_4 = \begin{bmatrix}
    1 & 0 \\
    0 & 1 \\
    0 & 0 \\
    0 & 0
    \end{bmatrix} \,,
\end{align}

with $\alpha$ being the probability of observing a reward. The goal prior is given by

\begin{align}
    \vect{c}_k &= \sigma \big( (0,0,c,-c)^T \otimes (1,1,1,1)^T \big) \forall k
\end{align}

with $c$ being the utility ascribed to a reward and $\sigma(\cdot)$ the softmax function.
Inference for the parts of the model not in $\mathcal{P}$ can be accomplished using belief propagation. We follow the experimental setup of \cite{friston_active_2015} and let $c = 2, \alpha = 0.9$. For inference, we perform two iterations of our \mpa procedure and use 20 Newton steps to obtain the parameters of the outgoing message from the P-substituted nodes.

We show the results in Fig.~\ref{fig:tmaze_results}. The number in each cell is the posterior probability mass assigned to the corresponding action, with the most likely actions highlighted in {\color{red} red}.

\begin{figure}[H]
    \centering
    \begin{tikzpicture}
[node distance=16mm, auto]
[font=\large, auto]

\node[bigbox](1){$0.25$};
\node[bigbox, above of=1](5){};
\node[bigbox, right of=5](3){$0.20$};
\node[bigbox, left of=5](2){$0.20$};
\node[bigbox, below of=1, color=red](4){$0.35$};
\node[above of=5, font=\large\bfseries](label1){Controls at time step 1};

\node[bigbox, right of=1, node distance=60mm](12){$0.13$};
\node[bigbox, above of=12](52){};
\node[bigbox, color=red, right of=52](32){$0.30$};
\node[bigbox, color=red, left of=52](22){$0.30$};
\node[bigbox, below of=12](42){$0.26$};
\node[above of=52, font=\large\bfseries](label2){Controls at time step 2};

\end{tikzpicture}
    \caption{Posterior controls for the T-maze experiment}
    \label{fig:tmaze_results}
\end{figure}

Fig.~\ref{fig:tmaze_results} shows an agent that initially prefers the epistemic action (move to state 4) at time $t+1$ and subsequently exhibits a preference for either of the potentially rewarding arms (indifferent between states 2 and 3). This shows that \laif is able to infer the optimal policy and that our approach can reproduce prior results on the T-maze.

\begin{figure}[H]
    \centering
    \begin{tikzpicture}
    [node distance=20mm,auto]

   \node[observation](d){}; 
   \node[bigstoc, right of=d](z0){$\mathcal{C}at$};
   \node[bigstoc, minimum size=8mm, right of=z0](t1){$\mathcal{T}M$};
   \node[deterministic, right of=t1](eq1){=};
   \node[bigstoc, minimum size=8mm, right of=eq1](t2){$\mathcal{T}M$};
   \node[right of=t2](eq2){};

    \draw(t1.south)++(-2.0mm,0.0mm)--++(0.0,-1.0) node[observation,pos=1.0, xshift=-1.2mm](){}node[left, pos=0.5](){$B_1$};
    \draw(t1.south)++(2.0mm,0.0mm)--++(0.0,-1.0) node[observation,pos=1.0, xshift=-1.2mm](){} node[right, pos=0.5](){$B_4$}node[pos=0.5](b41){};
    \node[left of=b41, node distance=3mm](){$\cdots$};
    
    \draw(t2.south)++(-2.0mm,0.0mm)--++(0.0,-1.0) node[observation,pos=1.0, xshift=-1.2mm](){}node[left, pos=0.5](){$B_1$};
    \draw(t2.south)++(2.0mm,0.0mm)--++(0.0,-1.0) node[observation,pos=1.0, xshift=-1.2mm](){} node[right, pos=0.5](){$B_4$}node[pos=0.5](b42){};
    \node[left of=b42, node distance=3mm](){$\cdots$};

    \bigstocarcnw{t1}
    \bigstocarcne{t1}
    \bigstocarcnw{t2}
    \bigstocarcne{t2}

   \node[deterministic, minimum size=8mm, below of=eq1](obs1){$T$};
   \node[deterministic, minimum size=8mm, below of=eq2](obs2){$T$};
   \node[stochastic, below of=obs1](c1){$\mathcal{C}at$};
   \node[stochastic, below of=obs2](c2){$\mathcal{C}at$};
   \node[observation, node distance=15mm, left of= obs1](a1){};
   \node[observation, node distance=15mm, left of= obs2](a2){};
   \node[observation, node distance=15mm, left of= c1](goal1){};
   \node[observation, node distance=15mm, left of= c2](goal2){};

   \node[stochastic,above of=t1](u1){$\mathcal{C}at$};
   \node[stochastic,above of=t2](u2){$\mathcal{C}at$};
   \node[observation, left of=u1](e1){};
   \node[observation, left of=u2](e2){};

   \path[line] (d) edge[-, draw=black] node[above]{}(z0);
   \path[line] (z0) edge[-, draw=black] node[q,pos=0.0]{}(t1);
   \path[line] (t1) edge[-, draw=black] (eq1);
   \path[line] (eq1) edge[-, draw=black] node[above]{}(t2);
   \path[line] (t2) edge[-, draw=black] (eq2.center);

   \path[line] (u1) edge[-, draw=black] node[right]{} node[delta,pos=0.5]{$\delta$} node[q,pos=0]{} (t1);
   \path[line] (u2) edge[-, draw=black] node[right]{} node[delta,pos=0.5]{$\delta$} node[q,pos=0]{}(t2);
   \path[line] (e1) edge[-, draw=black] node[above]{} (u1);
   \path[line] (e2) edge[-, draw=black] node[above]{} (u2);
   
   \path[line] (eq1) edge[-, draw=black] node[q, pos=1]{} (obs1);
   \path[line] (obs1) edge[-, draw=black] node[p, pos=0]{} node[q, pos=1]{} (c1);
   
   \path[line] (eq2.center) edge[-, draw=black] node[q, pos=1]{} (obs2);
   \path[line] (obs2) edge[-, draw=black] node[p, pos=0]{} node[q, pos=1]{} (c2);
                                                           
   \path[line] (a1) edge[-, draw=black] node[above, pos=0.4]{} (obs1);
   \path[line] (a2) edge[-, draw=black] node[above, pos=0.4]{} (obs2);
   
   \path[line] (goal1) edge[-, draw=black] node[above, pos=0.4]{} (c1);
   \path[line] (goal2) edge[-, draw=black] node[above, pos=0.4]{} (c2);

   \draw[densely dashed] (5.3,-1.3) rectangle ++(1.3, -3.3);
   \draw[densely dashed] (9.3,-1.3) rectangle ++(1.3, -3.3);
\end{tikzpicture}
    \caption{\cffg for the T-maze model with additional $\delta$-constraints}
    \label{fig:delta_tmaze}
\end{figure}
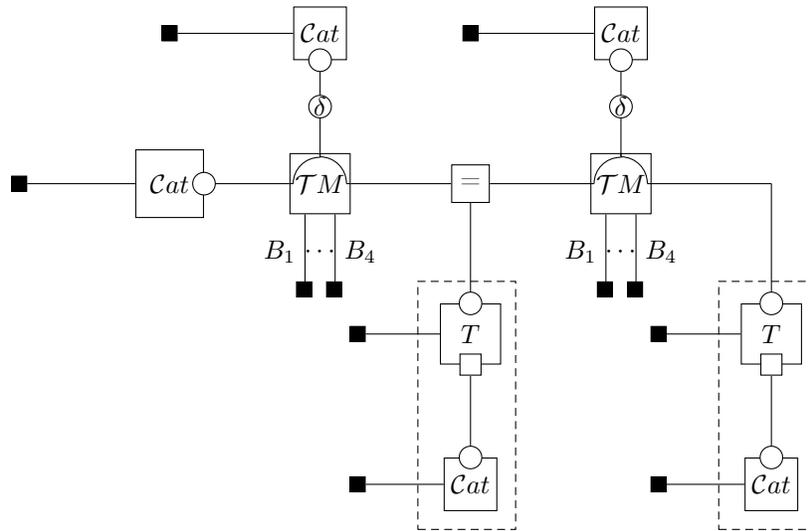

Since \cffgs are inherently modular, they allow us to modify the inference task without changing the model. To demonstrate, we now add $\delta$-constraints to the control variables. This corresponds to selecting the MAP estimate of the control posterior \cite{senoz_variational_2021} and results in the \cffg shown in Fig.~\ref{fig:delta_tmaze}. The schedule and all messages remain the same as our previous experiment - however now the agent will select the most likely course of action instead of providing us with a posterior distribution.

Performing this experiment yields the policy shown in Fig.~\ref{fig:tmaze_delta_results}

\begin{figure}[H]
    \centering
    \begin{tikzpicture}
[node distance=16mm, auto]
[font=\large, auto]

\node[bigbox](1){$0.0$};
\node[bigbox, above of=1](5){};
\node[bigbox, right of=5](3){$0.0$};
\node[bigbox, left of=5](2){$0.0$};
\node[bigbox, below of=1, color=red](4){$1.0$};
\node[above of=5, font=\large\bfseries](label1){Controls at time step 1};

\node[bigbox, right of=1, node distance=60mm](12){$0.0$};
\node[bigbox, above of=12](52){};
\node[bigbox, right of=52](32){$0.0$};
\node[bigbox, color=red, left of=52](22){$1.0$};
\node[bigbox, below of=12](42){$0.0$};
\node[above of=52, font=\large\bfseries](label2){Controls at time step 2};

\end{tikzpicture}
    \caption{Posterior controls for the T-maze experiment with $\delta$-constraints}
    \label{fig:tmaze_delta_results}
\end{figure}

Fig.~\ref{fig:tmaze_delta_results} once again shows that the agent is able to infer the optimal policy for solving the task. For repeated runs, the agent will randomly select to move to either position 2 or 3 at the second step due to minute differences in the Monte Carlo estimates used for computing the messages. The pointmass constraint obscures this since it forces the marginal to put all mass on the MAP estimate.

The point of repeating the experiments with $\delta$-constraints is not to show that the behaviour of the agent changes dramatically. Instead, the idea is to demonstrate that \cffgs allow for modular specification of \aif agents which allows for adapting parts of the model without having to touch the rest. In this case, the only parts of inference that change are those involving the control marginal. This means all messages out of the P-substituted composite nodes are unaffected since the $\delta$-constraints are only applied to the control marginals.

\section{Conclusions}
In this paper we have proposed a novel approach to \aif based on lagrangian optimisation which we have named \acf{laif}. We demonstrated \laif on a classic benchmark problem from the \aif literature and found that it inherits the epistemic drive that is a hallmark feature of \aif. \laif presents three main advantages over previous algorithms for \aif.

Firstly, an advantage of \laif is the computational efficiency afforded by being able to pass backward messages instead of needing to perform forwards rollouts for every policy. While \laif is still an iterative procedure, the computational complexity of each iteration scales linearly in the size of the planning horizon $T$ instead of exponentially.

A second advantage is that \laif allows for directly inferring posteriors over control signals instead of relying on a model comparison step based on \efe/\gfe. This means that \laif unifies inference for perception, learning, and actions into a single procedure without any overhead - it all becomes part of the same inference procedure. See our companion paper \cite{vandelaar2023realising} for more details.

Thirdly, \laif is inherently modular and consequently works for freely definable \cffgs, while prior work has focused mostly on specific generative models. Extensions to hierarchical or heterarchical models are straightforward and only require writing out the corresponding \cffg.

We have also introduced \cffg notation for writing down constraints and P-subtitutions on an \ffg. \cffgs are useful not just for \aif but for specifying free energy functionals in general. \cffgs accomplish this through a simple and intuitive graphical syntax. Our hope is that \cffgs can become a standard tool similar to \ffgs when it is desirable to write not just a model $f$ but also a family of approximating distributions $\mathcal{Q}$. Specifically in the context of \aif we have also introduced P-substitution as a way to modify the underlying free energy functional. In doing so we have formalised the relation between \aif and message passing on a \cffg, paving the way for future developments.

In future work, we plan to extend \laif to more node constructions to further open up the scope of available problems that can be attacked using \aif.

\section*{Acknowledgements}
This research was made possible by funding from GN Hearing A/S. This work is part of the research programme Efficient Deep Learning with project number P16-25 project 5, which is (partly) financed by the Netherlands Organisation for Scientific Research (NWO).

We gratefully acknowledge stimulating discussions with the members of the BIASlab research group at the Eindhoven University of Technology and members of VERSES Research Lab, in particular Ismail Senoz, Bart van Erp, Dmitry Bagaev, Karl Friston, Chris Buckley, Conor Heins and Tim Verbelen.

\printbibliography

\appendix
\section{Expected free energy}\label{sec:efe}

This section will focus on the most common alternative functional used for \aif, the \acf{efe}.

\efe is the standard choice for \aif models and is what is found in most \aif focused software such as \texttt{SPM} \cite{friston2014spm} and \texttt{PyMDP} \cite{heins2022pymdp}. In this section, we will cover the details of how \efe is commonly treated.

\efe is defined specifically over future time steps and on a particular choice of generative model. The model in question is a state space model (SSM) of the form

\begin{align}\label{eq:efe_model}
    p(\seq{x}, \seq{z} \mid \hat{\seq{u}}) &=
    \underbrace{p(z_t)}_{\substack{\text{State} \\ \text{prior}}}
    \prod_{k=t+1}^T \underbrace{p(x_k | z_k)}_{\substack{\text{Observation} \\ \text{model}}}
    \underbrace{p(z_k | z_{k-1}, \hat{u}_k)}_{\substack{\text{Transition} \\ \text{model}}}
\end{align}

where $\hat{u}_k$ denotes a particular control parameter that is known apriori and fixed. Note that \eqref{eq:efe_model} does not include a prior over actions $\seq{u}$ and is instead conditional on a fixed policy $\hat{\seq{u}} = \vect{\hat{u}}_{t+1:T}$. We use the $\hat{}$ notation on $\seq{u}$ to indicate that it is treated as a known value instead of a random variable.

The \efe is evaluated as a function of a particular policy. The \efe is defined as \cite{friston_active_2015}

\begin{subequations}\label{eq:efe_classic}
\begin{align}
    G[q; \hat{\seq{u}}]
    &= \sum_{k=t+1}^T \iint p(x_k | z_k) \underbrace{q(z_k|\hat{\seq{u}}_k) \log \frac{q(z_k | \hat{\seq{u}}_k)}{p(x_k, z_k | \hat{\seq{u}}_k)}}_{\text{VFE conditioned on } \hat{\seq{u}}_k} \dd x_k \dd z_k \\
    &= \sum_{k=t+1}^T \iint p(x_k | z_k) q(z_k|\hat{\seq{u}}_k) \log \frac{q(z_k | \hat{\seq{u}}_k)}{p(z_k | x_k, \hat{\seq{u}}_k)} - \log p(x_k) \dd x_k \dd z_k
\end{align}
\end{subequations}

where $p(x_k)$ denotes a goal prior over preferred observations. Note that $p(x_k)$ is not part of the generative model in Eq.~\eqref{eq:efe_model}. To compute Eq.~\eqref{eq:efe_classic} we also need

\begin{align}
    p(x_k, z_k \mid \hat{u}_k) &= \int p(x_k \mid z_k) p(z_k \mid z_{k-1}, \hat{u}_k) {\color{red} q(z_{k-1})} \dd z_{k-1}
\end{align}

meaning we use the {\color{red} forward prediction} from the previous time step to compute $p(x_k, z_k \mid \hat{u}_k)$. If we further we assume that

\begin{align}
    q(x_k,z_k \mid \hat{u}_k) &= p(x_k \mid z_k) q(z_k \mid \hat{u}_k)
\end{align}

It can be shown \cite{da_costa_active_2020,friston_active_2015,koudahl_2021_lgds,sajid_active_2020} that \eqref{eq:efe_classic} can be decomposed into a bound on a mutual information term and a cross-entropy loss between predicted observations and the goal prior.

\section{VFE and GFE}\label{sec:vfe_and_gfe}
In this section, we walk through how the formulation of \gfe given by \cite{parr_generalised_2019} reduces to the \vfe when observations are available. To show how this comes about, we note that \cite{parr_generalised_2019} assumes that

\begin{align}
    q(\vect{x}_k|\vect{z}_k) &=
    \begin{cases}
    \delta(\vect{x}_k - \vect{\hat{x}}_k) & \text{if } k \leq t \\
    p(\vect{x}_k | \vect{z}_k) & \text{if } k > t
    \end{cases} \,.
\end{align}

where $t$ is the current time step and $\vect{\hat{x}}_k$ denotes the observed data point at time step $k$. This is a problematic move since $q(\vect{x}_k \mid \vect{z}_k)$ is no longer a function of $\vect{z}_k$ for $k \leq t$. However if we do take Eq.~\eqref{eq:gfe_assumption} as valid and further assume

\begin{subequations}
\begin{align}
    q(\vect{x}_k, \vect{z}_k \mid \vect{\hat{u}}_k) &= q(\vect{x}_k \mid \vect{z}_k) q(\vect{z}_k \mid \vect{\hat{u}}_k) \\
    \tilde{p}(\vect{x}_k) &= 1 \text{ if } k \leq t \label{eq:p_eq_1}
\end{align}
\end{subequations}

and note that

\begin{align}
    q(\vect{x}_k | \vect{\hat{u}}_k) &= \int q(\vect{x}_k | \vect{z}_k) q(\vect{z}_k | \vect{\hat{u}}_k)  \dd \vect{z}_k \,.
\end{align}

We can plug this result into \eqref{eq:gfe} and obtain for $k \leq t$

\begin{subequations}
\begin{align}
   \notag  &\sum_{k=1}^t \iint q(\vect{x}_k | \vect{z}_k) q(\vect{z}_k \mid \vect{\hat{u}}_k) \log \frac{q(\vect{x}_k \mid \vect{\hat{u}}_k) q(\vect{z}_k \mid \vect{\hat{u}}_k)}{p(\vect{x}_k, \vect{z}_k \mid \vect{\hat{u}}_k )\tilde{p}(\vect{x}_k) } \dd \vect{x}_k \dd \vect{z}_k \\
     &= \sum_{k=1}^t \iint q(\vect{x}_k | \vect{z}_k) q(\vect{z}_k \mid \hat{u}_k) \log \frac{  \int \left[ q(\vect{x}_k | \vect{z}_k) q(\vect{z}_k \mid \vect{\hat{u}}_k) \right] \dd \vect{z}_k q(\vect{z}_k \mid \vect{\hat{u}}_k)}{p(\vect{x}_k, \vect{z}_k \mid \vect{\hat{u}}_k)} \dd \vect{x}_k \dd \vect{z}_k
\end{align}
\end{subequations}

where the last line follows from Eq.~\eqref{eq:p_eq_1}. Then by \eqref{eq:gfe_assumption} we have

\begin{align}
    &=\sum_{k=1}^t  \iint \delta(\vect{x}_k - \vect{\hat{x}}_k) q(\vect{z}_k \mid \vect{\hat{u}}_k) \log \frac{  \int \left[ \delta(\vect{x}_k - \vect{\hat{x}}_k) q(\vect{z}_k \mid \vect{\hat{u}}_k) \right] \dd \vect{z}_k q(\vect{z}_k \mid \vect{\hat{u}}_k)}{p(\vect{x}_k, \vect{z}_k \mid \vect{\hat{u}}_k)} \dd \vect{x}_k \dd \vect{z}_k
\end{align}

Now we pull $\delta(\vect{x}_k - \hat{x}_k)$ out of the integral and solve to find

\begin{subequations}
\begin{align}
    &=\sum_{k=1}^t   \iint \delta(\vect{x}_k - \vect{\hat{x}}_k) q(\vect{z}_k \mid \vect{\hat{u}}_k) \log \frac{  \delta(\vect{x}_k - \vect{\hat{x}}_k) \overbrace{\Big[ \int q(\vect{z}_k \mid \vect{\hat{u}}_k) \dd \vect{z}_k\Big] }^{\text{Integrates to 1}} q(\vect{z}_k \mid \vect{\hat{u}}_k)}{p(\vect{x}_k, \vect{z}_k \mid \vect{\hat{u}}_k) } \dd \vect{x}_k \dd \vect{z}_k \\
    &=\sum_{k=1}^t   \int q(\vect{z}_k \mid \vect{\hat{u}}_k) \log \frac{q(\vect{z}_k \mid \vect{\hat{u}}_k)}{p(\vect{\hat{x}}_k , \vect{z}_k \mid \vect{\hat{u}}_k) } \dd \vect{z}_k
\end{align}
\end{subequations}

which we can recognise as a \vfe with data constraints.

\section{The transition mixture node}\label{sec:transition_mixture}
In this section, we derive the \mpa rules required for the Transition Mixture node used in our experiments on policy inference. Before doing so, we first establish some preliminary results for the categorical distribution and the standard transition node. In this section we will also on occasion use an $\overline{\text{overbar}}$ to indicate an expectation, $\mathbb{E}_{q(x)}[g(x)] = \overline{g(x)}$ The categorical distribution is given by

\begin{subequations}
\begin{align}
    \mathcal{C}at(\vect{x} \mid \vect{z})
    &= \prod_{i = 1}^I \vect{z}_i^{\vect{x}_i} \label{eq:categorical_dist_1} \\
    &= \sum_{i = 1}^I \vect{x}_i \vect{z}_i \label{eq:categorical_dist_2}
\end{align}
\end{subequations}

where $\vect{x}$ and $\vect{z}$ are both one-hot encoded vectors. The move Eq.~\eqref{eq:categorical_dist_1} to Eq.~\eqref{eq:categorical_dist_2} is possible only because $\vect{x}$ is one-hot encoded. Similarly, we have for the standard Transition node that

\begin{subequations}
\begin{align}
    \mathcal{C}at(\vect{x} \mid \matr{A} \vect{z})
    &= \prod_{i = 1}^I \prod_{j = 1}^I \big[ \matr{A}_{ij} \big]^{\vect{x}_i \vect{z}_j } \\
    &= \sum_{i = 1}^I \sum_{j = 1}^I \vect{x}_i \vect{z}_j \matr{A}_{ij}
\end{align}
\end{subequations}

and again, since $\vect{x}$ and $\vect{z}$ are one-hot encoded

\begin{subequations}
\begin{align}
   \log \mathcal{C}at(\vect{x} \mid \matr{A} \vect{z})
   &= \log \Bigg[ \prod_{i = 1}^I \prod_{j = 1}^I \big[ \matr{A}_{ij} \big]^{\vect{x}_i \vect{z}_j } \Bigg] \\
   &= \sum_{i = 1}^I \sum_{j = 1}^I \log \Bigg[  \matr{A}_{ij}^{\vect{x}_i \vect{z}_j} \Bigg] \\
   &= \sum_{i = 1}^I \sum_{j = 1}^I \vect{x}_i \vect{z}_j \log \matr{A}_{ij}
\end{align}
\end{subequations}

Now, we are ready to start derivations for the transition mixture node. The transition mixture node has the node function

\begin{subequations}
\begin{align}
    f(\vect{x},\vect{y},\vect{z},\matr{A})
    &= \prod_k \mathcal{C}at \big( \vect{x} \mid \matr{A}_k \vect{z} \big)^{\vect{y}_k} \\
    &= \prod_k \Bigg[ \prod_i \prod_j \big[ \matr{A}_k \big]_{ij}^{\vect{z}_i \vect{x}_j} \Bigg]^{\vect{y}_k} \\
    &= \prod_{i,j,k} \matr{A}_{ijk}^{\vect{z}_i \vect{x}_j \vect{y}_k}
\end{align}
\end{subequations}

Where we use $i$ to index over columns, $j$ to index over rows and $k$ to index over factors. $\matr{A}$ is a 3-tensor that is normalised over columns. In other words, each slice of $\matr{A}$ corresponds to a valid transition matrix and slices are indexed by $k$. We assume a structured mean-field factorisation such that

\begin{align}
    q(\vect{x},\vect{y},\vect{z},\matr{A})
    &= q(\vect{x},\vect{y},\vect{z}) \prod_k q(\matr{A}_k)
\end{align}

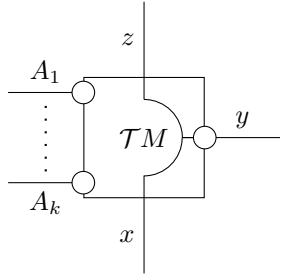
\begin{figure}[H]
    \centering
    \begin{tikzpicture}
\node[bigbox](f){$\mathcal{T}M$};

\bigarcne{f};
\bigarcse{f};

\draw[](f.north)--node[left](){$z$}++(0.0,1.0);
\draw[](f.south)--node[left](){$x$}++(0.0,-1.0);
\draw[](f.east)--node[q,pos=-0.0](){}node[above](){$y$}++(1.0,0.0);

\draw[](f.west)++(0.0,0.6)--node[q,pos=0](){}node[above](){$A_1$}++(-1.0,0.0);
\draw[](f.west)++(0.0,-0.6)--node[q,pos=0](){}node[below](){$A_k$}++(-1.0,0.0);

\draw[loosely dotted, thick](f.west)++(-0.5,-0.45)--++(0.0,1.0);

\end{tikzpicture}
    \caption{The transition mixture node}
    \label{fig:transition_mixture}
\end{figure}

The \cffg of the transition mixture node is shown in Fig.~\ref{fig:transition_mixture}

We now define

\begin{subequations}
\begin{align}
    \log f_{\matr{A}}(\vect{x},\vect{y},\vect{z})
    &= \mathbb{E}_{q(\matr{A})} \Big[ \log f(\vect{x},\vect{y},\vect{z},\matr{A}) \Big] \\
    &= \mathbb{E}_{q(\matr{A})} \Big[\sum_{ijk} \vect{z}_i \vect{x}_j \vect{y}_k \log \matr{A}_{ijk} \Big] \\
    &= \sum_{k} \vect{y}_k \mathbb{E}_{q(\matr{A}_k)} \Big[\sum_{ij} \vect{z}_i \vect{x}_j \log \matr{A}_{ijk} \Big] \\
    &= \sum_{k} \vect{y}_k \sum_{ij} \vect{z}_i \vect{x}_j \Big[ \overline{\log \matr{A}_{k}}\Big]_{ij}
\end{align}
\end{subequations}

which implies that

\begin{subequations}
\begin{align}
    f_A(\vect{x},\vect{y},\vect{z})
    &= \exp \Bigg( \sum_{k} \vect{y}_k \sum_{ijk} \vect{z}_i \vect{x}_j \Big[ \overline{\log \matr{A}_{k}} \Big]_{ij} \Bigg) \\
    &= \prod_{ijk} \exp \Big( \overline{\log \matr{A}_{k}} \Big)^{\vect{z}_i \vect{x}_j \vect{y}_k} \\
    &= \sum_{ijk} \vect{z}_i \vect{x}_j \vect{y}_k \underbrace{ \exp \Big( \overline{\log \matr{A}_k} \Big)_{ij}}_{\tilde{\matr{A}}_{ijk}}
\end{align}
\end{subequations}

Now, we are ready to derive the desired messages. The first message $\nu(\cdot)$ will be towards $\vect{x}$. If we use $\mu(\vect{y})$ to denote the incoming message on the edge $\vect{y}$ (similar for $\mu{\vect{z}}$), then the message towards $\vect{x}$ is given by

\begin{subequations}
\begin{align}
    \nu(\vect{x})
    &= \sum_{\vect{z}} \sum_{\vect{y}} \mu(\vect{z}) \mu(\vect{y}) \exp \Big( \mathbb{E}_{q(\matr{A})} \log f(\vect{x},\vect{y},\vect{z},\matr{A}) \Big) \\
    &= \sum_{\vect{z}} \sum_{\vect{y}} \mu(\vect{z}) \mu(\vect{y}) f_A(\vect{x},\vect{y},\vect{z}) \Big) \\
    &= \sum_{\vect{z}} \sum_{\vect{y}} \mu(\vect{z}) \mu(\vect{y}) \sum_{ijk} \vect{z}_i \vect{x}_j \vect{y}_k \tilde{\matr{A}}_{ijk}
\end{align}
\end{subequations}

Assuming the incoming messages are Categorically distributed, we can continue as

\begin{subequations}
\begin{align}
    &= \mathbb{E}_{\mathcal{C}at(\vect{z} \mid \pi_z)} \mathbb{E}_{\mathcal{C}at(\vect{y} \mid \pi_y)} \sum_{ijk} \vect{z}_i \vect{x}_j \vect{y}_k \tilde{\matr{A}}_{ijk} \\
    &= \sum_{ijk} \pi_{zi} \pi_{yk} \vect{x}_j \tilde{\matr{A}}_{ijk} \\
    &= \prod_j \Big[ \sum_{ik} \pi_{zi} \pi_{yk} \tilde{\matr{A}}_{ijk} \Big]^{\vect{x}_j}
\end{align}
\end{subequations}

where the last line follows from $\vect{x}$ being one-hot encoded. At this point, we can recognise the message

\begin{align}\label{eq:nu_x}
    \nu(\vect{x})
    &\propto \mathcal{C}at(\vect{x} \mid \vect{\rho}) \text{ where } \vect{\rho}_j = \frac{\sum_{ik} \pi_{zi} \pi_{yk} \tilde{\matr{A}}_{ijk}}{\sum_{ijk} \pi_{zi} \pi_{yk} \tilde{\matr{A}}_{ijk}}
\end{align}

By symmetry, similar results hold for messages $\nu(\vect{z})$ and $\nu(\vect{y})$. $\nu(\vect{z})$ is given by

\begin{subequations}
\begin{align}
    \nu(\vect{z})
    &=  \sum_x \sum_y \mu(x) \mu(\vect{y}) f_A(\vect{x},\vect{y},\vect{z}) \\
    &\propto \mathcal{C}at(\vect{z} \mid \vect{\rho}) \text{ where } \vect{\rho}_i = \frac{\sum_{jk} \pi_{xj} \pi_{yk} \tilde{\matr{A}}_{ijk}}{\sum_{ijk} \pi_{xj} \pi_{yk} \tilde{\matr{A}}_{ijk}} \,.
\end{align}
\end{subequations}

Similarly, $\nu(\vect{y})$ evaluates to
\begin{subequations}
\begin{align}
    \nu(\vect{y})
    &=  \sum_x \sum_z \mu(\vect{x}) \mu(\vect{z}) f_A(\vect{x},\vect{y},\vect{z}) \\
    &\propto \mathcal{C}at(\vect{y} \mid \vect{\rho}) \text{ where } \vect{\rho}_k = \frac{\sum_{ij} \pi_{x,j} \pi_{z,i} \tilde{\matr{A}}_{ijk}}{\sum_{ijk} \pi_{x,j} \pi_{z,i} \tilde{\matr{A}}_{ijk}} \,.
\end{align}
\end{subequations}

To compute the message towards the $n$'th candidate transition matrix $\matr{A}_n$, we will need to take an expectation with respect to $q(\vect{x},\vect{y},\vect{z})$. It is given by

\begin{subequations}
\begin{align}
    q(\vect{x},\vect{y},\vect{z})
    &= \mu(\vect{y}) \mu(\vect{x}) \mu(\vect{z}) f_A(\vect{x},\vect{y},\vect{z}) \\
    &= \prod_{ijk} \pi_{xj}^{\vect{x}_j} \pi_{zi}^{\vect{z}_i} \pi_{yk}^{\vect{y}_k} \tilde{\matr{A}}_{ijk}^{\vect{x}_j \vect{z}_i \vect{y}_k} \\
    &= \prod_{ijk} \Big[ \pi_{xj} \pi_{zi} \pi_{yk} \tilde{\matr{A}}_{ijk} \Big]^{\vect{x}_j \vect{z}_i \vect{y}_k} \\
    &\propto \mathcal{C}at(\vect{x},\vect{y},\vect{z} \mid \matr{B})
\end{align}
\end{subequations}

where $\matr{B}$ is a three-dimensional contingency tensor with entries

\begin{align}
    \matr{B}_{ijk} &= \frac{ \pi_{xj} \pi_{zi} \pi_{yk} \tilde{\matr{A}}_{ijk}}{ \sum_{ijk} \pi_{xj} \pi_{zi} \pi_{yk} \tilde{\matr{A}}_{ijk}}
\end{align}

Now we can compute the message towards a transition matrix $\matr{A}_n$ as

\begin{subequations}
\begin{align}
    \log \nu(\matr{A}_n)
    &= \mathbb{E}_{q(\matr{A}_{ k \backslash n}) q(\vect{x},\vect{y},\vect{z})} \Big[ \log f(\vect{x},\vect{y},\vect{z},\matr{A}) \Big] \\
    &= \mathbb{E}_{q(\matr{A}_{ k \backslash n}) q(\vect{x},\vect{y},\vect{z})} \Big[ \sum_{ijk} \vect{x}_j \vect{z}_i \vect{y}_k \log \matr{A}_{ijk} \Big] \\
    &= \mathbb{E}_{q(\matr{A}_{ k \backslash n})} \Big[ \sum_{ijk} \matr{B}_{ijk} \log \matr{A}_{ijk} \Big] \\
    &= \underbrace{\sum_{k \backslash n} \Big[ \sum_{ijk} \matr{B}_{ijk} \log \matr{A}_{ijk} \Big]}_{\text{Constant w.r.t $\matr{A}_n$}} + \sum_{ij} \matr{B}_{ijn} \log \matr{A}_{ijn} \\
    &\propto tr \Big( \matr{B}_{ijn} \log \matr{A}_{ijn} \Big)
\end{align}
\end{subequations}

which implies that

\begin{align}
    \nu(\matr{A}_n) &\propto \mathcal{D}ir(\matr{A}_n \mid \matr{B}_n +1) \,.
\end{align}

We see that messages towards each component in $\matr{A}$ are distributed according to a Dirichlet distribution with parameters $\matr{B}_n + 1$. The final expression we need to derive is the average energy term for the transition mixture node. It is given by

\begin{subequations}
\begin{align}
   U_x[q]
   &= - \mathbb{E}_{q(\matr{A})q(\vect{x},\vect{y},\vect{z})} \Big( \log f(\vect{x},\vect{y},\vect{z},\matr{A}) \Big) \\
   &= - \mathbb{E}_{q(\vect{x},\vect{y},\vect{z})} \Big( f_A(\vect{x},\vect{y},\vect{z}) \Big) \\
   &= - \mathbb{E}_{q(\vect{x},\vect{y},\vect{z})} \Big[ \sum_{ijk} \vect{x}_j \vect{z}_i \vect{y}_k \big[ \overline{ \log \matr{A}_k} \big]_{ij}  \Big) \\
   &= - \sum_{ijk} \matr{B}_{ijk} \big[ \overline{ \log \matr{A}_k}\big]_{ij} \\
   &= - \sum_k tr \Big( \matr{B}_{k}^T \overline{\log \matr{A}_k} \Big)
\end{align}
\end{subequations}

\end{document}